\documentclass[sigconf]{acmart} 

\AtBeginDocument{%
  \providecommand\BibTeX{{%
    \normalfont B\kern-0.5em{\scshape i\kern-0.25em b}\kern-0.8em\TeX}}}

\usepackage{color}
\definecolor{myred}{rgb}{.8,.0,.0}
\definecolor{myblue}{rgb}{.0,.0,.8}

\newcommand{\JimSan}{Jim\'{e}nez-S\'{a}nchez}

\usepackage{subcaption}
\usepackage{multirow}

\usepackage{array}
\newcolumntype{P}[1]{>{\centering}p{#1}}

\usepackage{mydefs}
\usepackage{acmart-taps}
\usepackage{framed}
\usepackage{fancybox}

\copyrightyear{2025}
\acmYear{2025}
\setcopyright{cc}
\setcctype{by}
\acmConference[FAccT '25]{The 2025 ACM Conference on Fairness, Accountability, and Transparency}{June 23--26, 2025}{Athens, Greece}
\acmBooktitle{The 2025 ACM Conference on Fairness, Accountability, and Transparency (FAccT '25), June 23--26, 2025, Athens, Greece}
\acmDOI{10.1145/3715275.3732035}
\acmISBN{979-8-4007-1482-5/2025/06}

\newcommand{\chexpert}{CheXpert}

\makeatletter
\ifcase\ACM@format@nr
\relax 
\or 
  \def\@authorfont{\large\sffamily}
  \def\@affiliationfont{\small\normalfont}
\or 
\or 
  \def\@authorfont{\LARGE\sffamily}
  \def\@affiliationfont{\large}
\or 
  \def\@authorfont{\LARGE}
  \def\@affiliationfont{\small}
\or 
  \def\@authorfont{\normalsize\normalfont}
  \def\@affiliationfont{\normalsize\normalfont}
\or 
  \def\@authorfont{\Large\normalfont}
  \def\@affiliationfont{\normalsize\normalfont}
\or 
  \def\@authorfont{\bfseries}
  \def\@affiliationfont{\mdseries}
\or 
  \def\@authorfont{\bfseries}
  \def\@affiliationfont{\mdseries}
\or 
  \def\@authorfont{\LARGE}
  \def\@affiliationfont{\large}
\or 
  \def\@authorfont{\large\sffamily}
  \def\@affiliationfont{\small\normalfont}
\fi
\makeatother 

\newcommand{\ourpar}[1]{\paragraph{\textbf{#1}}}
\begin{document}

\title{In the Picture: Medical Imaging Datasets, Artifacts, and their Living Review}

\author{Amelia \JimSan}
\affiliation{
  \institution{ITU}
  \country{Denmark}
}
\orcid{0000-0001-7870-0603}

\author{Natalia-Rozalia Avlona}
\affiliation{
  \institution{KU}
  \country{Denmark}
}

\author{Sarah de Boer}
\affiliation{
\institution{Radboudumc}
  \country{Netherlands}
}

\author{V\'{i}ctor M. Campello}
\affiliation{
\institution{UB}
  \country{Spain}
}

\author{Aasa Feragen}
\affiliation{
  \institution{DTU}
  \country{Denmark}
}

\author{Enzo Ferrante}
\affiliation{
\institution{CONICET \& UBA}
  \country{Argentina}
}

\author{Melanie Ganz}
\affiliation{
  \institution{KU \& Rigshospitalet}
  \country{Denmark}
}

\author{Judy Wawira Gichoya}
\affiliation{
  \institution{Emory University}
  \country{USA}
}

\author{Camila Gonz\'{a}lez}
\affiliation{
  \institution{Stanford University}
  \country{USA}
}

\author{Steff Groefsema}
\affiliation{
  \institution{RUG}
  \country{Netherlands}
}

\author{Alessa Hering}
\affiliation{
\institution{Radboudumc}
  \country{Netherlands}
}

\author{Adam Hulman}
\affiliation{
\institution{AUH \& AU}
  \country{Denmark}
}

\author{Leo Joskowicz}
\affiliation{
  \institution{HUJI}
  \country{Israel}
}

\author{Dovile Juodelyte}
\affiliation{
  \institution{ITU}
  \country{Denmark}
}

\author{Melih Kandemir}
\affiliation{
  \institution{SDU}
  \country{Denmark}
}

\author{Thijs Kooi}
\affiliation{
  \institution{Lunit}
  \country{South Korea}
}

\author{Jorge del Pozo L\'{e}rida}
\affiliation{
  \institution{ITU \& Cerebriu A/S}
  \country{Denmark}
}

\author{Livie Yumeng Li}
\affiliation{
\institution{AUH \& AU}
  \country{Denmark}
}

\author{Andre Pacheco}
\affiliation{
\institution{UFES}
  \country{Brazil}
}

\author{Tim R\"{a}dsch}
\affiliation{
\institution{DKFZ \& UHEI}
  \country{Germany}
}

\author{Mauricio Reyes}
\affiliation{
\institution{UniBE}
  \country{Switzerland}
}

\author{Th\'{e}o Sourget}
\affiliation{
  \institution{ITU}
  \country{Denmark}
}

\author{Bram van Ginneken}
\affiliation{
  \institution{Radboudumc \& Plain Medical}
  \country{Netherlands}
}

\author{David Wen}
\affiliation{
  \institution{Oxford University Hospitals}
  \country{UK}
}

\author{Nina Weng}
\affiliation{
  \institution{DTU}
  \country{Denmark}
}

\author{Jack Junchi Xu}
\affiliation{
  \institution{Copenhagen University Hospital \& RAIT}
  \country{Denmark}
}

\author{Hubert Dariusz Zając}
\affiliation{
  \institution{KU}
  \country{Denmark}
}

\author{Maria A. Zuluaga}
\affiliation{
\institution{EURECOM}
  \country{France}
}

\author{Veronika Cheplygina}
\affiliation{%
  \institution{ITU}
  \country{Denmark}
}
\orcid{0000-0003-0176-9324}

\renewcommand{\shortauthors}{\JimSan, et al.}

\newcommand{\ra}[2]{\renewcommand{\arraystretch}{#1}}

\begin{abstract}
Datasets play a critical role in medical imaging research, yet issues such as label quality, shortcuts, and metadata are often overlooked. This lack of attention may harm the generalizability of algorithms and, consequently, negatively impact patient outcomes. While existing medical imaging literature reviews mostly focus on machine learning (ML) methods, with only a few focusing on datasets for specific applications, these reviews remain static -- they are published once and not updated thereafter. This fails to account for emerging evidence, such as biases, shortcuts, and additional annotations that other researchers may contribute after the dataset is published. We refer to these newly discovered findings of datasets as \emph{research artifacts}. To address this gap, we propose a \emph{living review} that continuously tracks public datasets and their associated research artifacts across multiple medical imaging applications. Our approach includes a framework for the living review to monitor data documentation artifacts, and an SQL database to visualize the citation relationships between research artifact and dataset. Lastly, we discuss key considerations for creating medical imaging datasets, review best practices for data annotation, discuss the significance of shortcuts and demographic diversity, and emphasize the importance of managing datasets throughout their entire lifecycle. Our demo is publicly available at~\url{http://inthepicture.itu.dk/}. 

\end{abstract}



\maketitle


\section{Introduction} \label{sec:intro} 
High-quality datasets are a key element to the development of machine learning (ML) models for medical imaging applications and, more generally, for healthcare. Such datasets are characterized by having diverse representation of patients, sufficient sample sizes, accurate labels or annotations, and comprehensive documentation. Failing to meet these requirements has a direct impact on a model's robustness and reliability \cite{ricci2022addressing,seyyed2021underdiagnosis,compton2023more,bissoto2020debiasing,zajkac2023ground}, thereby affecting the model's clinical utility. Inaccurate or incomplete annotations can lead to models that make incorrect predictions. Insufficient or lack of documentation may miss information such as demographics or hospital scanner, leading to biased and inaccurate models \cite{oakden2020hidden, chen2021data}. A lack of diversity limits the generalizability of the model across heterogeneous patient populations, whereas a sufficient sample size is necessary to ensure that the model can learn meaningful patterns and avoid overfitting. Ultimately, dataset quality is as important as the choice of the method to build the ML model.

Despite the critical importance of data, current efforts in medical imaging do not consider the evolving nature of datasets \cite{hutchinson2021towards}. Such datasets often consist of two parts: images such as chest X-rays, and target labels such as lung diseases. However, additional evidence about these datasets, such as shortcuts, biases, or additional annotations, emerges over time, but is often not available in the original dataset documentation. We refer to these newly discovered aspects of datasets as \emph{research artifacts}. Literature reviews, both in medical imaging and in ML more generally, primarily focus on ML models (see \cite{litjens2017survey,cheplygina2019not,yi2019generative,shamshad2023transformers,budd2021survey} for examples), with only a few addressing specific issues such as fairness in model predictions \cite{chen2021ethical,chen2023algorithm,ricci2022addressing}. Some reviews summarize available datasets within a specific application such as dermatology \cite{daneshjou2021lack,wen2022characteristics} or ophthalmology \cite{khan2021ophtalmologicalreview}. A recent work \cite{galanty2024assessingdoc} assesses the documentation of publicly available magnetic resonance imaging (MRI), color fundus photography, and electrocardiogram datasets, yet carries out a static review that is reviewed and published once, without subsequent updates. Finally, datasets released by international competitions are often used after the competition for benchmarking state-of-the-art models \cite{eisenmann2022biomedical}. Although the construction of these datasets is crucial for interpreting results, post-competition analyses are centered around model performance, limiting the discussion about the data to a snapshot of its main features (e.g., sample size or scanning devices). On a positive note, competitions do sometimes track the evolution of performance over time, recognizing the dynamic nature of both the datasets and methods. Overall, we believe that the medical imaging field is in need of dynamic or living reviews, inspired by efforts like living reviews of evidence on COVID-19 \cite{wynants2020prediction}.

\begin{figure*}[t]
    \centering
    \includegraphics[width=0.75\linewidth]{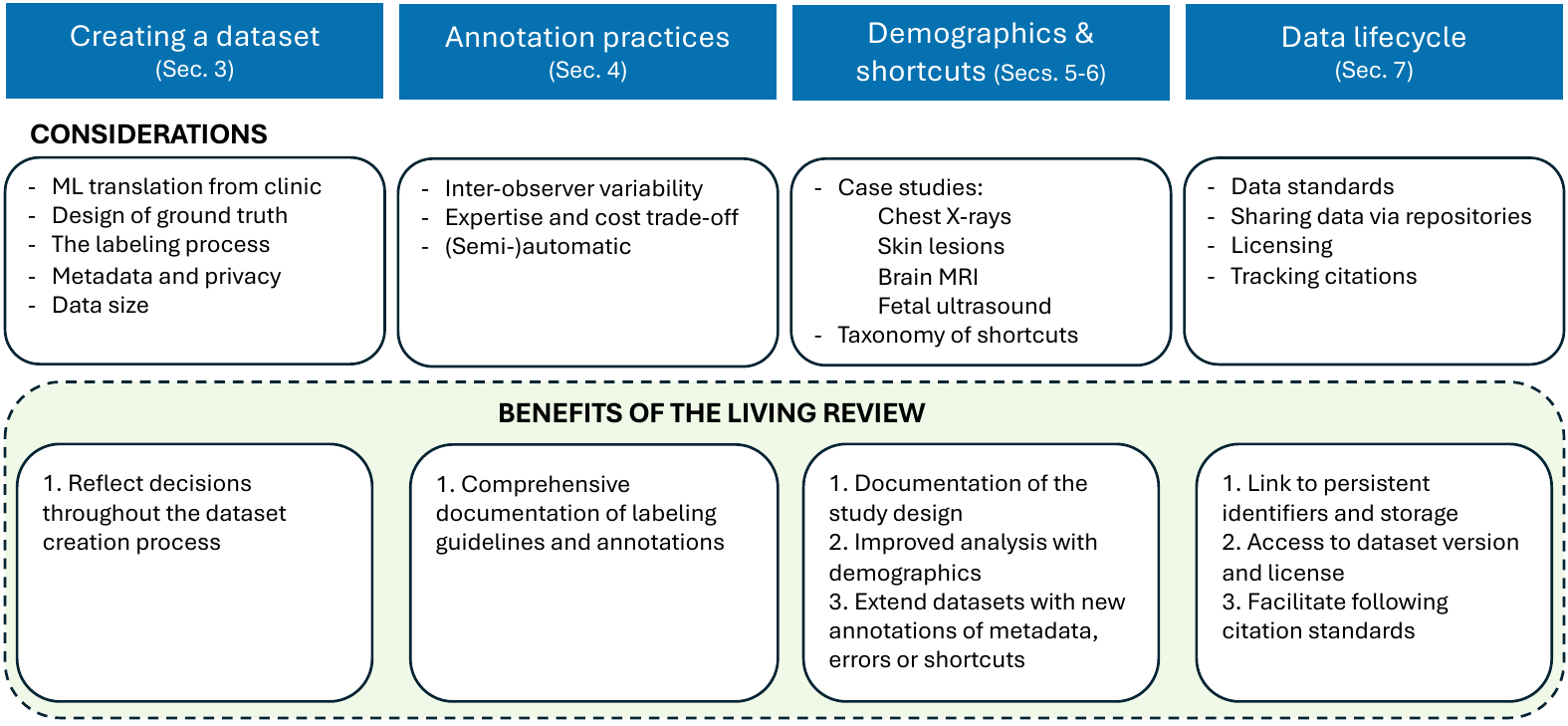}
    \caption{Benefits of our proposed living review framework (Sec.~\ref{sec:livingreview}). Overview of considerations for creating a dataset (Sec.~\ref{sec:creatingdata}), data annotation practices and quality (Sec.~\ref{sec:labelquality}), patient demographics (Sec.~\ref{sec:demographics}) and shortcuts (Sec.~\ref{sec:shortcuts}), and dataset lifecycle (Sec.~\ref{sec:lifecycle}).} \label{fig:schematic_overview_workflow}
    \Description{This figures summarizes the structure of this manuscript highlighting the benefits of our proposed living review for medical imaging datasets. It provides an overview of considerations for creating a dataset (Sec.~\ref{sec:creatingdata}), data annotation practices and quality (Sec.~\ref{sec:labelquality}), patient demographics (Sec.~\ref{sec:demographics}) and shortcuts (Sec.~\ref{sec:shortcuts}), and dataset lifecycle (Sec.~\ref{sec:lifecycle}).
    }
\end{figure*}

We aim to promote collaborative extensions of datasets and avoid snapshots or static datasets. The research we present was conducted in the context of a year-long, collaborative webinar, as well as an in-person workshop on current developments and challenges of medical imaging data. These events brought together a group of around 50 researchers from academia, industry, and clinicians, with backgrounds from ML to epidemiology and human-computer interaction, and research experience from 10+ countries in five continents. This paper is a synthesis of discussions during and after the webinars and workshop, which spanned topics across the entire dataset lifecycle, from creating medical imaging datasets to their use for validating algorithms to data governance, see Fig.~\ref{fig:schematic_overview_workflow}. Our contributions are as follows:
\begin{itemize}
    \item We present a proof of concept for a \emph{living review} (publications and database, see Fig.~\ref{fig:livingreviewzenododb}) - a framework for enhancing dataset metadata through research artifacts, facilitating the discovery of emerging information about medical imaging datasets.
    \item We discuss key considerations for creating datasets, best practices for data annotation, significance of demographics and shortcuts, and data management throughout the dataset lifecycle.
    \item We release a demo of a living database of 16 datasets along with their connections and relationships to 24 research artifacts: shortcuts, annotations, and derivatives, for two medical image applications.
    \item We discuss how the research community can contribute to our living review, and invite them to do so.
\end{itemize}

\section{Proposal: a living review of medical imaging datasets} \label{sec:livingreview}
Development of novel algorithms with state-of-the-art results is considered the more prestigious activity within the ML field \cite{birhane2021values,sambasivan2021everyone}, leading to datasets often being considered ``as-is'' benchmarks. However, medical imaging datasets not only are foundational to ML development but they also evolve over time, both explicitly and implicitly. A dataset evolves \emph{explicitly} if the data itself is updated, for example, due to errors. A dataset evolves \emph{implicitly} as new evidence emerges, such as erroneous target labels, additional annotations, and shortcuts \cite{oakden2020exploring,oakden2020hidden}. For example, the \chexpert{} dataset \cite{irvin2019chexpert} was initially published without the recommended datasheet \cite{gebru2021datasheets}. A datasheet co-authored by some of the \chexpert{} authors was later released \cite{garbin2021structured}, but this datasheet is not linked from the original dataset. By not incorporating such evidence into the original dataset, it is often not taken into consideration in subsequent research.
 
Traditional systematic reviews of medical imaging datasets, such as \cite{daneshjou2021lack,wen2022characteristics,khan2021ophtalmologicalreview,li2023systematic}, do not capture this evolution. We argue that we need a \emph{living review} to keep track of novel open datasets, as well as emerging connections and issues in existing datasets. Living systematic reviews are a more recent development in meta-research, but they are crucial for rapidly evolving topics. Given the rapid developments in ML, it would be advantageous to adopt a similar framework for datasets. Our proposed work is different from these efforts, since \cite{wynants2020prediction} focuses on COVID-19 findings, not specifically on medical image datasets. \cite{schmidt2021data,schmidt2023automated} focus on data extraction methods to aid doing systematic reviews, of these only \cite{schmidt2021data} is a living review itself, which could provide inspiration for how to carry out our plans.

Here we outline our vision for the living review of open datasets, studying the relationships between the datasets and their research artifacts. We propose a framework of these relationships, as well as protocols for the maintenance of the living review. 

\begin{figure*}[]
    \centering
    \includegraphics[width=0.7\linewidth]{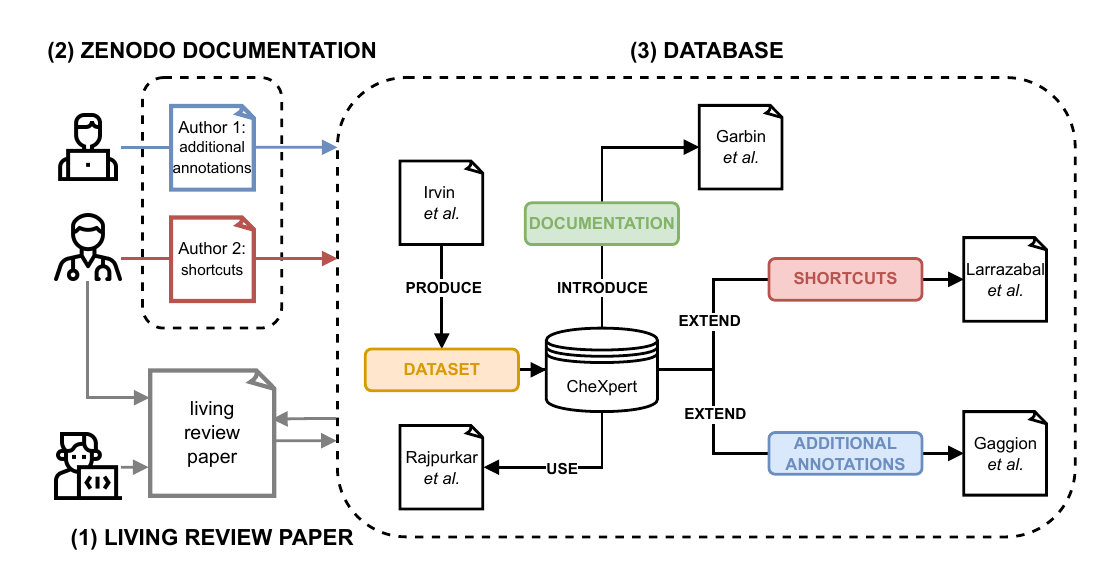}
    \caption{
    Our living review consists of three components: (1) an overarching living review paper, (2) documentation of research artifacts via publications on Zenodo linked by living review paper, and (3) a SQL database for exploring the links between the datasets and the research artifacts.}
    \label{fig:livingreviewzenododb}
    \Description{Our living review consists of three components: (1) an overarching living review paper, (2) documentation of research artifacts via publications on Zenodo linked by living review paper, and (3) a SQL database for exploring the links between the datasets and the research artifacts.}
\end{figure*}

\ourpar{Conceptualization}
Our proposal for the living review consists of three parts, as illustrated in Fig.~\ref{fig:livingreviewzenododb}: 

\begin{enumerate} 
\item an overarching living review publication,
\item documentation of research artifacts via dataset-specific publications on Zenodo, which the overarching paper links to,
\item a SQL database for exploring the links between the datasets and the research artifacts. 
\end{enumerate}

An ``artifact'', like a Greek amphora, is a produced object -- the output of a process. We define a research artifact as any additional evidence related to a dataset, for example derived datasets like PruneCXR \cite{holste2023prunecxr} and LongTailCXR \cite{holste2022longtailcxr}, additional annotations like spurious correlations \cite{damgaard2023augmenting},  segmentation masks \cite{gaggion2023chexmask}, or compressed information in the form of embeddings \cite{sellergrengeneralized}. Such artifacts are crucial as they provide the necessary context for understanding, reproducing and validating research findings. Moreover, computer science papers with shared artifacts receive about 75\% more citations than those without \cite{frachtenberg2022research}, underscoring the critical role of artifacts in the visibility and impact of research. 

We map the relation between the datasets and the research artifacts with the citation function (use, produce, extend, introduce, other). We exemplify this for the \chexpert{} dataset in Fig.~\ref{fig:livingreviewzenododb}, for more details and examples please see Tables~\ref{tab:resources}-\ref{tab:citationfunctions}-\ref{tab:relations}.

\ourpar{Implementation}
The overarching, continuously updated, living review publication will link the dataset-specific publications (peer-reviewed or preprints) where smaller teams can document datasets by building on existing guidelines \cite{rostamzadeh2022healthsheet,garbin2022assessing}, and extending them by documenting resources that provide additional evidence related to responsible use of these datasets, such as demographic biases (Section~\ref{sec:demographics}) or shortcuts (Section~\ref{sec:shortcuts}). These datasheets could be hosted on Zenodo \cite{zenodo2013} which allows each datasheet to have its own Digital Object Identifier (DOI). To enhance interoperability and findability, we could package the documentation in a structured format like Croissant \cite{akhtar2024croissant}. At regular intervals, we can update the overarching living publication, linking to the newly created datasheets.

The SQL database consists of three tables (see Fig.~\ref{fig:database}): datasets, papers documenting their research artifacts, and the dataset-artifact relations. Our interactive demo~\url{http://130.226.140.142} (built with PostgreSQL, Python, Streamlit and the \texttt{st\_link\_analysis} package), supports dynamic dataset-artifact exploration. For additional technical details, see Section~\ref{sec:applivingreview}. In the demo, we present 24 research artifacts related to 16 datasets, which include additional annotations, evidence of shortcuts, and derived datasets, focused on two applications: skin lesions and chest X-rays. 

\ourpar{Maintenance}
It is unrealistic to expect that we as authors of this work would be able to cover all areas of medical imaging, to maintain the living review in perpetuity, nor to ensure that all researchers are aware of its existence. We therefore propose a structure that promotes collaboration, incentivizes better practices around dataset documentation and citation, and regularly informs (new and existing) dataset users about new evidence. Further versions of the living review will require contributions from the community. During data-centric initiatives (in-person and online to include participants who are typically underrepresented at in-person events), we invite researchers to contribute to the living review. A limitation of the current proposal is that we do not yet integrate any mechanism for quality assessment, such as a data steward who will monitor the accuracy of the contributions. 

\section{Considerations for creating a dataset} \label{sec:creatingdata}
\aptLtoX[graphic=no,type=html]{\begin{aptdispbox}\noindent\textbf{Relevance to the living review.} Researchers can access the recommended documentation to better understand decisions made throughout the dataset creation process, including the translation from the clinical problem, balancing metadata sharing with patient privacy, and considerations regarding dataset size.\end{aptdispbox}}{
\noindent\fcolorbox{cyan}{cyan!10}{%
  \begin{minipage}{0.9\columnwidth}
\textbf{Relevance to the living review.} Researchers can access the recommended documentation to better understand decisions made throughout the dataset creation process, including the translation from the clinical problem, balancing metadata sharing with patient privacy, and considerations regarding dataset size.
  \end{minipage}
}}

\ourpar{Translation of clinical problem to ML problem}
ML applications in healthcare are often related to specific parts of the treatment of a patient. Consequently, the development of a dataset will depend on translating these clinical challenges into ML problems that can be evaluated with ML metrics. This translation requires aligning expectations among various stakeholders in the multidisciplinary team while also establishing a utility criterion, for example, a reduction of workload of the radiology department~\cite{Yoon2023,Tong2023}. The right definition of the ML problem is essential for determining the necessary data (whether newly collected or already existing, such as from retrospective studies), designing an appropriate labeling process, assessing the usefulness of the proposed methods, and understanding the potential limitations of such evaluations from a clinical standpoint~\cite{reinke2023commonlimitationsimageprocessing}. This step also helps to identify potential risks or pitfalls in the data creation process, where biases or shortcuts may be introduced. Importantly, we must remember that an ML dataset is always a \emph{proxy}, and therefore achieving high performance on the ML task, such as detecting cancer, does not necessarily translate to desired outcomes, such as detecting cancer at earlier stages and reducing patient mortality.  

\ourpar{Design of ground truth.} Data is never truly raw or objective \cite{gitelman2013RawOxymoron, Feinberg2017, zajkac2023ground}. While we might hope for medical imaging data to be different, and only capture objective reality, this is not the case. Research shows that medical datasets are defined through the painstaking work of multidisciplinary teams within specific clinic, geographical, legal and socioeconomic contexts \cite{zajkac2023ground, Muller2021, miceli2022studying}.

The processes shaping medical imaging datasets begin before any data is collected \cite{zajkac2023ground}. In particular, regulatory constraints determine what data can be collected and predetermine its purpose. However, issues arise during data acquisition, as data collection is often poorly defined, resulting in variable quality. In contrast, clinical trials follow more regulated protocols. Conducting such structured pilot studies can help refine and standardize consistent acquisition protocols. The context of creation and use direct the design of the datasets and can be accounted for through purposeful investigation of local assumptions and meanings embedded in the dataset. Similar to computer vision datasets \cite{scheuerman2021datasets}, commercial and operational pressures hint at the fact that medical imaging datasets have their own politics and are created with a specific purpose, whether commercial or public. To this end, the development of a data management plan is increasingly required to streamline the data lifecycle process \cite{hutchinson2021towards}. 

\ourpar{Impact of labeling process on data quality} The consequences of labeling decisions are better understood than those of earlier stages and include, among others, the clinical relevance of ML models \cite{oakden2020hidden, oakden2022validation}, the proliferation of social inequality and exclusion \cite{leavy2021ethical}, and the impact on the performance of trained ML models \cite{chen2021data}. To address those challenges, we must look into the processes, guidelines, and incentives of labeling \cite{fort2016}, as well as the interpersonal and organizational structures of entities responsible for that work \cite{miceli2022studying, Muller2021}. For example, epistemic differences within multidisciplinary teams creating medical imaging datasets -- such as misunderstandings regarding clinical terminology like ``opacity'' in chest X-rays, with different meanings across countries or lacking direct translations -- can affect the clinical outcomes of developed models \cite{zajkac2023ground}. Overlooking design choices in medical datasets can lead to misalignment between ML systems and the real-world needs, (see also Section~\ref{sec:labelquality}). Recognizing these datasets as constructed \cite{raji2021ai}, not objective, is essential to their fair and equitable use in medical practice.

\ourpar{Trade-off between metadata and patient privacy.} Metadata such as demographics is essential for evaluating the robustness and fairness of ML models. However, the need for detailed metadata often clashes with the imperative to protect patient privacy, as even anonymized data can carry risks of re-identification and misuse. We do not cover this in detail in this work, but various methods for federated and privacy-preserving ML have been developed, for example \cite{rieke2020future,kaissis2021end}. 

\ourpar{Data size} 
 ML systems are often expected to perform better with more data \cite{kaplan2020scaling}, as has been both observed in practice, and shown  by statistical learning theory \cite{vapnik1999overview} wherein error tolerance, statistical dependency of the samples, data dimensionality, and model capacity all interact with the model performance. Datasets in medical imaging have grown from hundreds to thousands, with the largest public datasets like CheXpert \cite{irvin2019chexpert}, MIMIC-CXR \cite{johnson2019mimic}, and Emory breast \cite{jeong2023emory} with up to hundreds of thousands of patients. This contrasts with general computer vision datasets, which often contain millions of images and serve as the basis for many empirical findings. In industry, medical datasets are also scaling up to millions, offering advantages for model development.

With (smaller) public datasets, a common solution is to inject additional knowledge from another source into the dataset, such as domain knowledge provided by experts, or using data or representations from a different source. Transfer learning \cite{pan2010survey,cheplygina2019not} is therefore often used in medical imaging. A common approach is to fine-tune models pretrained on ImageNet \cite{russakovsky2015imagenet}, although recent results show this strategy is more sensitive to shortcuts \cite{juodelyte2024source} than if training on RadImageNet \cite{mei2022radimagenet}, a recent dataset with a million images from different radiological modalities. 

Given the various factors contributing to the interaction between the data and the learning performance, there are no guarantees that larger datasets will necessarily result in better target models. For example, merging datasets from different sources can lead to shortcuts and biases \cite{compton2023more,shen2024data}. In this regard, dataset distillation \cite{yu2023dataset} which aims to reduce the size of the data while maintaining or improving its representativeness, could lead to more robust algorithms, despite the ``bigger is better'' intuition.

\section{Data annotation practices and quality} \label{sec:labelquality}
\aptLtoX[graphic=no,type=html]{\begin{aptdispbox}\noindent\textbf{Relevance to the living review.} Linked documentation of the annotation process (annotation guidelines + annotations) reflects observer variability and  trade-offs between annotator expertise and cost.\end{aptdispbox}}{\noindent\fcolorbox{cyan}{cyan!10}{%
  \begin{minipage}{0.9\columnwidth}
\textbf{Relevance to the living review.} Linked documentation of the annotation process (annotation guidelines + annotations) reflects observer variability and  trade-offs between annotator expertise and cost.
  \end{minipage}
}
\vspace{0.1cm}}

Labeling tasks in medical imaging depend on the context of the disease, the task itself (classification, segmentation, etc) and how the target labels are acquired. For example, ground truth labels for lesions or nodules could be confirmed via biopsies, while other (gold standard) labels or annotations could be based on interpretation of the experts or other annotators. Here we discuss two important considerations which vary with the task: inter-observer variability and expertise-cost trade-offs. 

\ourpar{Inter-observer variability}
Establishing observer variability of the task at hand is crucial before data collection and annotation, as it helps set accuracy targets, determine the required number of annotators, estimate the needed data, and assess variability. Observer variability is measured by having multiple annotators curate a small set of representative examples and then calculating agreement for binary classification, or metrics like Dice score for segmentation  \cite{maier2022metrics}.

Notably, observer variability varies greatly according to the task, anatomical structures and their sizes, image type, and expertise level \cite{joskowicz2019inter}. For example, mostly healthy large organs, e.g., the lungs, the liver, and the brain, imaged on volumetric scans will have a low observer variability, while small structures, e.g., lung nodules, will show larger variability. 

Observer variability is also influenced by inconsistent training across institutions (e.g.,~\cite{balagopal2021psa}) and insufficient guidelines for annotations~\cite{radsch2023labelling}. Comprehensive labeling instructions~\cite{radsch2023labelling}, task-specific training~\cite{daniel2018quality}, and a structured feedback loop can help mitigate this, but are not always feasible if the annotation task is not part of the clinical workflow, and therefore might be more often used in industry datasets. 

\ourpar{Expertise and cost trade-offs}
Selecting the right annotators for a task is another critical consideration~\cite{radsch2025quality}, and given the growing demand for annotated data, careful thought must be given to matching the annotators’ skills to the task’s requirements, both in terms of accuracy and detail of the annotations needed, and domain-specific contexts, for example when diseases have different prevalence across countries. 

Domain expertise of annotators can vary from clinicians to laypersons. 
\emph{Expert}-annotated data can be collected from (retrospective) studies at hospitals, although some types of annotations needed for ML (such as granular annotations like image contours) may not be created as part of the clinical workflow. In such cases only weakly-labeled data might be available, and/or additional annotations might be created by other experts or graduate students for example. In industry, the commonly cited phrase ``Garbage in - garbage out'' prompts
companies to allocate significant resources to ensure high quality annotations and curation, possibly hiring domain-specific experts through their customer base. This can result in datasets with hundreds of subcategories rather than a limited set of high-level categories, more granular annotations, and annotation workflows with various quality control measures. 

Medical expertise is not always required for medically-related annotation tasks. For example, studies shown that laypersons are able to detect surgical instruments in laparoscopic images \cite{maier2014can}, and several other successful results with \emph{crowdsourcing} in medical imaging have been reported ~\cite{orting2020survey}, although often missing details about the annotation process. Additionally, annotations for training data typically do not require the same level of accuracy and detail as testing data used for evaluating the generalizability of the developed algorithms. This can allow using weakly-labeled training with scribbles \cite{lin2016scribblesup,liu2022weakly} or bounding boxes \cite{oh2021background,dai2015boxsup}, or in 3D data, leveraging sparse annotations only from a few slices within a volume ~\cite{philipp2024annotation}. Taken together, these strategies can allow employing novice annotators for annotating training data while reserving domain experts for curating the test set.

It is crucial to emphasize that despite the apparent advantages of crowdsourcing approaches, there is a lot of \emph{hidden data work}, i.e., the labor involved in data collection and annotation is often invisible and undervalued \cite{paullada2021data,sambasivan2021everyone}. There are various reports of unethical practices, for example by tech companies, with regard to data workers who might be dealing with disturbing images or language. The motivations and conditions of workers annotating medical images might be different, but crowdsourcing studies often do not provide this information \cite{orting2020survey}.

\ourpar{Expertise vs. cost trade-offs: (semi-)automatic approaches}
Given the cost of annotation and the growing need for large datasets, various automated methods for annotation have been proposed. Computer vision methods can be used to extract additional features from the image, for example asymmetry, border irregularity or Fitzpatrick skin type in skin lesions \cite{raumanns2021enhance,groh2021evaluating}, and used as additional labels, for example via multi-task learning. Natural language processing (NLP) techniques have been proposed to extract diagnostic labels from unstructured clinical reports. However, this strategy has been shown to introduce labeling errors, for example in chest X-rays \cite{oakden2020exploring, wang2017chestx}. Large language models (LLMs) have shown groundbreaking performance in the general language domain and are expected to unlock new possibilities for analyzing medical texts~\cite{ten2024chatgpt}. However, due to domain-specific challenges (language, contextual nuances, and the ambiguity inherent in medical terminology) their effectiveness is still limited, see for example studies for radiography or free-text CT or MR reports \cite{fink2023potential,mukherjee2023feasibility,le2024performance}. 

\emph{Hybrid} approaches such as active learning \cite{budd2021survey,gal2017deep}, active label correction \cite{bernhardt2022active} and hard sample mining (collecting additional ``hard samples'' from specific device manufacturers or rare disease subtypes) offer opportunities to combine the advantages of both human annotation and automated methods. While such methods are popular in literature, in academic papers the human annotators are sometimes simulated by giving the algorithm access to the existing labels in the data, while in industry, the (re)-labeling process is more often done with domain experts. This also allows industry datasets to be more dynamic than publicly available datasets. However, due to the proprietary nature, it is difficult to comment the cost vs. quality trade-offs, and what learnings public dataset creators can extract from this. 

\ourpar{Final remarks}
In general, it is crucial to remember that any annotation method, human or otherwise, will always be based on some assumptions. For example, methods which directly generate synthetic images and labels, are always based on some underlying data and will inherit its biases (and one could also question why the data needs to be generated explicitly if the data generating process is known). Without comprehensive documentation of the annotation process and its assumptions -- which might not always be explicit -- can lead to failures when the assumptions do not hold for the data at hand. Another crucial point is that, for example, reporting bias will inevitably affect what we learn about about different annotation methods. A living review where authors of new studies (which might not be noticed due to factors like the publication venue) contribute to evidence around a specific dataset, could help reduce these problems.

\section{Inside a dataset: demographics} \label{sec:demographics}
\begin{figure*}[]
    \centering
    \includegraphics[width=0.8\linewidth]{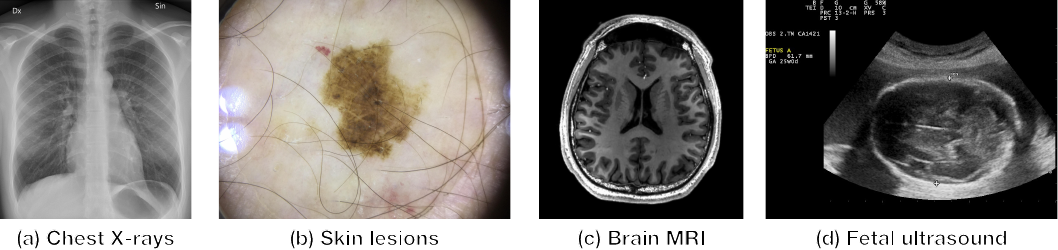}
    \caption{Case studies: (a) A normal posteroanterior chest radiograph of a patient with no visible signs of injury, showing letters that indicate left or right. (b) A malignant melanoma dermoscopic image from ISIC dataset, featuring dark corners and areas of intense brightness. (c) A cross-sectional T1-weighted MRI of a healthy human brain. (d) A fetal ultrasound image displaying the head of the fetus, where the text ``BDP'' refers to the head diameter, and ``GA'' indicates gestational age. The images (a), (c) and (d) are sourced from Wikimedia, and (b) from ISIC.}
    \label{fig:casestudies}
    \Description{Case studies: (a) A normal posteroanterior chest radiograph of a patient with no visible signs of injury, showing letters that indicate left or right. (b) A malignant melanoma dermoscopic image from ISIC dataset, featuring dark corners and areas of intense brightness. (c) A cross-sectional T1-weighted MRI of a healthy human brain. (d) A fetal ultrasound image displaying the head of the fetus showing information about the head diameter and gestational age.}
\end{figure*}

\aptLtoX[graphic=no,type=html]{\begin{aptdispbox}\noindent\textbf{Relevance to the living review.} We present four case studies (chest x-rays, skin lesions, brain MRI and fetal ultrasound), each with unique properties. We propose addressing these factors in our living review, emphasizing the importance of the documentation of the study design and patient demographics.\end{aptdispbox}}{\noindent\fcolorbox{cyan}{cyan!10}{%
  \begin{minipage}{0.9\columnwidth}
\textbf{Relevance to the living review.} We present four case studies (chest x-rays, skin lesions, brain MRI and fetal ultrasound), each with unique properties. We propose addressing these factors in our living review, emphasizing the importance of the documentation of the study design and patient demographics.
  \end{minipage}
}
\vspace{0.1cm}}

The potential of overfitting to established benchmark datasets is a long-standing debate in the ML community \cite{liao2021learning,gossmann2018test,hosseini2020tried}. While the past few years have seen a surge in public datasets, they typically lack demographic metadata about the subjects. This leads to a host of problems. If the data does not come with demographic information, researchers cannot assess whether the datasets and models trained on them have demographic biases. As a result, not only the resulting algorithms, but also what we learn about ML development, can be tainted by demographic biases without our knowledge. Equally important, research on algorithmic bias and fairness, which needs demographic metadata, has access to very few datasets. As a result, large amounts of research resources have been dedicated to a very small set of case studies, whose particularities -- for better and for worse -- drive the progress of the research field. Here we present four case studies, illustrated in Fig.~\ref{fig:casestudies}, further clinically relevant details in Section~\ref{sec:casestudies}. 

\ourpar{Case study 1: chest X-rays} Chest X-rays are the most commonly performed radiologic examinations worldwide~\cite{unscear2000sources}, requiring significant expertise for accurate and meaningful interpretation~\cite{van2001computer}. ML revolutionized chest X-ray diagnosis, highlighted by the release of NIH-CXR14 dataset~\cite{wang2017chestx} and CheXNet's model claiming radiologist-level pneumonia detection~\cite{rajpurkar2017chexnet}. However, these claims have been criticized for relying on shortcuts \cite{oakden2020hidden,jimenez2023detecting}, and for low inter-rater agreement \cite{damgaard2023augmenting}. 
Subsequently, additional datasets such as CheXpert~\cite{irvin2019chexpert}, MIMIC-CXR~\cite{johnson2019mimic}, and PadChest~\cite{bustos2020padchest} have become widely used in the research community. 

With the primary purpose of diagnosing and/or monitoring pathologies, chest X-ray datasets often include demographic information like age, gender, sex, race and ethnicity to analyze potential biases\footnote{We recognize that both gender and sex, as well as race and ethnicity, are distinct and not binary. However, datasets often do not document which variable was collected and/or use the terms interchangeably. We recognize that there are complexities in diagnosis when individuals belong to multiple categories. When possible, we use the terms used by the original authors throughout this paper.}. While age and gender are typically included, (self-reported) race or ethnicity are only available in a few datasets, such as MIMIC-CXR and CheXpert~\cite{paul2022demographic}. Age is generally skewed toward older populations (PadChest median age 62, MIMIC-CXR largest group aged 60–80~\cite{zhang2022improving}). Age can be predicted from chest X-rays~\cite{ieki2022deep}, which can lead to favoring well-represented age groups. Even without significant gender imbalance, performance disparities between genders persist \cite{larrazabal2020gender,seyyed2020chexclusion,seyyed2021underdiagnosis}. Balancing the data has proven ineffective, and studies have ruled out causes such as under-representation~\cite{larrazabal2020gender}, physiological differences~\cite{weng2023sex}, and shortcut learning~\cite{olesen2024slicing, oakden2020hidden, jimenez2023detecting}. Less is known about the effect of label errors, which have different effects on diagnostic labels -- in particular, the ``no finding'' label is known to often be associated with follow-up images of patients \cite{oakdenrayner2019xraysimpressions}. Performance disparities between racial groups favoring white individuals have been noted \cite{seyyed2020chexclusion}. ML can infer protected attributes like race, despite this being a challenging task for human experts~\cite{gichoya2022ai}. Recent research~\cite{glocker2023algorithmic} suggests that fine-tuning on specialized datasets alone cannot reduce the influence of these protected attributes.

\ourpar{Case study 2: skin lesions}
In recent years there has been significant growth in ML for dermatology~\cite{naqvi2023skin}, including tasks such as classification~\cite{pacheco2021attention}, segmentation~\cite{mirikharaji2023survey}, and lesion localization \cite{maqsood2023multiclass}. This surge may be largely explained by the availability of public datasets, such as Fitzpatrick17k~\cite{groh2021evaluating}, PAD-UFES-20~\cite{pacheco2020pad}, and HIBA~\cite{ricci2023dataset} -- as well as the International Skin Imaging Collaboration (ISIC) \cite{isicarchive}, which aggregates over 490K images from datasets such as HAM10000~\cite{tschandl2018ham10000} and BCN 20000 \cite{combalia2019bcn20000}. 

Despite the progress, the under-representation of demographic groups in these datasets limits the generalizability of ML~\cite{daneshjou2021lack}. Skin lesions manifest differently across populations, influenced by factors like skin tone, genetic background, age and UV light exposure \cite{wen2024data}. Nonetheless, most public datasets are limited in terms of geographic and skin tone diversity, with a predominance of lighter-skinned patients (Fitzpatrick I to III)~\cite{wen2022characteristics, groh2022towards}, potentially resulting in models that underperform for under-represented groups. For example, melanoma -- the deadliest type of skin cancer -- is much more prevalent in lighter-skinned individuals but can also affect those with darker skin tones, who might then be misdiagnosed \cite{groh2024deep}.

Recent efforts to diversify skin lesion datasets, such as the inclusion of PAD-UFES-20~\cite{pacheco2020pad} and HIBA~\cite{ricci2023dataset} in ISIC, have improved the representation of Latin American individuals, a region that was previously under-represented. However, there is still a strong lack of representation in terms of diversity of skin tone. Addressing these disparities is a global challenge that requires a collaborative effort from the research community, drawing on diverse perspectives and contributions from different backgrounds and regions worldwide. 

\ourpar{Case study 3: fetal ultrasound}
Ultrasound is the fundamental imaging modality for antenatal care. The acquisition process consists of a physical examination with an ultrasound probe looking for specific 2D slices called standard planes. Several sources of variation affect the image quality, such as the maternal body mass index (BMI) or the experience of the sonographer acquiring the scan. Another important factor is the ethnicity, which is associated with variations in the normal fetal growth according to several multi-ethnic studies~\citep{droogerEthnic2005,ogasawaraVariation2009,sletnerEthnic2015}. 

The few available datasets are the Fetal Planes DB~\citep{FetalPlanesDB} and data from the HC18~\citep{HC18}, FH-PS-AOP~\citep{AOP}, and ACOUSLIC-AI~\citep{ACOUSLIC} challenges, with none, to date, providing demographic information. More diverse datasets with detailed demographics, including BMI and ethnicity, image quality information, and expertise of the clinicians, are needed to develop fairer models and improve prenatal screening. This is true for low-, middle- and high-resource countries. In low and middle-income resource settings with poorer image quality due to portable devices this is crucial to reduce maternal complications and fetal mortality \cite{who2024maternalmortality}. A recent Danish study \cite{mikolaj2025} shows that across demographic subgroups, deep learning algorithms improve birth weight estimates from fetal ultrasound compared to clinical standard measurements extracted from those same ultrasound images -- potentially because the clinical standard measurements are unable to use additional image content to make up for suboptimal ultrasound planes. Future studies should therefore look with care at how both image quality and study design affect performance across groups -- for both ML and more traditional predictive approaches.

\ourpar{Case study 4: neuroimaging}
MRI is the third most commonly performed imaging modality after CT and X-rays. Its superior soft tissue contrast enables detailed visualization of brain anatomy, making it ideal for detecting abnormalities, and most public MRI datasets for ML research focus on the brain \cite{Dishner2024}. Key application areas for brain MRI and notable datasets, include neurodegenerative diseases (OASIS \cite{marcus2007oasis, marcus2010open, lamontagne2019oasis, koenig2020oasis}), brain cancer (BraTS \cite{Baid2021}, LUMIERE \cite{Suter2022}, Ocana \cite{ocana2023comprehensive}, and TCGA-GBM \cite{scarpace2016cancer}), and stroke, (ISLES 2022 \cite{hernandez2022isles}, ATLAS v2.0 \cite{Liew2022}, among others). 

The conversion from the standard clinical imaging format, DICOM, to NIfTI or other formats is complex and error-prone \cite{Li2016} and depends on how different formats implement DICOM standards. Moreover, many public neuroimaging datasets undergo extensive preprocessing prior to release, often to standarize datasets for open challenges, and ensure that model evaluations focus on algorithm performance rather than the effects of preprocessing. These factors may explain why the neuroimaging community --unlike other disciplines-- has made progress in advancing data-sharing practices by adopting standards and tools like BIDS \cite{Gorgolewski2016}, DataLad \cite{Halchenko2021}, NITRC.org \cite{kennedy2016nitrc} and OpenNeuro \cite{markiewicz2021openneuro}.

Demographic reporting in brain MRI is generally poor. For example, in US studies from 2010 to 2020, 77\% report sex, but only 10\% and 4\% report race and ethnicity, respectively~\cite{Sterling2022}. Studies show performance disparities in models by sex and race, with black females being most affected~\cite{dibaji2023studying,ioannou2022study}. Recent efforts to diversify brain MRI datasets include the addition of children (BraTS-PEDS~\cite{kazerooni2024bratspeds}) and Sub-Saharan African populations (BraTS-Africa~\cite{adewole2023brain}). Future work should focus on preserving raw data authenticity, enhancing diversity, and more complete metadata. 

\section{Inside a dataset: shortcuts} \label{sec:shortcuts}
\aptLtoX[graphic=no,type=html]{\begin{aptdispbox}\noindent\textbf{Relevance to the living review.}
We suggest documenting and annotating new evidence related to errors or shortcuts in existing datasets.\end{aptdispbox}}{\noindent\fcolorbox{cyan}{cyan!10}{%
  \begin{minipage}{0.9\columnwidth}
\textbf{Relevance to the living review.}
We suggest documenting and annotating new evidence related to errors or shortcuts in existing datasets.
  \end{minipage}
}
\vspace{0.1cm}}

Different terminology in the literature refers to shortcuts: confounders, spurious correlations, hidden stratification, etc. Shortcuts are decision rules that perform well on benchmark data but fail to transfer to challenging test cases, often with out-of-distribution data \cite{geirhos2018generalisation}.  When shortcuts fail, they result in biases and misdiagnosis, for example chest pain in women as anxiety or heart burn \cite{vanAssen2024implications}, misconception that Black patients have high pain threshold \cite{hoffman2016racial}, or delayed referrals for minority patients who may be judged as malingering or drug seeking \cite{haider2015unconscious}. Many shortcuts in medical imaging have been shown, often when the ML model memorizes irrelevant clinical characteristics, like the hospital where the patient was scanned \cite{compton2023more}. Hence, when the shortcut is missing, the model performance drops. For example, the model can rely on proxies, such as chest drains for pneumothorax \cite{oakden2020hidden,jimenez2023detecting}, radiographic markers containing scanning locations for pneumonia (especially in the intensive care unit, where the prevalence of pneumonia is high) \cite{zech2018variable}, dark corners or rulers for skin lesions \cite{bissoto2020debiasing}, or patient positioning for COVID-19 \cite{degrave2021ai}.

As we develop more novel datasets and dataset derivatives such as masks, embeddings and foundation models that extract features from multiple datasets, feature visualizations have demonstrated that the embeddings show clear separation of sex and risk groups, showing that these models also encode these characteristics \cite{glocker2023algorithmic} and lead to bias. It is important to note that these may be challenging to evaluate because of the complexity of ``model as a dataset''  where the original dataset may not be available for inspection, making it difficult to know whether the shortcut learning occurs due to the data, the bias of the model, or bias introduced by a fine-tuning dataset. It is therefore imperative to interpret the model outputs carefully and audit the false positives and false negatives using domain expertise.

\ourpar{Taxonomy of shortcuts}
Shortcuts can arise from spurious correlations (e.g. noise patterns or differences in intensity distributions across various scanners), demographic attributes, and their interactions. The Medical Imaging Contextualized Confounder Taxonomy (MICCAT) \cite{juodelyte2024source} (see Fig.~\ref{fig:miccat}) helps identify the shortcuts' origin and mitigation strategies. MICCAT extends beyond traditional demographic attributes to include a broader set of confounders that are domain- and context-specific confounders at both patient and environment levels.

Patient-level confounders include demographic attributes (sex/gender \cite{larrazabal2020gender,abbasi2020risk}, age \cite{abbasi2020risk}, and race/ethnicity \cite{gichoya2022ai}) and anatomical factors related to organs or conditions (BMI, tissue/breast/bone density). While demographic factors are standard in bias analysis, anatomical variations may form subgroups where models underperform, and their identification often requires analysis beyond standard demographic characteristics \cite{voneuler-chelpin2019sensitivity}.

Environment-level confounders include external and imaging confounders. External confounders involve visible elements like chest drains \cite{oakden2020hidden,jimenez2023detecting}, pen marks \cite{winkler2019association}, patient positioning \cite{degrave2021ai}, or text overlays \cite{lin2024shortcut}, creating localized artifacts.  In contrast, imaging confounders result from the imaging process (acquisition devices or parameters, noise, or motion artifacts), leading to global artifacts that may be imperceptible to the human eye. Systematic variations in exposure setting for chest X-rays \cite{lang2024using} or different characteristics of imaging equipment across centers \cite{compton2023more} can introduce shortcuts, causing models to rely on acquisition-specific factors rather than genuine clinical factors.

Documenting what type of shortcut (patient or environment, external or internal) a dataset has in our living review would enable researchers to learn across applications. For example, researchers working on skin lesions could identify and learn from relevant studies and experiments from ophthalmology, even if these studies would not be findable with traditional search strategies.

 \section{Data lifecycle} \label{sec:lifecycle}
\aptLtoX[graphic=no,type=html]{\begin{aptdispbox}\noindent\textbf{Relevance to the living review.} Our framework incentivizes researchers to adopt better dataset management practices. We recommend using persistent identifiers and storing, and versioning of datasets for reproducibility, and licensing and proper tracking for author attribution.\end{aptdispbox}}{\noindent\fcolorbox{cyan}{cyan!10}{%
  \begin{minipage}{0.9\columnwidth}
\textbf{Relevance to the living review.} Our framework incentivizes researchers to adopt better dataset management practices. We recommend using persistent identifiers and storing, and versioning of datasets for reproducibility, and licensing and proper tracking for author attribution.
  \end{minipage}
}
\vspace{0.1cm}}

Effective research data management is crucial for reproducibility, re-usability, and efficiency. The data lifecycle consists not only of data acquisition and analysis, but rather of data acquisition (see Section~\ref{sec:creatingdata}), data organization and standardization, data and metadata annotation (see Section~\ref{sec:labelquality}), data management and tracking during analysis, and ultimately, data maintenance. A comprehensive overview, with an example from neuroimaging, is provided in \cite{niso2022}.

\paragraph{\textbf{Data standards}} 
Since research, and especially medical research data, is often collected for one purpose, but then reused for another, data standardization is crucial to enhance re-usability. Converting to established data standards directly after acquisition makes it easier to ensure transparency and provenance of the data. Part of standardizing the data is also often to add metadata and perform additional annotations. Here proper documentation of where/how the metadata and annotations were acquired are important to ensure the usefulness of the additional data. 

Data standardization is not always easily achieved and very application dependent (see also Section~\ref{sec:demographics}). One example, the Brain Imaging Data Structure (BIDS) \cite{poldrack2024past} exemplifies the role of data standards in neuroimaging research. By providing a consistent framework for organizing and describing datasets, BIDS has enabled large scale data sharing, reproducibility, and collaboration across the scientific neuroimaging community, facilitating development of interoperable tools and workflows \cite{gorgolewski2017bids}, streamlining data analysis and reducing errors. This serves as an example for developing data standards for other medical imaging applications.

\paragraph{\textbf{Sharing data via data repositories}} 
Sharing data with the community is crucial for ensuring reproducibility, maximizing the scientific insights derived from the same set of participants, empowering researchers to reuse data, develop innovative analytical methods, refine scientific hypotheses, and conduct large-scale meta-analyses. Hosting services play a crucial role in advancing research and the development of ML models. Examples include repositories where researchers can access and share data (PhysioNet~\cite{goldberger2000physiobank}, Zenodo~\cite{zenodo2013}, OpenNeuro~\cite{markiewicz2021openneuro}) and host benchmarking competitions (Grand Challenge~\cite{grandchallenge}, Kaggle~\cite{kaggle}, HuggingFace~\cite{huggingFace}). Several of these are referred to as Community Contributed Platforms (CCPs).

Whenever a dataset is updated or changed, it should be versioned and the changes documented. While this functionality is available on some open data sharing platforms, it is not widely adopted. For example, HuggingFace provides a way to track the history of files and versions on their website, but does not explicitly version their datasets. Moreover, the issue of dataset version tracking is even more critical with copies of datasets on CCPs, where documentation, including pre-processing steps, is frequently lacking~\cite{jimenez2024copycats}.

Finally, open source tools for decentralized data sharing and processing, such as DataLad \cite{halchenko2021datalad}, facilitate data management and analysis tracking, and cost less than industry-standard solutions. Data maintenance costs can be optimized by considering dataset use. In neuroimaging for example, only a small percentage (<10\%) of datasets in large repositories are accessed after two years \cite{pernet2023long}, thus rarely accessed data could be archived, i.e., stored in ``cold'' / low-cost storage, reducing the data maintenance costs. 

\ourpar{Licensing} 
A concern for medical imaging datasets distributed on open repositories is their appropriateness for reuse. This has been partly addressed by the FAIR (Findable, Accessible, Interoperable, Reusable) principles, a data stewardship model for scientific data management and governance \cite{wilkinson2016fair}. Nevertheless, compliance with FAIR is a complex process that requires a series of activities which are dependent on expertise and oversight of the researchers distributing the datasets, e.g., metadata release and licensing \cite{jimenez2024copycats, yang2024navigating}.

A license is a standardized statement that specifies the permissible uses of the data and the associated constraints for the end user, with the most prevalent being the Creative Commons (CC) licensing suite. CC licenses allow specifying requirements for how to cite their attribution, as well as the conditions for the dataset reuse and potential re-sharing options for derivatives. Yet, there is still a lack of appropriate licenses for popular datasets \cite{jimenez2024copycats, longpre2023dataprovenance}. Possible reasons include the legal expertise of the dataset creators, the lack of auditing from the CCPs once the datasets are distributed, and the complexity of data protection laws, which often lead to confusion on how to license anonymized datasets which have already been approved in an initial release phase. 

It is important to emphasize that adhering to the FAIR principles does not require fully releasing the data under open licenses -- the data should be ``as open as possible, as closed as necessary''. Sensitive data can be shared under a Data Use Agreement (DUA), provided clear and precise guidelines are established for compliance with the DUA. As a step forward, conferences could offer workshops on data protection and licensing.

\ourpar{Tracking of dataset use}
Tracking the use of datasets throughout publications and experiments is important for several reasons: giving credit to dataset providers and annotators, identifying changes to datasets due to errors or ethical concerns \cite{peng2021mitigating}, and ensuring fair comparisons of new methods. Without proper tracking, it becomes extremely difficult to identify which methods are impacted by dataset-related issues.

Sources like PubMed, open citation tools like OpenAlex~\cite{priem2022openalex} or Semantic scholar~\cite{fricke2018semantic}, and platforms like Papers with Code help track dataset use through citations and benchmark results. However, citation practices often fail to fully capture dataset use, as datasets are sometimes cited via URLs or footnotes~\cite{heller2019role,sourget2024citation}. Some authors neglect citation altogether, and some datasets lack clear citation instructions~\cite{jimenez2024copycats}.  Journals and conferences can improve citation tracking by requiring a ``Data availability'' section, a practice already adopted by some. Finally, current citation practices lack sufficient context to confirm that a dataset is actually used in the experiment, and not just cited as an example. Modifying the citation graph to be more fine-grained, as proposed in~\cite{buneman2021data} and in our living review, would provide more context about dataset use. 

\section{Discussion and conclusions} \label{sec:discussion} 
We have comprehensively explored various factors that shape a medical imaging dataset, from translating the clinical problem into an ML problem to key considerations for creating high-quality datasets. These considerations include defining the ground truth, determining the necessary amount of data, and detailing the annotation process. We also examined the contents of datasets throughout four case studies, highlighting context-specific clinical factors, and describing a taxonomy of shortcuts. Furthermore, we discussed scientific data management and stewardship for the data lifecycle.

Our living review addresses these challenges by proposing a structure that encourages collaboration, better dataset documentation and citation practices, and regular updates on new evidence. Through a living review process, we facilitate reflection on decisions made throughout the dataset creation process, ensuring thorough documentation of annotation guidelines and annotations. The review promotes collaboration to extend open datasets through the addition of new annotations, and emphasizes the importance of including the study design and patient demographics. Our proposal considers persistent identifiers, dataset versioning, licensing, and ensuring proper citation standards, supporting researchers in adopting best practices for dataset management and accessibility. 

We acknowledge several limitations of our proposed living review, including covering all areas of medical imaging, the effort for ongoing maintenance, and the challenge of ensuring awareness among all researchers. While we emphasize the importance of patient metadata, like demographics (see Section~\ref{sec:demographics}), we do not explore data privacy in detail. 

\ourpar{Learning from other communities}
A medical imaging dataset is often a proxy for the actual healthcare problem and can lead to inaccurate diagnoses, but this is often not documented in the datasets or trained models, necessitating a living review of datasets as we propose. It is also clear that medical imaging can benefit significantly from other communities, for example clinical prediction research and epidemiology.

Data sharing in clinical prediction research can be improved. A recent review in ML for oncology reported that only 2 out of 46 studies shared their data \cite{collins2024open}, and data availability statements (reported in 76\% of cases) often stated that data was available on request. However, this practice is frequently done merely to meet editorial requirements, rather than to provide easy access to data \cite{savage2009empirical,rowhani2016has}, despite the introduction of the FAIR principles (see Section \ref{sec:lifecycle}). 

Algorithmic fairness was recently incorporated in the updated TRIPOD+AI statement, which provides guidance for reporting clinical prediction studies \cite{Collins2024}. Researchers publishing clinical prediction models are specifically asked to report any methods used to assess and address model fairness. The reporting of open science practices (e.g., accessible study protocols, study registration, data and code sharing) is also included as a new focus in the updated statement.

Systematic reviews of clinical prediction studies often use the PROBAST checklist to assess bias risks and model applicability \cite{moons2019probast}. One fundamental point relevant to medical imaging is about appropriate inclusion and exclusion of participants. For example, including images of participants who already progressed to a severe disease state and would never be assessed with ML, might reduce generalizability of the model. This could explain why clinical prediction models are abundant in the literature, but much rarer in the healthcare sector \cite{van2022developing}. 

Another question in PROBAST is whether predictors/images were similarly assessed for all participants -- if not, shortcuts (see Section~\ref{sec:shortcuts}) can occur. Shortcuts are similar to confounding in epidemiology, where, for example, sex is associated with both access to treatment and disease prevalence. Developing a simple benchmark including only metadata could provide useful information on the magnitude of the images’ additional predictive value. Confounding can also be dealt with stratification, but this can lead to smaller and smaller datasets, which is an issue given the already low sample sizes in prediction research \cite{dhiman2023sample}. While a list of potential shortcuts in the dataset documentation is helpful, it cannot replace domain expertise. However, in a survey of ML for (non-image) medical data, 35\% of the studies did not have any authors with a medical affiliation \cite{ebbehoj2022transfer}, and one could speculate that it is even lower at ML conferences. 

More generally in healthcare, the STANDING (STANdards for data Diversity, INclusivity and Generalizability) Together initiative \cite{ganapathi2022tackling}, involving many clinicians, aims to reduce ML health inequalities by providing recommendations to increase transparency and generalizability of datasets. Consensus recommendations were developed in two parts by an international multidisciplinary team \cite{alderman2025tackling}: (1) guidance on how dataset creators should document their datasets, emphasizing transparency about the dataset's origin, composition, limitations, and potential biases, and (2) how data users can best utilize these datasets to minimize potential harm and ensure equitable performance across different population groups.

To conclude, the value of interdisciplinary collaborations and sharing best research practices across disciplines is crucial to develop clinically relevant applications that high quality datasets provide the basis for.

\ourpar{Future outlook}
One of the core problems surrounding datasets are the current incentives that are rewarded and optimized for in ML. While we have a proposal for how to incentivize and maintain our living review, it will not fix all the problems since they require changes across many research fields. 

Although more data-centric initiatives, such as the Datasets and Benchmarks Track at NeurIPS, are emerging, it is clear that more attention to data work is needed. Perhaps in this context it is worth mentioning that despite the importance of the topic and a diverse group of researchers, our own workshop proposal was initially rejected by two different venues. However, given recent progress, we remain hopeful data work will be more addressed and better acknowledged in the future, and we are in the process of applying for funding for related research and networking events, to ensure the continuity of this work. 

Looking ahead, established peer-review processes in conferences could be adapted to support the publication and recognition of data artifacts, assigning them DOIs similar to standard papers. Conferences like Medical Imaging with Deep Learning (MIDL)\footnote{https://2025.midl.io/} could invite data artifacts as papers for a special track, similar to the Datasets and Benchmarks track at NeurIPS. However, a more scalable approach would involve multiple conferences adopting a ``rolling'' review style, as has been adopted in the natural language processing community \cite{aclrollingreview}. Another interesting option would be for conferences to invite collaborative contributions to selected datasets, similar to medical imaging competitions like those hosted at the International Conference on Medical Image Computing and Computer Assisted Interventions (MICCAI)\footnote{https://conferences.miccai.org/2025/en/challenges.asp}. In the existing scenario, the organizers select a dataset, competition participants develop algorithms, and after benchmarking, both organizers and many participants typically co-author a paper about the results. Contributions to a datasheet would, of course, not be trivial to ``benchmark''; however, there would not need to be a ``winner-takes-all'' mentality. If different teams highlight different aspects of the datasets, their contributions could simply be combined.

With our work, we want to promote a cultural shift toward the responsible use of high quality datasets, created through collaboration with relevant stakeholders or clinical experts. We therefore invite everyone involved in and affected by healthcare ML to contribute to our living review and other dataset efforts.

\section*{Ethical considerations statement}
We propose a living review framework to connect publicly available medical imaging datasets with associated research artifacts for the research community. Our proposal is based on publicly available data, and no additional private or sensitive data were collected. Researchers who want to contribute with new evidence to update our database should make sure that their annotation or documentation metadata is compliant with the General Data Protection Regulation (GDPR) in the European Union (EU), the EU AI Act, and other relevant national and international legislation governing data privacy. We briefly discuss considerations about the trade-off between metadata and patient privacy in Section~\ref{sec:creatingdata}, and considerations about FAIR principles and licensing in Section \ref{sec:lifecycle}. 

\section*{Adverse impact statement}
The goal of our living review framework is to benefit the research community by improving the usability and accountability of publicly available medical imaging datasets. By providing insights into annotations, errors, and additional findings, our framework helps reduce the risk of misinterpretation or reliance on flawed data. However, without proper stewardship, moderation and ongoing maintenance, there is a risk that insufficiently validated datasets or artifact could be included. It is important to note that our proposal is a data/literature exploration tool for researchers, and should not directly be used for clinical decisions. 
\section*{Author Contributions}
Authors, except for the first and last, are listed in alphabetical order. Contributions follow the CRediT author statement.

\noindent \textbf{Conceptualization, Methodology, Writing - Original Draft \& Final Draft}: Veronika Cheplygina, Amelia \JimSan.

\noindent \textbf{Data Curation, Visualization, Software}: Amelia \JimSan.

\noindent \textbf{Supervision, Funding Acquisition}: Veronika Cheplygina.

\noindent \textbf{Writing - Review \& Editing}: all authors contributed, with individual section contributions listed below in alphabetical order:
\begin{itemize}
    \item \textit{Introduction}: Veronika Cheplygina, Amelia \JimSan, Maria A. Zuluaga
    \item \textit{Proposal: a living review of medical imaging datasets}: Veronika Cheplygina, Amelia \JimSan
    \item \textit{Considerations for creating a dataset}: 
   Veronika Cheplygina, Camila González, Steff Groefsema, Amelia \JimSan, Leo Joskowicz, Melih Kandemir, Hubert Dariusz Zając.
    \item \textit{Data annotation practices and quality}: Veronika Cheplygina, Alessa Hering, Leo Joskowicz, Thijs Kooi, Tim Rädsch.
    \item \textit{Inside a dataset: demographics}: Víctor M. Campello, Aasa Feragen, Jorge del Pozo Lérida, Andre Pacheco, Mauricio Reyes, Nina Weng, Jack Junchi Xu.
    \item \textit{Inside a dataset: shortcuts}: Enzo Ferrante, Judy Wawira Gichoya, Dovile Juodelyte.
    \item \textit{Data lifecycle}: Natalia-Rozalia Avlona, Sarah de Boer, Melanie Ganz-Benjaminsen, Bram van Ginneken, Amelia \JimSan, Théo Sourget.
    \item \textit{Discussion}: Veronika Cheplygina, Adam Hulman, Amelia \JimSan, Livie Yumeng Li, David Wen.
\end{itemize}


\begin{acks}
    This project has received funding from the Independent Research Council Denmark (DFF) Inge Lehmann 1134-00017B. The ``In the Picture: Medical Imaging Datasets'' workshop was funded by DFF and the Danish Data Science Academy (DDSA) with Grant ID 2024-2342. We extend our gratitude to the speakers and participants of the ``Datasets through the Looking-Glass'' webinar, who have helped to shape this research. EF gratefully acknowledges the support of the Google Award for Inclusion Research (AIR) Program. AJS was financed by the DFF grant MMC. AH is employed at Steno Diabetes Center Aarhus that is partly funded by a donation from the Novo Nordisk Foundation. AH is supported by a Data Science Emerging Investigator grant by the Novo Nordisk Foundation (NNF22OC0076725). 
TR was supported by a scholarship from the Hanns Seidel Foundation with funds from the Federal Ministry of Education and Research Germany (BMBF). 
DW received funding from the Jill and Herbert Hunt Scholarship, University of Oxford. 
MAZ is funded by TRAIN (ANR-22-FAI1-0003-02).
We thank Freepick for the icons in Fig.~\ref{fig:livingreviewzenododb}.


\end{acks}

\bibliographystyle{ACM-Reference-Format}

\bibliography{refs_public,refs_amelia,refs_veronika}  


\begin{thebibliography}{200}


\ifx \showCODEN    \undefined \def \showCODEN     #1{\unskip}     \fi
\ifx \showDOI      \undefined \def \showDOI       #1{#1}\fi
\ifx \showISBNx    \undefined \def \showISBNx     #1{\unskip}     \fi
\ifx \showISBNxiii \undefined \def \showISBNxiii  #1{\unskip}     \fi
\ifx \showISSN     \undefined \def \showISSN      #1{\unskip}     \fi
\ifx \showLCCN     \undefined \def \showLCCN      #1{\unskip}     \fi
\ifx \shownote     \undefined \def \shownote      #1{#1}          \fi
\ifx \showarticletitle \undefined \def \showarticletitle #1{#1}   \fi
\ifx \showURL      \undefined \def \showURL       {\relax}        \fi
\providecommand\bibfield[2]{#2}
\providecommand\bibinfo[2]{#2}
\providecommand\natexlab[1]{#1}
\providecommand\showeprint[2][]{arXiv:#2}

\bibitem[Abbasi-Sureshjani et~al\mbox{.}(2020)]%
        {abbasi2020risk}
\bibfield{author}{\bibinfo{person}{Samaneh Abbasi-Sureshjani},
  \bibinfo{person}{Ralf Raumanns}, \bibinfo{person}{Britt~EJ Michels},
  \bibinfo{person}{Gerard Schouten}, {and} \bibinfo{person}{Veronika
  Cheplygina}.} \bibinfo{year}{2020}\natexlab{}.
\newblock \showarticletitle{Risk of Training Diagnostic Algorithms on Data with
  Demographic Bias}. In \bibinfo{booktitle}{\emph{MICCAI LABELS workshop,
  Lecture Notes in Computer Science}}, Vol.~\bibinfo{volume}{12446}.
  \bibinfo{publisher}{Springer}, \bibinfo{pages}{183--192}.
\newblock


\bibitem[Adewole et~al\mbox{.}(2023)]%
        {adewole2023brain}
\bibfield{author}{\bibinfo{person}{M Adewole}, \bibinfo{person}{JD Rudie},
  \bibinfo{person}{A Gbadamosi}, {et~al\mbox{.}}}
  \bibinfo{year}{2023}\natexlab{}.
\newblock \showarticletitle{The Brain Tumor Segmentation (BraTS) Challenge
  2023: Glioma Segmentation in Sub-Saharan Africa Patient Population
  (BraTS-Africa)}.
\newblock \bibinfo{journal}{\emph{arXiV preprint arXiV:2305.19369}}
  (\bibinfo{year}{2023}).
\newblock


\bibitem[Akhtar et~al\mbox{.}(2024)]%
        {akhtar2024croissant}
\bibfield{author}{\bibinfo{person}{Mubashara Akhtar}, \bibinfo{person}{Omar
  Benjelloun}, \bibinfo{person}{Costanza Conforti}, \bibinfo{person}{Joan
  Giner-Miguelez}, \bibinfo{person}{Nitisha Jain}, \bibinfo{person}{Michael
  Kuchnik}, \bibinfo{person}{Quentin Lhoest}, \bibinfo{person}{Pierre
  Marcenac}, \bibinfo{person}{Manil Maskey}, {et~al\mbox{.}}}
  \bibinfo{year}{2024}\natexlab{}.
\newblock \showarticletitle{{Croissant: A Metadata Format for ML-Ready
  Datasets}}.
\newblock \bibinfo{journal}{\emph{arXiv preprint arXiv:2403.19546}}
  (\bibinfo{year}{2024}).
\newblock


\bibitem[Alderman et~al\mbox{.}(2025)]%
        {alderman2025tackling}
\bibfield{author}{\bibinfo{person}{Joseph~E Alderman}, \bibinfo{person}{Joanne
  Palmer}, \bibinfo{person}{Elinor Laws}, \bibinfo{person}{Melissa~D
  McCradden}, \bibinfo{person}{Johan Ordish}, \bibinfo{person}{Marzyeh
  Ghassemi}, \bibinfo{person}{Stephen~R Pfohl}, \bibinfo{person}{Negar
  Rostamzadeh}, \bibinfo{person}{Heather Cole-Lewis}, \bibinfo{person}{Ben
  Glocker}, {et~al\mbox{.}}} \bibinfo{year}{2025}\natexlab{}.
\newblock \showarticletitle{Tackling algorithmic bias and promoting
  transparency in health datasets: the STANDING Together consensus
  recommendations}.
\newblock \bibinfo{journal}{\emph{The Lancet Digital Health}}
  \bibinfo{volume}{7}, \bibinfo{number}{1} (\bibinfo{year}{2025}),
  \bibinfo{pages}{e64--e88}.
\newblock


\bibitem[Baid et~al\mbox{.}(2021)]%
        {Baid2021}
\bibfield{author}{\bibinfo{person}{Ujjwal Baid}, \bibinfo{person}{Satyam
  Ghodasara}, \bibinfo{person}{Suyash Mohan}, \bibinfo{person}{Michel Bilello},
  \bibinfo{person}{Evan Calabrese}, \bibinfo{person}{Errol Colak},
  \bibinfo{person}{Keyvan Farahani}, \bibinfo{person}{Jayashree
  Kalpathy-Cramer}, \bibinfo{person}{Felipe~C. Kitamura},
  \bibinfo{person}{Sarthak Pati}, \bibinfo{person}{Luciano~M. Prevedello},
  \bibinfo{person}{Jeffrey~D. Rudie}, \bibinfo{person}{Chiharu Sako},
  \bibinfo{person}{Russell~T. Shinohara}, \bibinfo{person}{Timothy Bergquist},
  \bibinfo{person}{Rong Chai}, \bibinfo{person}{James Eddy},
  \bibinfo{person}{Julia Elliott}, \bibinfo{person}{Walter Reade},
  \bibinfo{person}{Thomas Schaffter}, \bibinfo{person}{Thomas Yu},
  \bibinfo{person}{Jiaxin Zheng}, \bibinfo{person}{Ahmed~W. Moawad},
  \bibinfo{person}{Luiz~Otavio Coelho}, \bibinfo{person}{Olivia McDonnell},
  \bibinfo{person}{Elka Miller}, \bibinfo{person}{Fanny~E. Moron},
  \bibinfo{person}{Mark~C. Oswood}, \bibinfo{person}{Robert~Y. Shih},
  \bibinfo{person}{Loizos Siakallis}, \bibinfo{person}{Yulia Bronstein},
  \bibinfo{person}{James~R. Mason}, \bibinfo{person}{Anthony~F. Miller},
  \bibinfo{person}{Gagandeep Choudhary}, \bibinfo{person}{Aanchal Agarwal},
  \bibinfo{person}{Cristina~H. Besada}, \bibinfo{person}{Jamal~J. Derakhshan},
  \bibinfo{person}{Mariana~C. Diogo}, \bibinfo{person}{Daniel~D. Do-Dai},
  \bibinfo{person}{Luciano Farage}, \bibinfo{person}{John~L. Go},
  \bibinfo{person}{Mohiuddin Hadi}, \bibinfo{person}{Virginia~B. Hill},
  \bibinfo{person}{Michael Iv}, \bibinfo{person}{David Joyner},
  \bibinfo{person}{Christie Lincoln}, \bibinfo{person}{Eyal Lotan},
  \bibinfo{person}{Asako Miyakoshi}, \bibinfo{person}{Mariana Sanchez-Montano},
  \bibinfo{person}{Jaya Nath}, \bibinfo{person}{Xuan~V. Nguyen},
  \bibinfo{person}{Manal Nicolas-Jilwan}, \bibinfo{person}{Johanna~Ortiz
  Jimenez}, \bibinfo{person}{Kerem Ozturk}, \bibinfo{person}{Bojan~D.
  Petrovic}, \bibinfo{person}{Chintan Shah}, \bibinfo{person}{Lubdha~M. Shah},
  \bibinfo{person}{Manas Sharma}, \bibinfo{person}{Onur Simsek},
  \bibinfo{person}{Achint~K. Singh}, \bibinfo{person}{Salil Soman},
  \bibinfo{person}{Volodymyr Statsevych}, \bibinfo{person}{Brent~D. Weinberg},
  \bibinfo{person}{Robert~J. Young}, \bibinfo{person}{Ichiro Ikuta},
  \bibinfo{person}{Amit~K. Agarwal}, \bibinfo{person}{Sword~C. Cambron},
  \bibinfo{person}{Richard Silbergleit}, \bibinfo{person}{Alexandru Dusoi},
  \bibinfo{person}{Alida~A. Postma}, \bibinfo{person}{Laurent
  Letourneau-Guillon}, \bibinfo{person}{Gloria J.~Guzman Perez-Carrillo},
  \bibinfo{person}{Atin Saha}, \bibinfo{person}{Neetu Soni},
  \bibinfo{person}{Greg Zaharchuk}, \bibinfo{person}{Vahe~M. Zohrabian},
  \bibinfo{person}{Yingming Chen}, \bibinfo{person}{Milos~M. Cekic},
  \bibinfo{person}{Akm Rahman}, \bibinfo{person}{Juan~E. Small},
  \bibinfo{person}{Varun Sethi}, \bibinfo{person}{Christos Davatzikos},
  \bibinfo{person}{John Mongan}, \bibinfo{person}{Christopher Hess},
  \bibinfo{person}{Soonmee Cha}, \bibinfo{person}{Javier Villanueva-Meyer},
  \bibinfo{person}{John~B. Freymann}, \bibinfo{person}{Justin~S. Kirby},
  \bibinfo{person}{Benedikt Wiestler}, \bibinfo{person}{Priscila Crivellaro},
  \bibinfo{person}{Rivka~R. Colen}, \bibinfo{person}{Aikaterini Kotrotsou},
  \bibinfo{person}{Daniel Marcus}, \bibinfo{person}{Mikhail Milchenko},
  \bibinfo{person}{Arash Nazeri}, \bibinfo{person}{Hassan Fathallah-Shaykh},
  \bibinfo{person}{Roland Wiest}, \bibinfo{person}{Andras Jakab},
  \bibinfo{person}{Marc-Andre Weber}, \bibinfo{person}{Abhishek Mahajan},
  \bibinfo{person}{Bjoern Menze}, \bibinfo{person}{Adam~E. Flanders}, {and}
  \bibinfo{person}{Spyridon Bakas}.} \bibinfo{year}{2021}\natexlab{}.
\newblock \showarticletitle{The RSNA-ASNR-MICCAI BraTS 2021 Benchmark on Brain
  Tumor Segmentation and Radiogenomic Classification}.
\newblock \bibinfo{journal}{\emph{arXiv preprint arXiv:2107.02314}}
  (\bibinfo{year}{2021}).
\newblock
\urldef\tempurl%
\url{http://arxiv.org/abs/2107.02314}
\showURL{%
\tempurl}


\bibitem[Balagopal et~al\mbox{.}(2021)]%
        {balagopal2021psa}
\bibfield{author}{\bibinfo{person}{Anjali Balagopal}, \bibinfo{person}{Howard
  Morgan}, \bibinfo{person}{Michael Dohopolski}, \bibinfo{person}{Ramsey
  Timmerman}, \bibinfo{person}{Jie Shan}, \bibinfo{person}{Daniel~F Heitjan},
  \bibinfo{person}{Wei Liu}, \bibinfo{person}{Dan Nguyen},
  \bibinfo{person}{Raquibul Hannan}, \bibinfo{person}{Aurelie Garant},
  {et~al\mbox{.}}} \bibinfo{year}{2021}\natexlab{}.
\newblock \showarticletitle{Psa-net: Deep learning--based physician
  style--aware segmentation network for postoperative prostate cancer clinical
  target volumes}.
\newblock \bibinfo{journal}{\emph{Artificial Intelligence in Medicine}}
  \bibinfo{volume}{121} (\bibinfo{year}{2021}), \bibinfo{pages}{102195}.
\newblock


\bibitem[Banerjee et~al\mbox{.}(2023)]%
        {banerjee2023shortcuts}
\bibfield{author}{\bibinfo{person}{Imon Banerjee}, \bibinfo{person}{Kamanasish
  Bhattacharjee}, \bibinfo{person}{John~L. Burns}, \bibinfo{person}{Hari
  Trivedi}, \bibinfo{person}{Saptarshi Purkayastha}, \bibinfo{person}{Laleh
  Seyyed-Kalantari}, \bibinfo{person}{Bhavik~N. Patel}, \bibinfo{person}{Rakesh
  Shiradkar}, {and} \bibinfo{person}{Judy Gichoya}.}
  \bibinfo{year}{2023}\natexlab{}.
\newblock \showarticletitle{“{Shortcuts}” {Causing} {Bias} in {Radiology}
  {Artificial} {Intelligence}: {Causes}, {Evaluation}, and {Mitigation}}.
\newblock \bibinfo{journal}{\emph{Journal of the American College of
  Radiology}} \bibinfo{volume}{20}, \bibinfo{number}{9} (\bibinfo{date}{Sept.}
  \bibinfo{year}{2023}), \bibinfo{pages}{842--851}.
\newblock
\showISSN{1546-1440}
\urldef\tempurl%
\url{https://doi.org/10.1016/j.jacr.2023.06.025}
\showDOI{\tempurl}


\bibitem[Bernhardt et~al\mbox{.}(2022)]%
        {bernhardt2022active}
\bibfield{author}{\bibinfo{person}{M{\'e}lanie Bernhardt},
  \bibinfo{person}{Daniel~C Castro}, \bibinfo{person}{Ryutaro Tanno},
  \bibinfo{person}{Anton Schwaighofer}, \bibinfo{person}{Kerem~C Tezcan},
  \bibinfo{person}{Miguel Monteiro}, \bibinfo{person}{Shruthi Bannur},
  \bibinfo{person}{Matthew~P Lungren}, \bibinfo{person}{Aditya Nori},
  \bibinfo{person}{Ben Glocker}, {et~al\mbox{.}}}
  \bibinfo{year}{2022}\natexlab{}.
\newblock \showarticletitle{Active label cleaning for improved dataset quality
  under resource constraints}.
\newblock \bibinfo{journal}{\emph{Nature Communications}} \bibinfo{volume}{13},
  \bibinfo{number}{1} (\bibinfo{year}{2022}), \bibinfo{pages}{1161}.
\newblock


\bibitem[Birhane et~al\mbox{.}(2022)]%
        {birhane2021values}
\bibfield{author}{\bibinfo{person}{Abeba Birhane}, \bibinfo{person}{Pratyusha
  Kalluri}, \bibinfo{person}{Dallas Card}, \bibinfo{person}{William Agnew},
  \bibinfo{person}{Ravit Dotan}, {and} \bibinfo{person}{Michelle Bao}.}
  \bibinfo{year}{2022}\natexlab{}.
\newblock \showarticletitle{The values encoded in machine learning research}.
  In \bibinfo{booktitle}{\emph{ACM Conference on Fairness, Accountability, and
  Transparency (FAccT)}}. \bibinfo{publisher}{ACM}.
\newblock


\bibitem[Bissoto et~al\mbox{.}(2020)]%
        {bissoto2020debiasing}
\bibfield{author}{\bibinfo{person}{Alceu Bissoto}, \bibinfo{person}{Eduardo
  Valle}, {and} \bibinfo{person}{Sandra Avila}.}
  \bibinfo{year}{2020}\natexlab{}.
\newblock \showarticletitle{{Debiasing Skin Lesion Datasets and Models? Not So
  Fast}}. In \bibinfo{booktitle}{\emph{Computer Vision and Pattern Recognition
  (CVPR) Workshops}}. \bibinfo{pages}{740--741}.
\newblock


\bibitem[Budd et~al\mbox{.}(2021)]%
        {budd2021survey}
\bibfield{author}{\bibinfo{person}{Samuel Budd}, \bibinfo{person}{Emma~C
  Robinson}, {and} \bibinfo{person}{Bernhard Kainz}.}
  \bibinfo{year}{2021}\natexlab{}.
\newblock \showarticletitle{A survey on active learning and human-in-the-loop
  deep learning for medical image analysis}.
\newblock \bibinfo{journal}{\emph{Medical Image Analysis}}
  \bibinfo{volume}{71} (\bibinfo{year}{2021}), \bibinfo{pages}{102062}.
\newblock


\bibitem[Buneman et~al\mbox{.}(2021)]%
        {buneman2021data}
\bibfield{author}{\bibinfo{person}{Peter Buneman}, \bibinfo{person}{Dennis
  Dosso}, \bibinfo{person}{Matteo Lissandrini}, {and}
  \bibinfo{person}{Gianmaria Silvello}.} \bibinfo{year}{2021}\natexlab{}.
\newblock \showarticletitle{Data citation and the citation graph}.
\newblock \bibinfo{journal}{\emph{Quantitative Science Studies}}
  \bibinfo{volume}{2}, \bibinfo{number}{4} (\bibinfo{year}{2021}),
  \bibinfo{pages}{1399--1422}.
\newblock


\bibitem[Burgos-Artizzu et~al\mbox{.}(2020)]%
        {FetalPlanesDB}
\bibfield{author}{\bibinfo{person}{Xavier~P. Burgos-Artizzu},
  \bibinfo{person}{David Coronado-Gutierrez}, \bibinfo{person}{Brenda
  Valenzuela-Alcaraz}, \bibinfo{person}{Elisenda Bonet-Carne},
  \bibinfo{person}{Elisenda Eixarch}, \bibinfo{person}{Fatima Crispi}, {and}
  \bibinfo{person}{Eduard Gratacós}.} \bibinfo{year}{2020}\natexlab{}.
\newblock \bibinfo{booktitle}{\emph{{FETAL\_PLANES\_DB: Common maternal-fetal
  ultrasound images}}}.
\newblock
\urldef\tempurl%
\url{https://doi.org/10.5281/zenodo.3904280}
\showDOI{\tempurl}


\bibitem[Bustos et~al\mbox{.}(2020)]%
        {bustos2020padchest}
\bibfield{author}{\bibinfo{person}{Aurelia Bustos}, \bibinfo{person}{Antonio
  Pertusa}, \bibinfo{person}{Jose-Maria Salinas}, {and} \bibinfo{person}{Maria
  De~La Iglesia-Vay{\'a}}.} \bibinfo{year}{2020}\natexlab{}.
\newblock \showarticletitle{Padchest: A large chest x-ray image dataset with
  multi-label annotated reports}.
\newblock \bibinfo{journal}{\emph{Medical Image Analysis}}
  \bibinfo{volume}{66} (\bibinfo{year}{2020}), \bibinfo{pages}{101797}.
\newblock


\bibitem[Cepeda et~al\mbox{.}(2023)]%
        {cepeda2023rio}
\bibfield{author}{\bibinfo{person}{Santiago Cepeda}, \bibinfo{person}{Sergio
  Garc{\'\i}a-Garc{\'\i}a}, \bibinfo{person}{Ignacio Arrese},
  \bibinfo{person}{Francisco Herrero}, \bibinfo{person}{Trinidad Escudero},
  \bibinfo{person}{Tom{\'a}s Zamora}, {and} \bibinfo{person}{Rosario Sarabia}.}
  \bibinfo{year}{2023}\natexlab{}.
\newblock \showarticletitle{The R{\'\i}o Hortega University Hospital
  Glioblastoma dataset: A comprehensive collection of preoperative, early
  postoperative and recurrence MRI scans (RHUH-GBM)}.
\newblock \bibinfo{journal}{\emph{Data in Brief}}  \bibinfo{volume}{50}
  (\bibinfo{year}{2023}), \bibinfo{pages}{109617}.
\newblock


\bibitem[Challenge(2025)]%
        {grandchallenge}
\bibfield{author}{\bibinfo{person}{Grand Challenge}.}
  \bibinfo{year}{2025}\natexlab{}.
\newblock \bibinfo{title}{A platform for end-to-end development of machine
  learning solutions in biomedical imaging.}
\newblock \bibinfo{howpublished}{https://grand-challenge.org/}.
\newblock
\newblock
\shownote{Accessed: 2025-01-17}.


\bibitem[Chen et~al\mbox{.}(2021a)]%
        {chen2021data}
\bibfield{author}{\bibinfo{person}{Haihua Chen}, \bibinfo{person}{Jiangping
  Chen}, {and} \bibinfo{person}{Junhua Ding}.}
  \bibinfo{year}{2021}\natexlab{a}.
\newblock \showarticletitle{Data evaluation and enhancement for quality
  improvement of machine learning}.
\newblock \bibinfo{journal}{\emph{IEEE Transactions on Reliability}}
  \bibinfo{volume}{70}, \bibinfo{number}{2} (\bibinfo{year}{2021}),
  \bibinfo{pages}{831--847}.
\newblock


\bibitem[Chen et~al\mbox{.}(2021b)]%
        {chen2021ethical}
\bibfield{author}{\bibinfo{person}{Irene~Y Chen}, \bibinfo{person}{Emma
  Pierson}, \bibinfo{person}{Sherri Rose}, \bibinfo{person}{Shalmali Joshi},
  \bibinfo{person}{Kadija Ferryman}, {and} \bibinfo{person}{Marzyeh Ghassemi}.}
  \bibinfo{year}{2021}\natexlab{b}.
\newblock \showarticletitle{Ethical machine learning in healthcare}.
\newblock \bibinfo{journal}{\emph{Annual Review of Biomedical Data Science}}
  \bibinfo{volume}{4} (\bibinfo{year}{2021}), \bibinfo{pages}{123--144}.
\newblock


\bibitem[Chen et~al\mbox{.}(2023)]%
        {chen2023algorithm}
\bibfield{author}{\bibinfo{person}{Richard~J. Chen}, \bibinfo{person}{Judy~J.
  Wang}, \bibinfo{person}{Drew F.~K. Williamson}, \bibinfo{person}{Tiffany~Y.
  Chen}, \bibinfo{person}{Jana Lipkova}, \bibinfo{person}{Ming~Y. Lu},
  \bibinfo{person}{Sharifa Sahai}, {and} \bibinfo{person}{Faisal Mahmood}.}
  \bibinfo{year}{2023}\natexlab{}.
\newblock \showarticletitle{Algorithmic fairness in artificial intelligence for
  medicine and healthcare}.
\newblock \bibinfo{journal}{\emph{Nature Biomedical Engineering}}
  \bibinfo{volume}{7}, \bibinfo{number}{6} (\bibinfo{date}{June}
  \bibinfo{year}{2023}), \bibinfo{pages}{719–742}.
\newblock
\showISSN{2157-846X}
\urldef\tempurl%
\url{https://doi.org/10.1038/s41551-023-01056-8}
\showDOI{\tempurl}


\bibitem[Cheplygina et~al\mbox{.}(2019)]%
        {cheplygina2019not}
\bibfield{author}{\bibinfo{person}{Veronika Cheplygina},
  \bibinfo{person}{Marleen de Bruijne}, {and} \bibinfo{person}{Josien~PW
  Pluim}.} \bibinfo{year}{2019}\natexlab{}.
\newblock \showarticletitle{Not-so-supervised: a survey of semi-supervised,
  multi-instance, and transfer learning in medical image analysis}.
\newblock \bibinfo{journal}{\emph{Medical image analysis}}
  \bibinfo{volume}{54} (\bibinfo{year}{2019}), \bibinfo{pages}{280--296}.
\newblock


\bibitem[Collins et~al\mbox{.}(2024a)]%
        {Collins2024}
\bibfield{author}{\bibinfo{person}{Gary~S Collins}, \bibinfo{person}{Karel G~M
  Moons}, \bibinfo{person}{Paula Dhiman}, \bibinfo{person}{Richard~D Riley},
  \bibinfo{person}{Andrew~L Beam}, \bibinfo{person}{Ben Van~Calster},
  \bibinfo{person}{Marzyeh Ghassemi}, \bibinfo{person}{Xiaoxuan Liu},
  \bibinfo{person}{Johannes~B Reitsma}, \bibinfo{person}{Maarten van Smeden},
  \bibinfo{person}{Anne-Laure Boulesteix}, \bibinfo{person}{Jennifer~Catherine
  Camaradou}, \bibinfo{person}{Leo~Anthony Celi}, \bibinfo{person}{Spiros
  Denaxas}, \bibinfo{person}{Alastair~K Denniston}, \bibinfo{person}{Ben
  Glocker}, \bibinfo{person}{Robert~M Golub}, \bibinfo{person}{Hugh Harvey},
  \bibinfo{person}{Georg Heinze}, \bibinfo{person}{Michael~M Hoffman},
  \bibinfo{person}{André~Pascal Kengne}, \bibinfo{person}{Emily Lam},
  \bibinfo{person}{Naomi Lee}, \bibinfo{person}{Elizabeth~W Loder},
  \bibinfo{person}{Lena Maier-Hein}, \bibinfo{person}{Bilal~A Mateen},
  \bibinfo{person}{Melissa~D McCradden}, \bibinfo{person}{Lauren
  Oakden-Rayner}, \bibinfo{person}{Johan Ordish}, \bibinfo{person}{Richard
  Parnell}, \bibinfo{person}{Sherri Rose}, \bibinfo{person}{Karandeep Singh},
  \bibinfo{person}{Laure Wynants}, {and} \bibinfo{person}{Patricia Logullo}.}
  \bibinfo{year}{2024}\natexlab{a}.
\newblock \showarticletitle{{TRIPOD+AI} statement: updated guidance for
  reporting clinical prediction models that use regression or machine learning
  methods}.
\newblock \bibinfo{journal}{\emph{BMJ}} (\bibinfo{date}{April}
  \bibinfo{year}{2024}), \bibinfo{pages}{e078378}.
\newblock
\showISSN{1756-1833}
\urldef\tempurl%
\url{https://doi.org/10.1136/bmj-2023-078378}
\showDOI{\tempurl}


\bibitem[Collins et~al\mbox{.}(2024b)]%
        {collins2024open}
\bibfield{author}{\bibinfo{person}{Gary~S Collins}, \bibinfo{person}{Rebecca
  Whittle}, \bibinfo{person}{Garrett~S Bullock}, \bibinfo{person}{Patricia
  Logullo}, \bibinfo{person}{Paula Dhiman}, \bibinfo{person}{Jennifer~A de
  Beyer}, \bibinfo{person}{Richard~D Riley}, {and} \bibinfo{person}{Michael~M
  Schlussel}.} \bibinfo{year}{2024}\natexlab{b}.
\newblock \showarticletitle{Open science practices need substantial improvement
  in prognostic model studies in oncology using machine learning}.
\newblock \bibinfo{journal}{\emph{Journal of Clinical Epidemiology}}
  \bibinfo{volume}{165} (\bibinfo{year}{2024}), \bibinfo{pages}{111199}.
\newblock


\bibitem[Compton et~al\mbox{.}(2023)]%
        {compton2023more}
\bibfield{author}{\bibinfo{person}{Rhys Compton}, \bibinfo{person}{Lily Zhang},
  \bibinfo{person}{Aahlad Puli}, {and} \bibinfo{person}{Rajesh Ranganath}.}
  \bibinfo{year}{2023}\natexlab{}.
\newblock \showarticletitle{When more is less: Incorporating additional
  datasets can hurt performance by introducing spurious correlations}. In
  \bibinfo{booktitle}{\emph{Machine Learning for Healthcare Conference}}. PMLR,
  \bibinfo{pages}{110--127}.
\newblock


\bibitem[Dai et~al\mbox{.}(2015)]%
        {dai2015boxsup}
\bibfield{author}{\bibinfo{person}{Jifeng Dai}, \bibinfo{person}{Kaiming He},
  {and} \bibinfo{person}{Jian Sun}.} \bibinfo{year}{2015}\natexlab{}.
\newblock \showarticletitle{Boxsup: Exploiting bounding boxes to supervise
  convolutional networks for semantic segmentation}. In
  \bibinfo{booktitle}{\emph{International Conference on Computer Vision
  (ICCV)}}. \bibinfo{pages}{1635--1643}.
\newblock


\bibitem[Damgaard et~al\mbox{.}(2023)]%
        {damgaard2023augmenting}
\bibfield{author}{\bibinfo{person}{Cathrine Damgaard},
  \bibinfo{person}{Trine~Naja Eriksen}, \bibinfo{person}{Dovile Juodelyte},
  \bibinfo{person}{Veronika Cheplygina}, {and} \bibinfo{person}{Amelia
  Jim{\'e}nez-S{\'a}nchez}.} \bibinfo{year}{2023}\natexlab{}.
\newblock \showarticletitle{Augmenting Chest X-ray Datasets with Non-Expert
  Annotations}.
\newblock \bibinfo{journal}{\emph{arXiv preprint arXiv:2309.02244}}
  (\bibinfo{year}{2023}).
\newblock


\bibitem[Daneshjou et~al\mbox{.}(2021)]%
        {daneshjou2021lack}
\bibfield{author}{\bibinfo{person}{Roxana Daneshjou}, \bibinfo{person}{Mary~P
  Smith}, \bibinfo{person}{Mary~D Sun}, \bibinfo{person}{Veronica Rotemberg},
  {and} \bibinfo{person}{James Zou}.} \bibinfo{year}{2021}\natexlab{}.
\newblock \showarticletitle{Lack of transparency and potential bias in
  artificial intelligence data sets and algorithms: a scoping review}.
\newblock \bibinfo{journal}{\emph{JAMA Dermatology}} \bibinfo{volume}{157},
  \bibinfo{number}{11} (\bibinfo{year}{2021}), \bibinfo{pages}{1362--1369}.
\newblock


\bibitem[Daniel et~al\mbox{.}(2018)]%
        {daniel2018quality}
\bibfield{author}{\bibinfo{person}{Florian Daniel}, \bibinfo{person}{Pavel
  Kucherbaev}, \bibinfo{person}{Cinzia Cappiello}, \bibinfo{person}{Boualem
  Benatallah}, {and} \bibinfo{person}{Mohammad Allahbakhsh}.}
  \bibinfo{year}{2018}\natexlab{}.
\newblock \showarticletitle{Quality control in crowdsourcing: A survey of
  quality attributes, assessment techniques, and assurance actions}.
\newblock \bibinfo{journal}{\emph{ACM Computing Surveys (CSUR)}}
  \bibinfo{volume}{51}, \bibinfo{number}{1} (\bibinfo{year}{2018}),
  \bibinfo{pages}{1--40}.
\newblock


\bibitem[Dashe et~al\mbox{.}(2009)]%
        {dasheObesity2009}
\bibfield{author}{\bibinfo{person}{Jodi~S. Dashe}, \bibinfo{person}{Donald~D.
  McIntire}, {and} \bibinfo{person}{Diane~M. Twickler}.}
  \bibinfo{year}{2009}\natexlab{}.
\newblock \showarticletitle{Maternal {{Obesity Limits}} the {{Ultrasound
  Evaluation}} of {{Fetal Anatomy}}}.
\newblock \bibinfo{journal}{\emph{Journal of Ultrasound in Medicine}}
  \bibinfo{volume}{28}, \bibinfo{number}{8} (\bibinfo{year}{2009}),
  \bibinfo{pages}{1025--1030}.
\newblock
\showISSN{1550-9613}
\urldef\tempurl%
\url{https://doi.org/10.7863/jum.2009.28.8.1025}
\showDOI{\tempurl}


\bibitem[DeGrave et~al\mbox{.}(2021)]%
        {degrave2021ai}
\bibfield{author}{\bibinfo{person}{Alex~J DeGrave}, \bibinfo{person}{Joseph~D
  Janizek}, {and} \bibinfo{person}{Su-In Lee}.}
  \bibinfo{year}{2021}\natexlab{}.
\newblock \showarticletitle{AI for radiographic COVID-19 detection selects
  shortcuts over signal}.
\newblock \bibinfo{journal}{\emph{Nature Machine Intelligence}}
  \bibinfo{volume}{3}, \bibinfo{number}{7} (\bibinfo{year}{2021}),
  \bibinfo{pages}{610--619}.
\newblock


\bibitem[Dhiman et~al\mbox{.}(2023)]%
        {dhiman2023sample}
\bibfield{author}{\bibinfo{person}{Paula Dhiman}, \bibinfo{person}{Jie Ma},
  \bibinfo{person}{Cathy Qi}, \bibinfo{person}{Garrett Bullock},
  \bibinfo{person}{Jamie~C Sergeant}, \bibinfo{person}{Richard~D Riley}, {and}
  \bibinfo{person}{Gary~S Collins}.} \bibinfo{year}{2023}\natexlab{}.
\newblock \showarticletitle{Sample size requirements are not being considered
  in studies developing prediction models for binary outcomes: a systematic
  review}.
\newblock \bibinfo{journal}{\emph{BMC Medical Research Methodology}}
  \bibinfo{volume}{23}, \bibinfo{number}{1} (\bibinfo{year}{2023}),
  \bibinfo{pages}{188}.
\newblock


\bibitem[Dibaji et~al\mbox{.}(2023)]%
        {dibaji2023studying}
\bibfield{author}{\bibinfo{person}{Mahsa Dibaji}, \bibinfo{person}{Neha
  Gianchandani}, \bibinfo{person}{Akhil Nair}, \bibinfo{person}{Mansi Singhal},
  \bibinfo{person}{Roberto Souza}, {and} \bibinfo{person}{Mariana Bento}.}
  \bibinfo{year}{2023}\natexlab{}.
\newblock \showarticletitle{Studying the Effects of Sex-Related Differences on
  Brain Age Prediction Using Brain MR Imaging}. In
  \bibinfo{booktitle}{\emph{MICCAI Workshop on Clinical Image-Based
  Procedures}}. Springer, \bibinfo{pages}{205--214}.
\newblock


\bibitem[Dishner et~al\mbox{.}(2024)]%
        {Dishner2024}
\bibfield{author}{\bibinfo{person}{Katharine~A. Dishner}, \bibinfo{person}{Bala
  McRae-Posani}, \bibinfo{person}{Arka Bhowmik}, \bibinfo{person}{Maxine~S.
  Jochelson}, \bibinfo{person}{Andrei Holodny}, \bibinfo{person}{Katja Pinker},
  \bibinfo{person}{Sarah Eskreis-Winkler}, {and} \bibinfo{person}{Joseph~N.
  Stember}.} \bibinfo{year}{2024}\natexlab{}.
\newblock \bibinfo{title}{A Survey of Publicly Available MRI Datasets for
  Potential Use in Artificial Intelligence Research}.
\newblock , \bibinfo{numpages}{450-480}~pages.
\newblock
Issue 2.
\showISSN{15222586}
\urldef\tempurl%
\url{https://doi.org/10.1002/jmri.29101}
\showDOI{\tempurl}


\bibitem[Drooger et~al\mbox{.}(2005)]%
        {droogerEthnic2005}
\bibfield{author}{\bibinfo{person}{J.~C. Drooger}, \bibinfo{person}{J.~W.~M.
  Troe}, \bibinfo{person}{G.~J. J.~M. Borsboom}, \bibinfo{person}{A. Hofman},
  \bibinfo{person}{J.~P. Mackenbach}, \bibinfo{person}{H.~A. Moll},
  \bibinfo{person}{R.~J.~M. Snijders}, \bibinfo{person}{F.~C. Verhulst},
  \bibinfo{person}{J.~C.~M. Witteman}, \bibinfo{person}{E.~a.~P. Steegers},
  {and} \bibinfo{person}{I.~M.~A. Joung}.} \bibinfo{year}{2005}\natexlab{}.
\newblock \showarticletitle{Ethnic Differences in Prenatal Growth and the
  Association with Maternal and Fetal Characteristics}.
\newblock \bibinfo{journal}{\emph{Ultrasound in Obstetrics \& Gynecology}}
  \bibinfo{volume}{26}, \bibinfo{number}{2} (\bibinfo{year}{2005}),
  \bibinfo{pages}{115--122}.
\newblock
\showISSN{1469-0705}
\urldef\tempurl%
\url{https://doi.org/10.1002/uog.1962}
\showDOI{\tempurl}


\bibitem[Ebbehoj et~al\mbox{.}(2022)]%
        {ebbehoj2022transfer}
\bibfield{author}{\bibinfo{person}{Andreas Ebbehoj},
  \bibinfo{person}{Mette~{\O}stergaard Thunbo}, \bibinfo{person}{Ole~Emil
  Andersen}, \bibinfo{person}{Michala~Vilstrup Glindtvad}, {and}
  \bibinfo{person}{Adam Hulman}.} \bibinfo{year}{2022}\natexlab{}.
\newblock \showarticletitle{Transfer learning for non-image data in clinical
  research: a scoping review}.
\newblock \bibinfo{journal}{\emph{PLOS Digital Health}} \bibinfo{volume}{1},
  \bibinfo{number}{2} (\bibinfo{year}{2022}), \bibinfo{pages}{e0000014}.
\newblock


\bibitem[Eisenmann et~al\mbox{.}(2022)]%
        {eisenmann2022biomedical}
\bibfield{author}{\bibinfo{person}{Matthias Eisenmann}, \bibinfo{person}{Annika
  Reinke}, \bibinfo{person}{Vivienn Weru}, \bibinfo{person}{Minu~Dietlinde
  Tizabi}, \bibinfo{person}{Fabian Isensee}, \bibinfo{person}{Tim~J Adler},
  \bibinfo{person}{Patrick Godau}, \bibinfo{person}{Veronika Cheplygina},
  \bibinfo{person}{Michal Kozubek}, \bibinfo{person}{Sharib Ali},
  {et~al\mbox{.}}} \bibinfo{year}{2022}\natexlab{}.
\newblock \showarticletitle{Biomedical image analysis competitions: The state
  of current participation practice}.
\newblock \bibinfo{journal}{\emph{arXiv preprint arXiv:2212.08568}}
  (\bibinfo{year}{2022}).
\newblock


\bibitem[{European Organization For Nuclear Research} and {OpenAIRE}(2013)]%
        {zenodo2013}
\bibfield{author}{\bibinfo{person}{{European Organization For Nuclear
  Research}} {and} \bibinfo{person}{{OpenAIRE}}.}
  \bibinfo{year}{2013}\natexlab{}.
\newblock \bibinfo{title}{Zenodo}.
\newblock
\newblock
\urldef\tempurl%
\url{https://doi.org/10.25495/7GXK-RD71}
\showDOI{\tempurl}


\bibitem[Fang et~al\mbox{.}(2024)]%
        {fang2024source}
\bibfield{author}{\bibinfo{person}{Yuqi Fang}, \bibinfo{person}{Pew-Thian Yap},
  \bibinfo{person}{Weili Lin}, \bibinfo{person}{Hongtu Zhu}, {and}
  \bibinfo{person}{Mingxia Liu}.} \bibinfo{year}{2024}\natexlab{}.
\newblock \showarticletitle{Source-free unsupervised domain adaptation: A
  survey}.
\newblock \bibinfo{journal}{\emph{Neural Networks}} (\bibinfo{year}{2024}),
  \bibinfo{pages}{106230}.
\newblock


\bibitem[Feinberg(2017)]%
        {Feinberg2017}
\bibfield{author}{\bibinfo{person}{Melanie Feinberg}.}
  \bibinfo{year}{2017}\natexlab{}.
\newblock \showarticletitle{A design perspective on data}. In
  \bibinfo{booktitle}{\emph{Conference on Human Factors in Computing Systems
  (CHI)}}, Vol.~\bibinfo{volume}{2017-May}. \bibinfo{pages}{2952--2963}.
\newblock
\showISBNx{9781450346559}
\urldef\tempurl%
\url{https://doi.org/10.1145/3025453.3025837}
\showDOI{\tempurl}


\bibitem[Fink et~al\mbox{.}(2023)]%
        {fink2023potential}
\bibfield{author}{\bibinfo{person}{Matthias~A Fink}, \bibinfo{person}{Arved
  Bischoff}, \bibinfo{person}{Christoph~A Fink}, \bibinfo{person}{Martin Moll},
  \bibinfo{person}{Jonas Kroschke}, \bibinfo{person}{Luca Dulz},
  \bibinfo{person}{Claus~Peter Heu{\ss}el}, \bibinfo{person}{Hans-Ulrich
  Kauczor}, {and} \bibinfo{person}{Tim~F Weber}.}
  \bibinfo{year}{2023}\natexlab{}.
\newblock \showarticletitle{Potential of ChatGPT and GPT-4 for data mining of
  free-text CT reports on lung cancer}.
\newblock \bibinfo{journal}{\emph{Radiology}} \bibinfo{volume}{308},
  \bibinfo{number}{3} (\bibinfo{year}{2023}), \bibinfo{pages}{e231362}.
\newblock


\bibitem[Fort(2016)]%
        {fort2016}
\bibfield{author}{\bibinfo{person}{Karën Fort}.}
  \bibinfo{year}{2016}\natexlab{}.
\newblock \bibinfo{booktitle}{\emph{Collaborative Annotation for Reliable
  Natural Language Processing: Technical and Sociological Aspects}}.
\newblock \bibinfo{publisher}{Wiley}. 1--164 pages.
\newblock
\showISBNx{9781119306696}
\urldef\tempurl%
\url{https://doi.org/10.1002/9781119306696}
\showDOI{\tempurl}


\bibitem[Frachtenberg(2022)]%
        {frachtenberg2022research}
\bibfield{author}{\bibinfo{person}{Eitan Frachtenberg}.}
  \bibinfo{year}{2022}\natexlab{}.
\newblock \showarticletitle{Research artifacts and citations in computer
  systems papers}.
\newblock \bibinfo{journal}{\emph{PeerJ Computer Science}}  \bibinfo{volume}{8}
  (\bibinfo{year}{2022}), \bibinfo{pages}{e887}.
\newblock


\bibitem[Fricke(2018)]%
        {fricke2018semantic}
\bibfield{author}{\bibinfo{person}{Suzanne Fricke}.}
  \bibinfo{year}{2018}\natexlab{}.
\newblock \showarticletitle{Semantic scholar}.
\newblock \bibinfo{journal}{\emph{Journal of the Medical Library Association}}
  \bibinfo{volume}{106}, \bibinfo{number}{1} (\bibinfo{year}{2018}),
  \bibinfo{pages}{145}.
\newblock


\bibitem[Gaggion et~al\mbox{.}(2023)]%
        {gaggion2023chexmask}
\bibfield{author}{\bibinfo{person}{Nicol{\'a}s Gaggion},
  \bibinfo{person}{Candelaria Mosquera}, \bibinfo{person}{Lucas Mansilla},
  \bibinfo{person}{Martina Aineseder}, \bibinfo{person}{Diego~H Milone}, {and}
  \bibinfo{person}{Enzo Ferrante}.} \bibinfo{year}{2023}\natexlab{}.
\newblock \showarticletitle{CheXmask: a large-scale dataset of anatomical
  segmentation masks for multi-center chest x-ray images}.
\newblock \bibinfo{journal}{\emph{arXiv preprint arXiv:2307.03293}}
  (\bibinfo{year}{2023}).
\newblock


\bibitem[Gal et~al\mbox{.}(2017)]%
        {gal2017deep}
\bibfield{author}{\bibinfo{person}{Yarin Gal}, \bibinfo{person}{Riashat Islam},
  {and} \bibinfo{person}{Zoubin Ghahramani}.} \bibinfo{year}{2017}\natexlab{}.
\newblock \showarticletitle{Deep bayesian active learning with image data}. In
  \bibinfo{booktitle}{\emph{International Conference on Machine Learning
  (ICML)}}. PMLR, \bibinfo{pages}{1183--1192}.
\newblock


\bibitem[Galanty et~al\mbox{.}(2024)]%
        {galanty2024assessingdoc}
\bibfield{author}{\bibinfo{person}{Maria Galanty}, \bibinfo{person}{Dieuwertje
  Luitse}, \bibinfo{person}{Sijm~H. Noteboom}, \bibinfo{person}{Philip Croon},
  \bibinfo{person}{Alexander~P. Vlaar}, \bibinfo{person}{Thomas Poell},
  \bibinfo{person}{Clara~I. Sanchez}, \bibinfo{person}{Tobias Blanke}, {and}
  \bibinfo{person}{Ivana Išgum}.} \bibinfo{year}{2024}\natexlab{}.
\newblock \showarticletitle{Assessing the documentation of publicly available
  medical image and signal datasets and their impact on bias using the BEAMRAD
  tool}.
\newblock \bibinfo{journal}{\emph{Scientific Reports}} \bibinfo{volume}{14},
  \bibinfo{number}{1} (\bibinfo{date}{Dec.} \bibinfo{year}{2024}).
\newblock
\showISSN{2045-2322}
\urldef\tempurl%
\url{https://doi.org/10.1038/s41598-024-83218-5}
\showDOI{\tempurl}


\bibitem[Ganapathi et~al\mbox{.}(2022)]%
        {ganapathi2022tackling}
\bibfield{author}{\bibinfo{person}{Shaswath Ganapathi}, \bibinfo{person}{Jo
  Palmer}, \bibinfo{person}{Joseph~E Alderman}, \bibinfo{person}{Melanie
  Calvert}, \bibinfo{person}{Cyrus Espinoza}, \bibinfo{person}{Jacqui Gath},
  \bibinfo{person}{Marzyeh Ghassemi}, \bibinfo{person}{Katherine Heller},
  \bibinfo{person}{Francis Mckay}, \bibinfo{person}{Alan Karthikesalingam},
  {et~al\mbox{.}}} \bibinfo{year}{2022}\natexlab{}.
\newblock \showarticletitle{Tackling bias in AI health datasets through the
  STANDING Together initiative}.
\newblock \bibinfo{journal}{\emph{Nature Medicine}} \bibinfo{volume}{28},
  \bibinfo{number}{11} (\bibinfo{year}{2022}), \bibinfo{pages}{2232--2233}.
\newblock


\bibitem[Garbin and Marques(2022)]%
        {garbin2022assessing}
\bibfield{author}{\bibinfo{person}{Christian Garbin} {and} \bibinfo{person}{Oge
  Marques}.} \bibinfo{year}{2022}\natexlab{}.
\newblock \showarticletitle{Assessing methods and tools to improve reporting,
  increase transparency, and reduce failures in machine learning applications
  in health care}.
\newblock \bibinfo{journal}{\emph{Radiology: Artificial Intelligence}}
  \bibinfo{volume}{4}, \bibinfo{number}{2} (\bibinfo{year}{2022}),
  \bibinfo{pages}{e210127}.
\newblock


\bibitem[Garbin et~al\mbox{.}(2021)]%
        {garbin2021structured}
\bibfield{author}{\bibinfo{person}{Christian Garbin}, \bibinfo{person}{Pranav
  Rajpurkar}, \bibinfo{person}{Jeremy Irvin}, \bibinfo{person}{Matthew~P
  Lungren}, {and} \bibinfo{person}{Oge Marques}.}
  \bibinfo{year}{2021}\natexlab{}.
\newblock \showarticletitle{Structured dataset documentation: a datasheet for
  {CheXpert}}.
\newblock \bibinfo{journal}{\emph{arXiv preprint arXiv:2105.03020}}
  (\bibinfo{year}{2021}).
\newblock


\bibitem[Gebru et~al\mbox{.}(2021)]%
        {gebru2021datasheets}
\bibfield{author}{\bibinfo{person}{Timnit Gebru}, \bibinfo{person}{Jamie
  Morgenstern}, \bibinfo{person}{Briana Vecchione},
  \bibinfo{person}{Jennifer~Wortman Vaughan}, \bibinfo{person}{Hanna Wallach},
  \bibinfo{person}{Hal~Daum{\'e} Iii}, {and} \bibinfo{person}{Kate Crawford}.}
  \bibinfo{year}{2021}\natexlab{}.
\newblock \showarticletitle{Datasheets for datasets}.
\newblock \bibinfo{journal}{\emph{Commun. ACM}} \bibinfo{volume}{64},
  \bibinfo{number}{12} (\bibinfo{year}{2021}), \bibinfo{pages}{86--92}.
\newblock


\bibitem[Geirhos et~al\mbox{.}(2018)]%
        {geirhos2018generalisation}
\bibfield{author}{\bibinfo{person}{Robert Geirhos}, \bibinfo{person}{Carlos~RM
  Temme}, \bibinfo{person}{Jonas Rauber}, \bibinfo{person}{Heiko~H Sch{\"u}tt},
  \bibinfo{person}{Matthias Bethge}, {and} \bibinfo{person}{Felix~A Wichmann}.}
  \bibinfo{year}{2018}\natexlab{}.
\newblock \showarticletitle{Generalisation in humans and deep neural networks}.
  In \bibinfo{booktitle}{\emph{Neural Information Processing Systems
  (NeurIPS)}}. \bibinfo{pages}{7538--7550}.
\newblock


\bibitem[Gichoya et~al\mbox{.}(2022)]%
        {gichoya2022ai}
\bibfield{author}{\bibinfo{person}{Judy~Wawira Gichoya}, \bibinfo{person}{Imon
  Banerjee}, \bibinfo{person}{Ananth~Reddy Bhimireddy}, \bibinfo{person}{John~L
  Burns}, \bibinfo{person}{Leo~Anthony Celi}, \bibinfo{person}{Li-Ching Chen},
  \bibinfo{person}{Ramon Correa}, \bibinfo{person}{Natalie Dullerud},
  \bibinfo{person}{Marzyeh Ghassemi}, \bibinfo{person}{Shih-Cheng Huang},
  {et~al\mbox{.}}} \bibinfo{year}{2022}\natexlab{}.
\newblock \showarticletitle{AI recognition of patient race in medical imaging:
  a modelling study}.
\newblock \bibinfo{journal}{\emph{The Lancet Digital Health}}
  \bibinfo{volume}{4}, \bibinfo{number}{6} (\bibinfo{year}{2022}),
  \bibinfo{pages}{e406--e414}.
\newblock


\bibitem[Gitelman(2013)]%
        {gitelman2013RawOxymoron}
\bibfield{editor}{\bibinfo{person}{Lisa Gitelman}} (Ed.).
  \bibinfo{year}{2013}\natexlab{}.
\newblock \bibinfo{booktitle}{\emph{“Raw Data” Is an Oxymoron}}.
\newblock \bibinfo{publisher}{MIT Press}.
\newblock
\showISBNx{0262518287}
\newblock
\shownote{Includes bibliographical references and index}.


\bibitem[Glocker et~al\mbox{.}(2023)]%
        {glocker2023algorithmic}
\bibfield{author}{\bibinfo{person}{Ben Glocker}, \bibinfo{person}{Charles
  Jones}, \bibinfo{person}{M{\'e}lanie Bernhardt}, {and}
  \bibinfo{person}{Stefan Winzeck}.} \bibinfo{year}{2023}\natexlab{}.
\newblock \showarticletitle{Algorithmic encoding of protected characteristics
  in chest X-ray disease detection models}.
\newblock \bibinfo{journal}{\emph{EBioMedicine}}  \bibinfo{volume}{89}
  (\bibinfo{year}{2023}).
\newblock


\bibitem[Goldberger et~al\mbox{.}(2000)]%
        {goldberger2000physiobank}
\bibfield{author}{\bibinfo{person}{Ary~L Goldberger}, \bibinfo{person}{Luis~AN
  Amaral}, \bibinfo{person}{Leon Glass}, \bibinfo{person}{Jeffrey~M Hausdorff},
  \bibinfo{person}{Plamen~Ch Ivanov}, \bibinfo{person}{Roger~G Mark},
  \bibinfo{person}{Joseph~E Mietus}, \bibinfo{person}{George~B Moody},
  \bibinfo{person}{Chung-Kang Peng}, {and} \bibinfo{person}{H~Eugene Stanley}.}
  \bibinfo{year}{2000}\natexlab{}.
\newblock \showarticletitle{{PhysioBank, PhysioToolkit, and PhysioNet:}
  components of a new research resource for complex physiologic signals}.
\newblock \bibinfo{journal}{\emph{Circulation}} \bibinfo{volume}{101},
  \bibinfo{number}{23} (\bibinfo{year}{2000}), \bibinfo{pages}{e215--e220}.
\newblock


\bibitem[Gorgolewski et~al\mbox{.}(2017)]%
        {gorgolewski2017bids}
\bibfield{author}{\bibinfo{person}{Krzysztof~J Gorgolewski},
  \bibinfo{person}{Fidel Alfaro-Almagro}, \bibinfo{person}{Tibor Auer},
  \bibinfo{person}{Pierre Bellec}, \bibinfo{person}{Mihai Capot{\u{a}}},
  \bibinfo{person}{M~Mallar Chakravarty}, \bibinfo{person}{Nathan~W Churchill},
  \bibinfo{person}{Alexander~Li Cohen}, \bibinfo{person}{R~Cameron Craddock},
  \bibinfo{person}{Gabriel~A Devenyi}, {et~al\mbox{.}}}
  \bibinfo{year}{2017}\natexlab{}.
\newblock \showarticletitle{BIDS apps: Improving ease of use, accessibility,
  and reproducibility of neuroimaging data analysis methods}.
\newblock \bibinfo{journal}{\emph{PLoS Computational Biology}}
  \bibinfo{volume}{13}, \bibinfo{number}{3} (\bibinfo{year}{2017}),
  \bibinfo{pages}{e1005209}.
\newblock


\bibitem[Gorgolewski et~al\mbox{.}(2016)]%
        {Gorgolewski2016}
\bibfield{author}{\bibinfo{person}{Krzysztof~J. Gorgolewski},
  \bibinfo{person}{Tibor Auer}, \bibinfo{person}{Vince~D. Calhoun},
  \bibinfo{person}{R.~Cameron Craddock}, \bibinfo{person}{Samir Das},
  \bibinfo{person}{Eugene~P. Duff}, \bibinfo{person}{Guillaume Flandin},
  \bibinfo{person}{Satrajit~S. Ghosh}, \bibinfo{person}{Tristan Glatard},
  \bibinfo{person}{Yaroslav~O. Halchenko}, \bibinfo{person}{Daniel~A.
  Handwerker}, \bibinfo{person}{Michael Hanke}, \bibinfo{person}{David Keator},
  \bibinfo{person}{Xiangrui Li}, \bibinfo{person}{Zachary Michael},
  \bibinfo{person}{Camille Maumet}, \bibinfo{person}{B.~Nolan Nichols},
  \bibinfo{person}{Thomas~E. Nichols}, \bibinfo{person}{John Pellman},
  \bibinfo{person}{Jean~Baptiste Poline}, \bibinfo{person}{Ariel Rokem},
  \bibinfo{person}{Gunnar Schaefer}, \bibinfo{person}{Vanessa Sochat},
  \bibinfo{person}{William Triplett}, \bibinfo{person}{Jessica~A. Turner},
  \bibinfo{person}{Gaël Varoquaux}, {and} \bibinfo{person}{Russell~A.
  Poldrack}.} \bibinfo{year}{2016}\natexlab{}.
\newblock \showarticletitle{The brain imaging data structure, a format for
  organizing and describing outputs of neuroimaging experiments}.
\newblock \bibinfo{journal}{\emph{Scientific Data}}  \bibinfo{volume}{3}
  (\bibinfo{date}{6} \bibinfo{year}{2016}).
\newblock
\showISSN{20524463}
\urldef\tempurl%
\url{https://doi.org/10.1038/sdata.2016.44}
\showDOI{\tempurl}


\bibitem[Gossmann et~al\mbox{.}(2018)]%
        {gossmann2018test}
\bibfield{author}{\bibinfo{person}{Alexej Gossmann}, \bibinfo{person}{Aria
  Pezeshk}, {and} \bibinfo{person}{Berkman Sahiner}.}
  \bibinfo{year}{2018}\natexlab{}.
\newblock \showarticletitle{Test data reuse for evaluation of adaptive machine
  learning algorithms: over-fitting to a fixed 'test' dataset and a potential
  solution}. In \bibinfo{booktitle}{\emph{SPIE Medical Imaging}},
  Vol.~\bibinfo{volume}{10577}. SPIE, \bibinfo{pages}{121--132}.
\newblock


\bibitem[Groh et~al\mbox{.}(2024)]%
        {groh2024deep}
\bibfield{author}{\bibinfo{person}{Matthew Groh}, \bibinfo{person}{Omar Badri},
  \bibinfo{person}{Roxana Daneshjou}, \bibinfo{person}{Arash Koochek},
  \bibinfo{person}{Caleb Harris}, \bibinfo{person}{Luis~R Soenksen},
  \bibinfo{person}{P~Murali Doraiswamy}, {and} \bibinfo{person}{Rosalind
  Picard}.} \bibinfo{year}{2024}\natexlab{}.
\newblock \showarticletitle{Deep learning-aided decision support for diagnosis
  of skin disease across skin tones}.
\newblock \bibinfo{journal}{\emph{Nature Medicine}} \bibinfo{volume}{30},
  \bibinfo{number}{2} (\bibinfo{year}{2024}), \bibinfo{pages}{573--583}.
\newblock


\bibitem[Groh et~al\mbox{.}(2022)]%
        {groh2022towards}
\bibfield{author}{\bibinfo{person}{Matthew Groh}, \bibinfo{person}{Caleb
  Harris}, \bibinfo{person}{Roxana Daneshjou}, \bibinfo{person}{Omar Badri},
  {and} \bibinfo{person}{Arash Koochek}.} \bibinfo{year}{2022}\natexlab{}.
\newblock \showarticletitle{Towards transparency in dermatology image datasets
  with skin tone annotations by experts, crowds, and an algorithm}.
\newblock \bibinfo{journal}{\emph{Proceedings of the ACM on Human-Computer
  Interaction}} \bibinfo{volume}{6}, \bibinfo{number}{CSCW2}
  (\bibinfo{year}{2022}), \bibinfo{pages}{1--26}.
\newblock


\bibitem[Groh et~al\mbox{.}(2021)]%
        {groh2021evaluating}
\bibfield{author}{\bibinfo{person}{Matthew Groh}, \bibinfo{person}{Caleb
  Harris}, \bibinfo{person}{Luis Soenksen}, \bibinfo{person}{Felix Lau},
  \bibinfo{person}{Rachel Han}, \bibinfo{person}{Aerin Kim},
  \bibinfo{person}{Arash Koochek}, {and} \bibinfo{person}{Omar Badri}.}
  \bibinfo{year}{2021}\natexlab{}.
\newblock \showarticletitle{Evaluating deep neural networks trained on clinical
  images in dermatology with the fitzpatrick 17k dataset}. In
  \bibinfo{booktitle}{\emph{Computer Vision and Pattern Recognition (CVPR)}}.
  \bibinfo{pages}{1820--1828}.
\newblock


\bibitem[Guan and Liu(2021)]%
        {guan2021domain}
\bibfield{author}{\bibinfo{person}{Hao Guan} {and} \bibinfo{person}{Mingxia
  Liu}.} \bibinfo{year}{2021}\natexlab{}.
\newblock \showarticletitle{Domain adaptation for medical image analysis: a
  survey}.
\newblock \bibinfo{journal}{\emph{IEEE Transactions on Biomedical Engineering}}
  \bibinfo{volume}{69}, \bibinfo{number}{3} (\bibinfo{year}{2021}),
  \bibinfo{pages}{1173--1185}.
\newblock


\bibitem[Haider et~al\mbox{.}(2015)]%
        {haider2015unconscious}
\bibfield{author}{\bibinfo{person}{Adil~H Haider}, \bibinfo{person}{Eric~B
  Schneider}, \bibinfo{person}{N Sriram}, \bibinfo{person}{Valerie~K Scott},
  \bibinfo{person}{Sandra~M Swoboda}, \bibinfo{person}{Cheryl~K Zogg},
  \bibinfo{person}{Nitasha Dhiman}, \bibinfo{person}{Elliott~R Haut},
  \bibinfo{person}{David~T Efron}, \bibinfo{person}{Peter~J Pronovost},
  {et~al\mbox{.}}} \bibinfo{year}{2015}\natexlab{}.
\newblock \showarticletitle{Unconscious race and class biases among registered
  nurses: vignette-based study using implicit association testing}.
\newblock \bibinfo{journal}{\emph{Journal of the American College of Surgeons}}
  \bibinfo{volume}{220}, \bibinfo{number}{6} (\bibinfo{year}{2015}),
  \bibinfo{pages}{1077--1086}.
\newblock


\bibitem[Halchenko et~al\mbox{.}(2021b)]%
        {Halchenko2021}
\bibfield{author}{\bibinfo{person}{Yaroslav Halchenko}, \bibinfo{person}{Kyle
  Meyer}, \bibinfo{person}{Benjamin Poldrack}, \bibinfo{person}{Debanjum
  Solanky}, \bibinfo{person}{Adina Wagner}, \bibinfo{person}{Jason Gors},
  \bibinfo{person}{Dave MacFarlane}, \bibinfo{person}{Dorian Pustina},
  \bibinfo{person}{Vanessa Sochat}, \bibinfo{person}{Satrajit Ghosh},
  \bibinfo{person}{Christian Mönch}, \bibinfo{person}{Christopher Markiewicz},
  \bibinfo{person}{Laura Waite}, \bibinfo{person}{Ilya Shlyakhter},
  \bibinfo{person}{Alejandro de~la Vega}, \bibinfo{person}{Soichi Hayashi},
  \bibinfo{person}{Christian Häusler}, \bibinfo{person}{Jean-Baptiste Poline},
  \bibinfo{person}{Tobias Kadelka}, \bibinfo{person}{Kusti Skytén},
  \bibinfo{person}{Dorota Jarecka}, \bibinfo{person}{David Kennedy},
  \bibinfo{person}{Ted Strauss}, \bibinfo{person}{Matt Cieslak},
  \bibinfo{person}{Peter Vavra}, \bibinfo{person}{Horea-Ioan Ioanas},
  \bibinfo{person}{Robin Schneider}, \bibinfo{person}{Mika Pflüger},
  \bibinfo{person}{James Haxby}, \bibinfo{person}{Simon Eickhoff}, {and}
  \bibinfo{person}{Michael Hanke}.} \bibinfo{year}{2021}\natexlab{b}.
\newblock \showarticletitle{DataLad: distributed system for joint management of
  code, data, and their relationship}.
\newblock \bibinfo{journal}{\emph{Journal of Open Source Software}}
  \bibinfo{volume}{6} (\bibinfo{date}{7} \bibinfo{year}{2021}),
  \bibinfo{pages}{3262}.
\newblock
Issue 63.
\urldef\tempurl%
\url{https://doi.org/10.21105/joss.03262}
\showDOI{\tempurl}


\bibitem[Halchenko et~al\mbox{.}(2021a)]%
        {halchenko2021datalad}
\bibfield{author}{\bibinfo{person}{Yaroslav~O Halchenko}, \bibinfo{person}{Kyle
  Meyer}, \bibinfo{person}{Benjamin Poldrack}, \bibinfo{person}{Debanjum~Singh
  Solanky}, \bibinfo{person}{Adina~S Wagner}, \bibinfo{person}{Jason Gors},
  \bibinfo{person}{Dave MacFarlane}, \bibinfo{person}{Dorian Pustina},
  \bibinfo{person}{Vanessa Sochat}, \bibinfo{person}{Satrajit~S Ghosh},
  {et~al\mbox{.}}} \bibinfo{year}{2021}\natexlab{a}.
\newblock \showarticletitle{DataLad: distributed system for joint management of
  code, data, and their relationship}.
\newblock \bibinfo{journal}{\emph{Journal of Open Source Software}}
  \bibinfo{volume}{6}, \bibinfo{number}{63} (\bibinfo{year}{2021}),
  \bibinfo{pages}{3262}.
\newblock


\bibitem[Heller et~al\mbox{.}(2019)]%
        {heller2019role}
\bibfield{author}{\bibinfo{person}{Nicholas Heller}, \bibinfo{person}{Jack
  Rickman}, \bibinfo{person}{Christopher Weight}, {and}
  \bibinfo{person}{Nikolaos Papanikolopoulos}.}
  \bibinfo{year}{2019}\natexlab{}.
\newblock \showarticletitle{The role of publicly available data in miccai
  papers from 2014 to 2018}. In \bibinfo{booktitle}{\emph{Large-Scale
  Annotation of Biomedical Data and Expert Label Synthesis (MICCAI LABELS)}}.
  Springer, \bibinfo{pages}{70--77}.
\newblock


\bibitem[Hernandez~Petzsche et~al\mbox{.}(2022)]%
        {hernandez2022isles}
\bibfield{author}{\bibinfo{person}{Moritz~R Hernandez~Petzsche},
  \bibinfo{person}{Ezequiel de~la Rosa}, \bibinfo{person}{Uta Hanning},
  \bibinfo{person}{Roland Wiest}, \bibinfo{person}{Waldo Valenzuela},
  \bibinfo{person}{Mauricio Reyes}, \bibinfo{person}{Maria Meyer},
  \bibinfo{person}{Sook-Lei Liew}, \bibinfo{person}{Florian Kofler},
  \bibinfo{person}{Ivan Ezhov}, {et~al\mbox{.}}}
  \bibinfo{year}{2022}\natexlab{}.
\newblock \showarticletitle{ISLES 2022: A multi-center magnetic resonance
  imaging stroke lesion segmentation dataset}.
\newblock \bibinfo{journal}{\emph{Scientific Data}} \bibinfo{volume}{9},
  \bibinfo{number}{1} (\bibinfo{year}{2022}), \bibinfo{pages}{762}.
\newblock


\bibitem[Hernández-Pérez et~al\mbox{.}(2024)]%
        {combalia2019bcn20000}
\bibfield{author}{\bibinfo{person}{Carlos Hernández-Pérez},
  \bibinfo{person}{Marc Combalia}, \bibinfo{person}{Sebastian Podlipnik},
  \bibinfo{person}{Noel C.~F. Codella}, \bibinfo{person}{Veronica Rotemberg},
  \bibinfo{person}{Allan~C. Halpern}, \bibinfo{person}{Ofer Reiter},
  \bibinfo{person}{Cristina Carrera}, \bibinfo{person}{Alicia Barreiro},
  \bibinfo{person}{Brian Helba}, \bibinfo{person}{Susana Puig},
  \bibinfo{person}{Veronica Vilaplana}, {and} \bibinfo{person}{Josep Malvehy}.}
  \bibinfo{year}{2024}\natexlab{}.
\newblock \showarticletitle{{BCN20000: Dermoscopic Lesions in the Wild}}.
\newblock \bibinfo{journal}{\emph{Scientific Data}} \bibinfo{volume}{11},
  \bibinfo{number}{1} (\bibinfo{date}{June} \bibinfo{year}{2024}).
\newblock
\showISSN{2052-4463}
\urldef\tempurl%
\url{https://doi.org/10.1038/s41597-024-03387-w}
\showDOI{\tempurl}


\bibitem[Hoffman et~al\mbox{.}(2016)]%
        {hoffman2016racial}
\bibfield{author}{\bibinfo{person}{Kelly~M Hoffman}, \bibinfo{person}{Sophie
  Trawalter}, \bibinfo{person}{Jordan~R Axt}, {and} \bibinfo{person}{M~Norman
  Oliver}.} \bibinfo{year}{2016}\natexlab{}.
\newblock \showarticletitle{Racial bias in pain assessment and treatment
  recommendations, and false beliefs about biological differences between
  blacks and whites}.
\newblock \bibinfo{journal}{\emph{Proceedings of the National Academy of
  Sciences}} \bibinfo{volume}{113}, \bibinfo{number}{16}
  (\bibinfo{year}{2016}), \bibinfo{pages}{4296--4301}.
\newblock


\bibitem[Holste et~al\mbox{.}(2023)]%
        {holste2023prunecxr}
\bibfield{author}{\bibinfo{person}{Gregory Holste}, \bibinfo{person}{Ziyu
  Jiang}, \bibinfo{person}{Ajay Jaiswal}, \bibinfo{person}{Maria Hanna},
  \bibinfo{person}{Shlomo Minkowitz}, \bibinfo{person}{Alan~C Legasto},
  \bibinfo{person}{Joanna~G Escalon}, \bibinfo{person}{Sharon Steinberger},
  \bibinfo{person}{Mark Bittman}, \bibinfo{person}{Thomas~C Shen},
  {et~al\mbox{.}}} \bibinfo{year}{2023}\natexlab{}.
\newblock \showarticletitle{How Does Pruning Impact Long-Tailed Multi-label
  Medical Image Classifiers?}. In \bibinfo{booktitle}{\emph{Medical Image
  Computing and Computer-Assisted Intervention (MICCAI)}}. Springer,
  \bibinfo{pages}{663--673}.
\newblock


\bibitem[Holste et~al\mbox{.}(2022)]%
        {holste2022longtailcxr}
\bibfield{author}{\bibinfo{person}{Gregory Holste}, \bibinfo{person}{Song
  Wang}, \bibinfo{person}{Ziyu Jiang}, \bibinfo{person}{Thomas~C Shen},
  \bibinfo{person}{George Shih}, \bibinfo{person}{Ronald~M Summers},
  \bibinfo{person}{Yifan Peng}, {and} \bibinfo{person}{Zhangyang Wang}.}
  \bibinfo{year}{2022}\natexlab{}.
\newblock \showarticletitle{Long-tailed classification of thorax diseases on
  chest x-ray: A new benchmark study}. In \bibinfo{booktitle}{\emph{MICCAI
  Workshop on Data Augmentation, Labelling, and Imperfections}}. Springer,
  \bibinfo{pages}{22--32}.
\newblock


\bibitem[Hosseini et~al\mbox{.}(2020)]%
        {hosseini2020tried}
\bibfield{author}{\bibinfo{person}{Mahan Hosseini}, \bibinfo{person}{Michael
  Powell}, \bibinfo{person}{John Collins}, \bibinfo{person}{Chloe
  Callahan-Flintoft}, \bibinfo{person}{William Jones}, \bibinfo{person}{Howard
  Bowman}, {and} \bibinfo{person}{Brad Wyble}.}
  \bibinfo{year}{2020}\natexlab{}.
\newblock \showarticletitle{I tried a bunch of things: The dangers of
  unexpected overfitting in classification of brain data}.
\newblock \bibinfo{journal}{\emph{Neuroscience \& Biobehavioral Reviews}}
  \bibinfo{volume}{119} (\bibinfo{year}{2020}), \bibinfo{pages}{456--467}.
\newblock


\bibitem[HuggingFace(2025)]%
        {huggingFace}
\bibfield{author}{\bibinfo{person}{HuggingFace}.}
  \bibinfo{year}{2025}\natexlab{}.
\newblock \bibinfo{title}{HuggingFace}.
\newblock \bibinfo{howpublished}{https://huggingface.co/}.
\newblock
\newblock
\shownote{Accessed: 2025-01-17}.


\bibitem[Hutchinson et~al\mbox{.}(2021)]%
        {hutchinson2021towards}
\bibfield{author}{\bibinfo{person}{Ben Hutchinson}, \bibinfo{person}{Andrew
  Smart}, \bibinfo{person}{Alex Hanna}, \bibinfo{person}{Emily Denton},
  \bibinfo{person}{Christina Greer}, \bibinfo{person}{Oddur Kjartansson},
  \bibinfo{person}{Parker Barnes}, {and} \bibinfo{person}{Margaret Mitchell}.}
  \bibinfo{year}{2021}\natexlab{}.
\newblock \showarticletitle{Towards accountability for machine learning
  datasets: Practices from software engineering and infrastructure}. In
  \bibinfo{booktitle}{\emph{Fairness, Accountability, and Transparency
  (FAccT)}}. \bibinfo{pages}{560--575}.
\newblock


\bibitem[Ieki et~al\mbox{.}(2022)]%
        {ieki2022deep}
\bibfield{author}{\bibinfo{person}{Hirotaka Ieki}, \bibinfo{person}{Kaoru Ito},
  \bibinfo{person}{Mike Saji}, \bibinfo{person}{Rei Kawakami},
  \bibinfo{person}{Yuji Nagatomo}, \bibinfo{person}{Kaori Takada},
  \bibinfo{person}{Toshiya Kariyasu}, \bibinfo{person}{Haruhiko Machida},
  \bibinfo{person}{Satoshi Koyama}, \bibinfo{person}{Hiroki Yoshida},
  {et~al\mbox{.}}} \bibinfo{year}{2022}\natexlab{}.
\newblock \showarticletitle{Deep learning-based age estimation from chest
  X-rays indicates cardiovascular prognosis}.
\newblock \bibinfo{journal}{\emph{Communications Medicine}}
  \bibinfo{volume}{2}, \bibinfo{number}{1} (\bibinfo{year}{2022}),
  \bibinfo{pages}{159}.
\newblock


\bibitem[Ioannou et~al\mbox{.}(2022)]%
        {ioannou2022study}
\bibfield{author}{\bibinfo{person}{Stefanos Ioannou}, \bibinfo{person}{Hana
  Chockler}, \bibinfo{person}{Alexander Hammers}, \bibinfo{person}{Andrew~P
  King}, {and} \bibinfo{person}{Alzheimer’s Disease~Neuroimaging
  Initiative}.} \bibinfo{year}{2022}\natexlab{}.
\newblock \showarticletitle{A study of demographic bias in CNN-based brain MR
  segmentation}. In \bibinfo{booktitle}{\emph{MICCAI Workshop on Machine
  Learning in Clinical Neuroimaging}}. Springer, \bibinfo{pages}{13--22}.
\newblock


\bibitem[Irvin et~al\mbox{.}(2019)]%
        {irvin2019chexpert}
\bibfield{author}{\bibinfo{person}{Jeremy Irvin}, \bibinfo{person}{Pranav
  Rajpurkar}, \bibinfo{person}{Michael Ko}, \bibinfo{person}{Yifan Yu},
  \bibinfo{person}{Silviana Ciurea-Ilcus}, \bibinfo{person}{Chris Chute},
  \bibinfo{person}{Henrik Marklund}, \bibinfo{person}{Behzad Haghgoo},
  \bibinfo{person}{Robyn Ball}, \bibinfo{person}{Katie Shpanskaya},
  {et~al\mbox{.}}} \bibinfo{year}{2019}\natexlab{}.
\newblock \showarticletitle{Chexpert: A large chest radiograph dataset with
  uncertainty labels and expert comparison}. In \bibinfo{booktitle}{\emph{AAAI
  Conference on Artificial Intelligence}}, Vol.~\bibinfo{volume}{33}.
  \bibinfo{pages}{590--597}.
\newblock


\bibitem[(ISIC)(2024)]%
        {isicarchive}
\bibfield{author}{\bibinfo{person}{The International Skin Imaging~Collaboration
  (ISIC)}.} \bibinfo{year}{2024}\natexlab{}.
\newblock \bibinfo{title}{ISIC archive}.
\newblock \bibinfo{howpublished}{https://www.isic-archive.com/}.
\newblock
\newblock
\shownote{Accessed: 2024-05-22}.


\bibitem[Jeong et~al\mbox{.}(2023)]%
        {jeong2023emory}
\bibfield{author}{\bibinfo{person}{Jiwoong~J Jeong}, \bibinfo{person}{Brianna~L
  Vey}, \bibinfo{person}{Ananth Bhimireddy}, \bibinfo{person}{Thomas Kim},
  \bibinfo{person}{Thiago Santos}, \bibinfo{person}{Ramon Correa},
  \bibinfo{person}{Raman Dutt}, \bibinfo{person}{Marina Mosunjac},
  \bibinfo{person}{Gabriela Oprea-Ilies}, \bibinfo{person}{Geoffrey Smith},
  {et~al\mbox{.}}} \bibinfo{year}{2023}\natexlab{}.
\newblock \showarticletitle{The EMory BrEast imaging Dataset (EMBED): A
  racially diverse, granular dataset of 3.4 million screening and diagnostic
  mammographic images}.
\newblock \bibinfo{journal}{\emph{Radiology: Artificial Intelligence}}
  \bibinfo{volume}{5}, \bibinfo{number}{1} (\bibinfo{year}{2023}),
  \bibinfo{pages}{e220047}.
\newblock


\bibitem[Jieyun and ZhanHong(2023)]%
        {AOP}
\bibfield{author}{\bibinfo{person}{Bai Jieyun} {and} \bibinfo{person}{Ou
  ZhanHong}.} \bibinfo{year}{2023}\natexlab{}.
\newblock \bibinfo{booktitle}{\emph{Pubic {{Symphysis-Fetal Head Segmentation}}
  and {{Angle}} of {{Progression}}}}.
\newblock
\urldef\tempurl%
\url{https://doi.org/10.5281/zenodo.7851339}
\showDOI{\tempurl}


\bibitem[Jim{\'e}nez-S{\'a}nchez et~al\mbox{.}(2024)]%
        {jimenez2024copycats}
\bibfield{author}{\bibinfo{person}{Amelia Jim{\'e}nez-S{\'a}nchez},
  \bibinfo{person}{Natalia-Rozalia Avlona}, \bibinfo{person}{Dovile Juodelyte},
  \bibinfo{person}{Th{\'e}o Sourget}, \bibinfo{person}{Caroline Vang-Larsen},
  \bibinfo{person}{Hubert~Dariusz Zaj{\k{a}}c}, \bibinfo{person}{Anna Rogers},
  {and} \bibinfo{person}{Veronika Cheplygina}.}
  \bibinfo{year}{2024}\natexlab{}.
\newblock \showarticletitle{Copycats: the many lives of a publicly available
  medical imaging dataset}. In \bibinfo{booktitle}{\emph{Neural Information
  Processing Systems (NeurIPS) Datasets and Benchmarks Track}}.
\newblock
\urldef\tempurl%
\url{https://openreview.net/forum?id=X4KImMSIRq}
\showURL{%
\tempurl}


\bibitem[Jim{\'e}nez-S{\'a}nchez et~al\mbox{.}(2023)]%
        {jimenez2023detecting}
\bibfield{author}{\bibinfo{person}{Amelia Jim{\'e}nez-S{\'a}nchez},
  \bibinfo{person}{Dovile Juodelyte}, \bibinfo{person}{Bethany Chamberlain},
  {and} \bibinfo{person}{Veronika Cheplygina}.}
  \bibinfo{year}{2023}\natexlab{}.
\newblock \showarticletitle{{Detecting Shortcuts in Medical Images} - {A Case
  Study in Chest X-rays}}. In \bibinfo{booktitle}{\emph{2023 IEEE 20th
  International Symposium on Biomedical Imaging (ISBI)}}. IEEE,
  \bibinfo{pages}{1--5}.
\newblock


\bibitem[Johnson et~al\mbox{.}(2019)]%
        {johnson2019mimic}
\bibfield{author}{\bibinfo{person}{Alistair~EW Johnson}, \bibinfo{person}{Tom~J
  Pollard}, \bibinfo{person}{Seth~J Berkowitz}, \bibinfo{person}{Nathaniel~R
  Greenbaum}, \bibinfo{person}{Matthew~P Lungren}, \bibinfo{person}{Chih-ying
  Deng}, \bibinfo{person}{Roger~G Mark}, {and} \bibinfo{person}{Steven Horng}.}
  \bibinfo{year}{2019}\natexlab{}.
\newblock \showarticletitle{MIMIC-CXR, a de-identified publicly available
  database of chest radiographs with free-text reports}.
\newblock \bibinfo{journal}{\emph{Scientific Data}} \bibinfo{volume}{6},
  \bibinfo{number}{1} (\bibinfo{year}{2019}), \bibinfo{pages}{317}.
\newblock


\bibitem[Joskowicz et~al\mbox{.}(2019)]%
        {joskowicz2019inter}
\bibfield{author}{\bibinfo{person}{Leo Joskowicz}, \bibinfo{person}{D Cohen},
  \bibinfo{person}{N Caplan}, {and} \bibinfo{person}{Jacob Sosna}.}
  \bibinfo{year}{2019}\natexlab{}.
\newblock \showarticletitle{Inter-observer variability of manual contour
  delineation of structures in CT}.
\newblock \bibinfo{journal}{\emph{European Radiology}}  \bibinfo{volume}{29}
  (\bibinfo{year}{2019}), \bibinfo{pages}{1391--1399}.
\newblock


\bibitem[Juodelyte et~al\mbox{.}(2025)]%
        {juodelyte2024source}
\bibfield{author}{\bibinfo{person}{Dovile Juodelyte}, \bibinfo{person}{Yucheng
  Lu}, \bibinfo{person}{Amelia Jim{\'e}nez-S{\'a}nchez},
  \bibinfo{person}{Sabrina Bottazzi}, \bibinfo{person}{Enzo Ferrante}, {and}
  \bibinfo{person}{Veronika Cheplygina}.} \bibinfo{year}{2025}\natexlab{}.
\newblock \showarticletitle{Source Matters: Source Dataset Impact on Model
  Robustness in Medical Imaging}. In \bibinfo{booktitle}{\emph{Applications of
  Medical Artificial Intelligence}},
  \bibfield{editor}{\bibinfo{person}{Shandong Wu}, \bibinfo{person}{Behrouz
  Shabestari}, {and} \bibinfo{person}{Lei Xing}} (Eds.).
  \bibinfo{publisher}{Springer Nature Switzerland}, \bibinfo{address}{Cham},
  \bibinfo{pages}{105--115}.
\newblock
\showISBNx{978-3-031-82007-6}


\bibitem[Kaggle(2025)]%
        {kaggle}
\bibfield{author}{\bibinfo{person}{Kaggle}.} \bibinfo{year}{2025}\natexlab{}.
\newblock \bibinfo{title}{Kaggle}.
\newblock \bibinfo{howpublished}{https://www.kaggle.com/}.
\newblock
\newblock
\shownote{Accessed: 2025-01-17}.


\bibitem[Kaissis et~al\mbox{.}(2021)]%
        {kaissis2021end}
\bibfield{author}{\bibinfo{person}{Georgios Kaissis},
  \bibinfo{person}{Alexander Ziller}, \bibinfo{person}{Jonathan
  Passerat-Palmbach}, \bibinfo{person}{Th{\'e}o Ryffel},
  \bibinfo{person}{Dmitrii Usynin}, \bibinfo{person}{Andrew Trask},
  \bibinfo{person}{Ion{\'e}sio Lima~Jr}, \bibinfo{person}{Jason Mancuso},
  \bibinfo{person}{Friederike Jungmann}, \bibinfo{person}{Marc-Matthias
  Steinborn}, {et~al\mbox{.}}} \bibinfo{year}{2021}\natexlab{}.
\newblock \showarticletitle{End-to-end privacy preserving deep learning on
  multi-institutional medical imaging}.
\newblock \bibinfo{journal}{\emph{Nature Machine Intelligence}}
  \bibinfo{volume}{3}, \bibinfo{number}{6} (\bibinfo{year}{2021}),
  \bibinfo{pages}{473--484}.
\newblock


\bibitem[Kaplan et~al\mbox{.}(2020)]%
        {kaplan2020scaling}
\bibfield{author}{\bibinfo{person}{Jared Kaplan}, \bibinfo{person}{Sam
  McCandlish}, \bibinfo{person}{Tom Henighan}, \bibinfo{person}{Tom~B Brown},
  \bibinfo{person}{Benjamin Chess}, \bibinfo{person}{Rewon Child},
  \bibinfo{person}{Scott Gray}, \bibinfo{person}{Alec Radford},
  \bibinfo{person}{Jeffrey Wu}, {and} \bibinfo{person}{Dario Amodei}.}
  \bibinfo{year}{2020}\natexlab{}.
\newblock \showarticletitle{Scaling laws for neural language models}.
\newblock \bibinfo{journal}{\emph{arXiv preprint arXiv:2001.08361}}
  (\bibinfo{year}{2020}).
\newblock


\bibitem[Kazerooni et~al\mbox{.}(2024)]%
        {kazerooni2024bratspeds}
\bibfield{author}{\bibinfo{person}{Anahita~Fathi Kazerooni},
  \bibinfo{person}{Nastaran Khalili}, \bibinfo{person}{Xinyang Liu},
  \bibinfo{person}{Debanjan Haldar}, \bibinfo{person}{Zhifan Jiang},
  \bibinfo{person}{Anna Zapaishchykova}, \bibinfo{person}{Julija Pavaine},
  \bibinfo{person}{Lubdha~M Shah}, \bibinfo{person}{Blaise~V Jones},
  \bibinfo{person}{Nakul Sheth}, {et~al\mbox{.}}}
  \bibinfo{year}{2024}\natexlab{}.
\newblock \showarticletitle{Brats-peds: Results of the multi-consortium
  international pediatric brain tumor segmentation challenge 2023}.
\newblock \bibinfo{journal}{\emph{arXiv preprint arXiv:2407.08855}}
  (\bibinfo{year}{2024}).
\newblock


\bibitem[Keenan et~al\mbox{.}(2022)]%
        {Keenan2022}
\bibfield{author}{\bibinfo{person}{Kathryn~E. Keenan}, \bibinfo{person}{Jana~G.
  Delfino}, \bibinfo{person}{Kalina~V. Jordanova}, \bibinfo{person}{Megan~E.
  Poorman}, \bibinfo{person}{Prathyush Chirra}, \bibinfo{person}{Akshay~S.
  Chaudhari}, \bibinfo{person}{Bettina Baessler}, \bibinfo{person}{Jessica
  Winfield}, \bibinfo{person}{Satish~E. Viswanath}, {and}
  \bibinfo{person}{Nandita~M. deSouza}.} \bibinfo{year}{2022}\natexlab{}.
\newblock \bibinfo{title}{Challenges in ensuring the generalizability of image
  quantitation methods for MRI}.
\newblock , \bibinfo{numpages}{2820-2835}~pages.
\newblock
Issue 4.
\showISSN{24734209}
\urldef\tempurl%
\url{https://doi.org/10.1002/mp.15195}
\showDOI{\tempurl}


\bibitem[Kennedy et~al\mbox{.}(2016)]%
        {kennedy2016nitrc}
\bibfield{author}{\bibinfo{person}{David~N Kennedy}, \bibinfo{person}{Christian
  Haselgrove}, \bibinfo{person}{Jon Riehl}, \bibinfo{person}{Nina Preuss},
  {and} \bibinfo{person}{Robert Buccigrossi}.} \bibinfo{year}{2016}\natexlab{}.
\newblock \showarticletitle{The NITRC image repository}.
\newblock \bibinfo{journal}{\emph{NeuroImage}}  \bibinfo{volume}{124}
  (\bibinfo{year}{2016}), \bibinfo{pages}{1069--1073}.
\newblock


\bibitem[Khan et~al\mbox{.}(2021)]%
        {khan2021ophtalmologicalreview}
\bibfield{author}{\bibinfo{person}{Saad~M Khan}, \bibinfo{person}{Xiaoxuan
  Liu}, \bibinfo{person}{Siddharth Nath}, \bibinfo{person}{Edward Korot},
  \bibinfo{person}{Livia Faes}, \bibinfo{person}{Siegfried~K Wagner},
  \bibinfo{person}{Pearse~A Keane}, \bibinfo{person}{Neil~J Sebire},
  \bibinfo{person}{Matthew~J Burton}, {and} \bibinfo{person}{Alastair~K
  Denniston}.} \bibinfo{year}{2021}\natexlab{}.
\newblock \showarticletitle{A global review of publicly available datasets for
  ophthalmological imaging: barriers to access, usability, and
  generalisability}.
\newblock \bibinfo{journal}{\emph{The Lancet Digital Health}}
  \bibinfo{volume}{3}, \bibinfo{number}{1} (\bibinfo{year}{2021}),
  \bibinfo{pages}{e51--e66}.
\newblock
\showISSN{2589-7500}
\urldef\tempurl%
\url{https://doi.org/10.1016/S2589-7500(20)30240-5}
\showDOI{\tempurl}


\bibitem[Koenig et~al\mbox{.}(2020)]%
        {koenig2020oasis}
\bibfield{author}{\bibinfo{person}{Lauren~N. Koenig},
  \bibinfo{person}{Gregory~S. Day}, \bibinfo{person}{Amber Salter},
  \bibinfo{person}{Sarah Keefe}, \bibinfo{person}{Laura~M. Marple},
  \bibinfo{person}{Justin Long}, \bibinfo{person}{Pamela LaMontagne},
  \bibinfo{person}{Parinaz Massoumzadeh}, \bibinfo{person}{B.~Joy Snider},
  \bibinfo{person}{Manasa Kanthamneni}, \bibinfo{person}{Cyrus~A. Raji},
  \bibinfo{person}{Nupur Ghoshal}, \bibinfo{person}{Brian~A. Gordon},
  \bibinfo{person}{Michelle Miller-Thomas}, \bibinfo{person}{John~C. Morris},
  \bibinfo{person}{Joshua~S. Shimony}, {and} \bibinfo{person}{Tammie~L.S.
  Benzinger}.} \bibinfo{year}{2020}\natexlab{}.
\newblock \showarticletitle{{Select Atrophied Regions in Alzheimer disease
  (SARA): An improved volumetric model for identifying Alzheimer disease
  dementia}}.
\newblock \bibinfo{journal}{\emph{NeuroImage: Clinical}}  \bibinfo{volume}{26}
  (\bibinfo{year}{2020}), \bibinfo{pages}{102248}.
\newblock
\showISSN{2213-1582}
\urldef\tempurl%
\url{https://doi.org/10.1016/j.nicl.2020.102248}
\showDOI{\tempurl}


\bibitem[Kushol et~al\mbox{.}(2023)]%
        {Kushol2023}
\bibfield{author}{\bibinfo{person}{Rafsanjany Kushol}, \bibinfo{person}{Pedram
  Parnianpour}, \bibinfo{person}{Alan~H. Wilman}, \bibinfo{person}{Sanjay
  Kalra}, {and} \bibinfo{person}{Yee~Hong Yang}.}
  \bibinfo{year}{2023}\natexlab{}.
\newblock \showarticletitle{Effects of MRI scanner manufacturers in
  classification tasks with deep learning models}.
\newblock \bibinfo{journal}{\emph{Scientific Reports}}  \bibinfo{volume}{13}
  (\bibinfo{date}{12} \bibinfo{year}{2023}).
\newblock
Issue 1.
\showISSN{20452322}
\urldef\tempurl%
\url{https://doi.org/10.1038/s41598-023-43715-5}
\showDOI{\tempurl}


\bibitem[LaMontagne et~al\mbox{.}(2019)]%
        {lamontagne2019oasis}
\bibfield{author}{\bibinfo{person}{Pamela~J LaMontagne},
  \bibinfo{person}{Tammie~LS Benzinger}, \bibinfo{person}{John~C Morris},
  \bibinfo{person}{Sarah Keefe}, \bibinfo{person}{Russ Hornbeck},
  \bibinfo{person}{Chengjie Xiong}, \bibinfo{person}{Elizabeth Grant},
  \bibinfo{person}{Jason Hassenstab}, \bibinfo{person}{Krista Moulder},
  \bibinfo{person}{Andrei~G Vlassenko}, {et~al\mbox{.}}}
  \bibinfo{year}{2019}\natexlab{}.
\newblock \showarticletitle{{OASIS-3}: longitudinal neuroimaging, clinical, and
  cognitive dataset for normal aging and {Alzheimer disease}}.
\newblock \bibinfo{journal}{\emph{medrxiv}} (\bibinfo{year}{2019}),
  \bibinfo{pages}{2019--12}.
\newblock


\bibitem[Lang et~al\mbox{.}(2024)]%
        {lang2024using}
\bibfield{author}{\bibinfo{person}{Oran Lang}, \bibinfo{person}{Doron
  Yaya-Stupp}, \bibinfo{person}{Ilana Traynis}, \bibinfo{person}{Heather
  Cole-Lewis}, \bibinfo{person}{Chloe~R Bennett}, \bibinfo{person}{Courtney~R
  Lyles}, \bibinfo{person}{Charles Lau}, \bibinfo{person}{Michal Irani},
  \bibinfo{person}{Christopher Semturs}, \bibinfo{person}{Dale~R Webster},
  {et~al\mbox{.}}} \bibinfo{year}{2024}\natexlab{}.
\newblock \showarticletitle{Using generative AI to investigate medical imagery
  models and datasets}.
\newblock \bibinfo{journal}{\emph{EBioMedicine}}  \bibinfo{volume}{102}
  (\bibinfo{year}{2024}).
\newblock


\bibitem[Larrazabal et~al\mbox{.}(2020)]%
        {larrazabal2020gender}
\bibfield{author}{\bibinfo{person}{Agostina~J Larrazabal},
  \bibinfo{person}{Nicol{\'a}s Nieto}, \bibinfo{person}{Victoria Peterson},
  \bibinfo{person}{Diego~H Milone}, {and} \bibinfo{person}{Enzo Ferrante}.}
  \bibinfo{year}{2020}\natexlab{}.
\newblock \showarticletitle{Gender imbalance in medical imaging datasets
  produces biased classifiers for computer-aided diagnosis}.
\newblock \bibinfo{journal}{\emph{Proceedings of the National Academy of
  Sciences}} \bibinfo{volume}{117}, \bibinfo{number}{23}
  (\bibinfo{year}{2020}), \bibinfo{pages}{12592--12594}.
\newblock


\bibitem[Le~Guellec et~al\mbox{.}(2024)]%
        {le2024performance}
\bibfield{author}{\bibinfo{person}{Bastien Le~Guellec},
  \bibinfo{person}{Alexandre Lef{\`e}vre}, \bibinfo{person}{Charlotte Geay},
  \bibinfo{person}{Lucas Shorten}, \bibinfo{person}{Cyril Bruge},
  \bibinfo{person}{Lotfi Hacein-Bey}, \bibinfo{person}{Philippe Amouyel},
  \bibinfo{person}{Jean-Pierre Pruvo}, \bibinfo{person}{Gr{\'e}gory
  Kuchcinski}, {and} \bibinfo{person}{Aghiles Hamroun}.}
  \bibinfo{year}{2024}\natexlab{}.
\newblock \showarticletitle{Performance of an open-source large language model
  in extracting information from free-text radiology reports}.
\newblock \bibinfo{journal}{\emph{Radiology: Artificial Intelligence}}
  (\bibinfo{year}{2024}), \bibinfo{pages}{e230364}.
\newblock


\bibitem[Leavy et~al\mbox{.}(2021)]%
        {leavy2021ethical}
\bibfield{author}{\bibinfo{person}{Susan Leavy}, \bibinfo{person}{Eugenia
  Siapera}, {and} \bibinfo{person}{Barry O'Sullivan}.}
  \bibinfo{year}{2021}\natexlab{}.
\newblock \showarticletitle{Ethical data curation for ai: An approach based on
  feminist epistemology and critical theories of race}. In
  \bibinfo{booktitle}{\emph{AI, Ethics, and Society (AIES)}}.
  \bibinfo{publisher}{AAAI/ACM}, \bibinfo{pages}{695--703}.
\newblock


\bibitem[Li et~al\mbox{.}(2023)]%
        {li2023systematic}
\bibfield{author}{\bibinfo{person}{Johann Li}, \bibinfo{person}{Guangming Zhu},
  \bibinfo{person}{Cong Hua}, \bibinfo{person}{Mingtao Feng},
  \bibinfo{person}{Basheer Bennamoun}, \bibinfo{person}{Ping Li},
  \bibinfo{person}{Xiaoyuan Lu}, \bibinfo{person}{Juan Song},
  \bibinfo{person}{Peiyi Shen}, \bibinfo{person}{Xu Xu}, {et~al\mbox{.}}}
  \bibinfo{year}{2023}\natexlab{}.
\newblock \showarticletitle{A systematic collection of medical image datasets
  for deep learning}.
\newblock \bibinfo{journal}{\emph{Comput. Surveys}} \bibinfo{volume}{56},
  \bibinfo{number}{5} (\bibinfo{year}{2023}), \bibinfo{pages}{1--51}.
\newblock


\bibitem[Li et~al\mbox{.}(2016)]%
        {Li2016}
\bibfield{author}{\bibinfo{person}{Xiangrui Li}, \bibinfo{person}{Paul~S.
  Morgan}, \bibinfo{person}{John Ashburner}, \bibinfo{person}{Jolinda Smith},
  {and} \bibinfo{person}{Christopher Rorden}.} \bibinfo{year}{2016}\natexlab{}.
\newblock \showarticletitle{The first step for neuroimaging data analysis:
  DICOM to NIfTI conversion}.
\newblock \bibinfo{journal}{\emph{Journal of Neuroscience Methods}}
  \bibinfo{volume}{264} (\bibinfo{date}{5} \bibinfo{year}{2016}),
  \bibinfo{pages}{47--56}.
\newblock
\showISSN{1872678X}
\urldef\tempurl%
\url{https://doi.org/10.1016/j.jneumeth.2016.03.001}
\showDOI{\tempurl}


\bibitem[Liao et~al\mbox{.}(2021)]%
        {liao2021learning}
\bibfield{author}{\bibinfo{person}{Thomas Liao}, \bibinfo{person}{Rohan Taori},
  \bibinfo{person}{Inioluwa~Deborah Raji}, {and} \bibinfo{person}{Ludwig
  Schmidt}.} \bibinfo{year}{2021}\natexlab{}.
\newblock \showarticletitle{Are we learning yet? a meta review of evaluation
  failures across machine learning}. In \bibinfo{booktitle}{\emph{Neural
  Information Processing Systems (NeurIPS) Datasets and Benchmarks Track}}.
\newblock


\bibitem[Liew et~al\mbox{.}(2022)]%
        {Liew2022}
\bibfield{author}{\bibinfo{person}{Sook~Lei Liew}, \bibinfo{person}{Bethany~P.
  Lo}, \bibinfo{person}{Miranda~R. Donnelly}, \bibinfo{person}{Artemis
  Zavaliangos-Petropulu}, \bibinfo{person}{Jessica~N. Jeong},
  \bibinfo{person}{Giuseppe Barisano}, \bibinfo{person}{Alexandre Hutton},
  \bibinfo{person}{Julia~P. Simon}, \bibinfo{person}{Julia~M. Juliano},
  \bibinfo{person}{Anisha Suri}, \bibinfo{person}{Zhizhuo Wang},
  \bibinfo{person}{Aisha Abdullah}, \bibinfo{person}{Jun Kim},
  \bibinfo{person}{Tyler Ard}, \bibinfo{person}{Nerisa Banaj},
  \bibinfo{person}{Michael~R. Borich}, \bibinfo{person}{Lara~A. Boyd},
  \bibinfo{person}{Amy Brodtmann}, \bibinfo{person}{Cathrin~M. Buetefisch},
  \bibinfo{person}{Lei Cao}, \bibinfo{person}{Jessica~M. Cassidy},
  \bibinfo{person}{Valentina Ciullo}, \bibinfo{person}{Adriana~B. Conforto},
  \bibinfo{person}{Steven~C. Cramer}, \bibinfo{person}{Rosalia Dacosta-Aguayo},
  \bibinfo{person}{Ezequiel de~la Rosa}, \bibinfo{person}{Martin Domin},
  \bibinfo{person}{Adrienne~N. Dula}, \bibinfo{person}{Wuwei Feng},
  \bibinfo{person}{Alexandre~R. Franco}, \bibinfo{person}{Fatemeh Geranmayeh},
  \bibinfo{person}{Alexandre Gramfort}, \bibinfo{person}{Chris~M. Gregory},
  \bibinfo{person}{Colleen~A. Hanlon}, \bibinfo{person}{Brenton~G. Hordacre},
  \bibinfo{person}{Steven~A. Kautz}, \bibinfo{person}{Mohamed~Salah Khlif},
  \bibinfo{person}{Hosung Kim}, \bibinfo{person}{Jan~S. Kirschke},
  \bibinfo{person}{Jingchun Liu}, \bibinfo{person}{Martin Lotze},
  \bibinfo{person}{Bradley~J. MacIntosh}, \bibinfo{person}{Maria Mataró},
  \bibinfo{person}{Feroze~B. Mohamed}, \bibinfo{person}{Jan~E. Nordvik},
  \bibinfo{person}{Gilsoon Park}, \bibinfo{person}{Amy Pienta},
  \bibinfo{person}{Fabrizio Piras}, \bibinfo{person}{Shane~M. Redman},
  \bibinfo{person}{Kate~P. Revill}, \bibinfo{person}{Mauricio Reyes},
  \bibinfo{person}{Andrew~D. Robertson}, \bibinfo{person}{Na~Jin Seo},
  \bibinfo{person}{Surjo~R. Soekadar}, \bibinfo{person}{Gianfranco Spalletta},
  \bibinfo{person}{Alison Sweet}, \bibinfo{person}{Maria Telenczuk},
  \bibinfo{person}{Gregory Thielman}, \bibinfo{person}{Lars~T. Westlye},
  \bibinfo{person}{Carolee~J. Winstein}, \bibinfo{person}{George~F.
  Wittenberg}, \bibinfo{person}{Kristin~A. Wong}, {and}
  \bibinfo{person}{Chunshui Yu}.} \bibinfo{year}{2022}\natexlab{}.
\newblock \showarticletitle{A large, curated, open-source stroke neuroimaging
  dataset to improve lesion segmentation algorithms}.
\newblock \bibinfo{journal}{\emph{Scientific Data}}  \bibinfo{volume}{9}
  (\bibinfo{date}{12} \bibinfo{year}{2022}).
\newblock
Issue 1.
\showISSN{20524463}
\urldef\tempurl%
\url{https://doi.org/10.1038/s41597-022-01401-7}
\showDOI{\tempurl}


\bibitem[Lin et~al\mbox{.}(2016)]%
        {lin2016scribblesup}
\bibfield{author}{\bibinfo{person}{Di Lin}, \bibinfo{person}{Jifeng Dai},
  \bibinfo{person}{Jiaya Jia}, \bibinfo{person}{Kaiming He}, {and}
  \bibinfo{person}{Jian Sun}.} \bibinfo{year}{2016}\natexlab{}.
\newblock \showarticletitle{Scribblesup: Scribble-supervised convolutional
  networks for semantic segmentation}. In \bibinfo{booktitle}{\emph{Computer
  Vision and Pattern Recognition (CVPR)}}. \bibinfo{pages}{3159--3167}.
\newblock


\bibitem[Lin et~al\mbox{.}(2024)]%
        {lin2024shortcut}
\bibfield{author}{\bibinfo{person}{Manxi Lin}, \bibinfo{person}{Nina Weng},
  \bibinfo{person}{Kamil Mikolaj}, \bibinfo{person}{Zahra Bashir},
  \bibinfo{person}{Morten~BS Svendsen}, \bibinfo{person}{Martin~G Tolsgaard},
  \bibinfo{person}{Anders~N Christensen}, {and} \bibinfo{person}{Aasa
  Feragen}.} \bibinfo{year}{2024}\natexlab{}.
\newblock \showarticletitle{Shortcut learning in medical image segmentation}.
  In \bibinfo{booktitle}{\emph{Medical Image Computing and Computer-Assisted
  Intervention (MICCAI)}}. Springer, \bibinfo{pages}{623--633}.
\newblock


\bibitem[Litjens et~al\mbox{.}(2017)]%
        {litjens2017survey}
\bibfield{author}{\bibinfo{person}{Geert Litjens}, \bibinfo{person}{Thijs
  Kooi}, \bibinfo{person}{Babak~Ehteshami Bejnordi}, \bibinfo{person}{Arnaud
  Arindra~Adiyoso Setio}, \bibinfo{person}{Francesco Ciompi},
  \bibinfo{person}{Mohsen Ghafoorian}, \bibinfo{person}{Jeroen~AWM van~der
  Laak}, \bibinfo{person}{Bram Van~Ginneken}, {and} \bibinfo{person}{Clara~I
  S{\'a}nchez}.} \bibinfo{year}{2017}\natexlab{}.
\newblock \showarticletitle{A survey on deep learning in medical image
  analysis}.
\newblock \bibinfo{journal}{\emph{Medical Image Analysis}}
  \bibinfo{volume}{42} (\bibinfo{year}{2017}), \bibinfo{pages}{60--88}.
\newblock


\bibitem[Liu et~al\mbox{.}(2022)]%
        {liu2022weakly}
\bibfield{author}{\bibinfo{person}{Xiaoming Liu}, \bibinfo{person}{Quan Yuan},
  \bibinfo{person}{Yaozong Gao}, \bibinfo{person}{Kelei He},
  \bibinfo{person}{Shuo Wang}, \bibinfo{person}{Xiao Tang},
  \bibinfo{person}{Jinshan Tang}, {and} \bibinfo{person}{Dinggang Shen}.}
  \bibinfo{year}{2022}\natexlab{}.
\newblock \showarticletitle{Weakly supervised segmentation of COVID19 infection
  with scribble annotation on CT images}.
\newblock \bibinfo{journal}{\emph{Pattern Recognition}}  \bibinfo{volume}{122}
  (\bibinfo{year}{2022}), \bibinfo{pages}{108341}.
\newblock


\bibitem[Longpre et~al\mbox{.}(2023)]%
        {longpre2023dataprovenance}
\bibfield{author}{\bibinfo{person}{Shayne Longpre}, \bibinfo{person}{Robert
  Mahari}, \bibinfo{person}{Anthony Chen}, \bibinfo{person}{Naana Obeng-Marnu},
  \bibinfo{person}{Damien Sileo}, \bibinfo{person}{William Brannon},
  \bibinfo{person}{Niklas Muennighoff}, \bibinfo{person}{Nathan Khazam},
  \bibinfo{person}{Jad Kabbara}, \bibinfo{person}{Kartik Perisetla},
  {et~al\mbox{.}}} \bibinfo{year}{2023}\natexlab{}.
\newblock \showarticletitle{{The data provenance initiative: A large scale
  audit of dataset licensing \& attribution in AI}}.
\newblock \bibinfo{journal}{\emph{arXiv preprint arXiv:2310.16787}}
  (\bibinfo{year}{2023}).
\newblock


\bibitem[Maier-Hein et~al\mbox{.}(2014)]%
        {maier2014can}
\bibfield{author}{\bibinfo{person}{Lena Maier-Hein}, \bibinfo{person}{Sven
  Mersmann}, \bibinfo{person}{Daniel Kondermann}, \bibinfo{person}{Sebastian
  Bodenstedt}, \bibinfo{person}{Alexandro Sanchez}, \bibinfo{person}{Christian
  Stock}, \bibinfo{person}{Hannes~Gotz Kenngott}, \bibinfo{person}{Mathias
  Eisenmann}, {and} \bibinfo{person}{Stefanie Speidel}.}
  \bibinfo{year}{2014}\natexlab{}.
\newblock \showarticletitle{Can Masses of Non-Experts Train Highly Accurate
  Image Classifiers?}
\newblock In \bibinfo{booktitle}{\emph{Medical Image Computing and
  Computer-Assisted Intervention (MICCAI)}}. \bibinfo{publisher}{Springer},
  \bibinfo{pages}{438--445}.
\newblock


\bibitem[Maier-Hein et~al\mbox{.}(2022)]%
        {maier2022metrics}
\bibfield{author}{\bibinfo{person}{Lena Maier-Hein}, \bibinfo{person}{Annika
  Reinke}, \bibinfo{person}{Evangelia Christodoulou}, \bibinfo{person}{Ben
  Glocker}, \bibinfo{person}{Patrick Godau}, \bibinfo{person}{Fabian Isensee},
  \bibinfo{person}{Jens Kleesiek}, \bibinfo{person}{Michal Kozubek},
  \bibinfo{person}{Mauricio Reyes}, \bibinfo{person}{Michael~A Riegler},
  {et~al\mbox{.}}} \bibinfo{year}{2022}\natexlab{}.
\newblock \showarticletitle{Metrics reloaded: Pitfalls and recommendations for
  image analysis validation}.
\newblock \bibinfo{journal}{\emph{arXiv preprint arXiv:2206.01653}}
  (\bibinfo{year}{2022}).
\newblock


\bibitem[Maqsood and Dama{\v{s}}evi{\v{c}}ius(2023)]%
        {maqsood2023multiclass}
\bibfield{author}{\bibinfo{person}{Sarmad Maqsood} {and}
  \bibinfo{person}{Robertas Dama{\v{s}}evi{\v{c}}ius}.}
  \bibinfo{year}{2023}\natexlab{}.
\newblock \showarticletitle{Multiclass skin lesion localization and
  classification using deep learning based features fusion and selection
  framework for smart healthcare}.
\newblock \bibinfo{journal}{\emph{Neural Networks}}  \bibinfo{volume}{160}
  (\bibinfo{year}{2023}), \bibinfo{pages}{238--258}.
\newblock


\bibitem[Marcus et~al\mbox{.}(2010)]%
        {marcus2010open}
\bibfield{author}{\bibinfo{person}{Daniel~S Marcus}, \bibinfo{person}{Anthony~F
  Fotenos}, \bibinfo{person}{John~G Csernansky}, \bibinfo{person}{John~C
  Morris}, {and} \bibinfo{person}{Randy~L Buckner}.}
  \bibinfo{year}{2010}\natexlab{}.
\newblock \showarticletitle{Open access series of imaging studies: longitudinal
  MRI data in nondemented and demented older adults}.
\newblock \bibinfo{journal}{\emph{{Journal of Cognitive Neuroscience}}}
  \bibinfo{volume}{22}, \bibinfo{number}{12} (\bibinfo{year}{2010}),
  \bibinfo{pages}{2677--2684}.
\newblock


\bibitem[Marcus et~al\mbox{.}(2007)]%
        {marcus2007oasis}
\bibfield{author}{\bibinfo{person}{Daniel~S. Marcus}, \bibinfo{person}{Tracy~H.
  Wang}, \bibinfo{person}{Jamie Parker}, \bibinfo{person}{John~G. Csernansky},
  \bibinfo{person}{John~C. Morris}, {and} \bibinfo{person}{Randy~L. Buckner}.}
  \bibinfo{year}{2007}\natexlab{}.
\newblock \showarticletitle{{Open Access Series of Imaging Studies (OASIS):
  Cross-sectional MRI Data in Young, Middle Aged, Nondemented, and Demented
  Older Adults}}.
\newblock \bibinfo{journal}{\emph{Journal of Cognitive Neuroscience}}
  \bibinfo{volume}{19}, \bibinfo{number}{9} (\bibinfo{date}{09}
  \bibinfo{year}{2007}), \bibinfo{pages}{1498--1507}.
\newblock
\showISSN{0898-929X}
\urldef\tempurl%
\url{https://doi.org/10.1162/jocn.2007.19.9.1498}
\showDOI{\tempurl}
\showeprint{https://direct.mit.edu/jocn/article-pdf/19/9/1498/1936514/jocn.2007.19.9.1498.pdf}


\bibitem[Markiewicz et~al\mbox{.}(2021)]%
        {markiewicz2021openneuro}
\bibfield{author}{\bibinfo{person}{Christopher~J. Markiewicz},
  \bibinfo{person}{Krzysztof~J. Gorgolewski}, \bibinfo{person}{Franklin
  Feingold}, \bibinfo{person}{Ross Blair}, \bibinfo{person}{Yaroslav~O.
  Halchenko}, \bibinfo{person}{Eric Miller}, \bibinfo{person}{Nell Hardcastle},
  \bibinfo{person}{Joe Wexler}, \bibinfo{person}{Oscar Esteban},
  \bibinfo{person}{Mathias Goncavles}, \bibinfo{person}{Anita Jwa}, {and}
  \bibinfo{person}{Russell Poldrack}.} \bibinfo{year}{2021}\natexlab{}.
\newblock \showarticletitle{The OpenNeuro resource for sharing of neuroscience
  data}.
\newblock \bibinfo{journal}{\emph{eLife}}  \bibinfo{volume}{10}
  (\bibinfo{date}{10} \bibinfo{year}{2021}).
\newblock
\showISSN{2050084X}
\urldef\tempurl%
\url{https://doi.org/10.7554/eLife.71774}
\showDOI{\tempurl}


\bibitem[Mei et~al\mbox{.}(2022)]%
        {mei2022radimagenet}
\bibfield{author}{\bibinfo{person}{Xueyan Mei}, \bibinfo{person}{Zelong Liu},
  \bibinfo{person}{Philip~M Robson}, \bibinfo{person}{Brett Marinelli},
  \bibinfo{person}{Mingqian Huang}, \bibinfo{person}{Amish Doshi},
  \bibinfo{person}{Adam Jacobi}, \bibinfo{person}{Chendi Cao},
  \bibinfo{person}{Katherine~E Link}, \bibinfo{person}{Thomas Yang},
  {et~al\mbox{.}}} \bibinfo{year}{2022}\natexlab{}.
\newblock \showarticletitle{RadImageNet: an open radiologic deep learning
  research dataset for effective transfer learning}.
\newblock \bibinfo{journal}{\emph{Radiology: Artificial Intelligence}}
  \bibinfo{volume}{4}, \bibinfo{number}{5} (\bibinfo{year}{2022}),
  \bibinfo{pages}{e210315}.
\newblock


\bibitem[Miceli et~al\mbox{.}(2022)]%
        {miceli2022studying}
\bibfield{author}{\bibinfo{person}{Milagros Miceli}, \bibinfo{person}{Julian
  Posada}, {and} \bibinfo{person}{Tianling Yang}.}
  \bibinfo{year}{2022}\natexlab{}.
\newblock \showarticletitle{Studying up machine learning data: Why talk about
  bias when we mean power?}
\newblock \bibinfo{journal}{\emph{Proceedings of the ACM on Human-Computer
  Interaction}} \bibinfo{volume}{6}, \bibinfo{number}{GROUP}
  (\bibinfo{year}{2022}), \bibinfo{pages}{1--14}.
\newblock


\bibitem[Mikolaj et~al\mbox{.}(2025)]%
        {mikolaj2025}
\bibfield{author}{\bibinfo{person}{Kamil~Wojciech Mikolaj},
  \bibinfo{person}{Anders~Nymark Christensen}, \bibinfo{person}{Caroline~Amalie
  Taksøe-Vester}, \bibinfo{person}{Aasa Feragen}, \bibinfo{person}{Olav~Bjørn
  Petersen}, \bibinfo{person}{Manxi Lin}, \bibinfo{person}{Mads Nielsen},
  \bibinfo{person}{Morten Bo~Søndergaard Svendsen}, {and}
  \bibinfo{person}{Martin~Grønnebæk Tolsgaard}.}
  \bibinfo{year}{2025}\natexlab{}.
\newblock \showarticletitle{Predicting Abnormal Fetal Growth Using Deep
  Learning}.
\newblock \bibinfo{journal}{\emph{npj Digital Medicine, in press}}
  (\bibinfo{year}{2025}).
\newblock


\bibitem[Mirikharaji et~al\mbox{.}(2023)]%
        {mirikharaji2023survey}
\bibfield{author}{\bibinfo{person}{Zahra Mirikharaji}, \bibinfo{person}{Kumar
  Abhishek}, \bibinfo{person}{Alceu Bissoto}, \bibinfo{person}{Catarina
  Barata}, \bibinfo{person}{Sandra Avila}, \bibinfo{person}{Eduardo Valle},
  \bibinfo{person}{M~Emre Celebi}, {and} \bibinfo{person}{Ghassan Hamarneh}.}
  \bibinfo{year}{2023}\natexlab{}.
\newblock \showarticletitle{A survey on deep learning for skin lesion
  segmentation}.
\newblock \bibinfo{journal}{\emph{Medical Image Analysis}}
  \bibinfo{volume}{88} (\bibinfo{year}{2023}), \bibinfo{pages}{102863}.
\newblock


\bibitem[Moons et~al\mbox{.}(2019)]%
        {moons2019probast}
\bibfield{author}{\bibinfo{person}{Karel~GM Moons}, \bibinfo{person}{Robert~F
  Wolff}, \bibinfo{person}{Richard~D Riley}, \bibinfo{person}{Penny~F Whiting},
  \bibinfo{person}{Marie Westwood}, \bibinfo{person}{Gary~S Collins},
  \bibinfo{person}{Johannes~B Reitsma}, \bibinfo{person}{Jos Kleijnen}, {and}
  \bibinfo{person}{Sue Mallett}.} \bibinfo{year}{2019}\natexlab{}.
\newblock \showarticletitle{PROBAST: a tool to assess risk of bias and
  applicability of prediction model studies: explanation and elaboration}.
\newblock \bibinfo{journal}{\emph{Annals of Internal Medicine}}
  \bibinfo{volume}{170}, \bibinfo{number}{1} (\bibinfo{year}{2019}),
  \bibinfo{pages}{W1--W33}.
\newblock


\bibitem[Mukherjee et~al\mbox{.}(2023)]%
        {mukherjee2023feasibility}
\bibfield{author}{\bibinfo{person}{Pritam Mukherjee}, \bibinfo{person}{Benjamin
  Hou}, \bibinfo{person}{Ricardo~B Lanfredi}, {and} \bibinfo{person}{Ronald~M
  Summers}.} \bibinfo{year}{2023}\natexlab{}.
\newblock \showarticletitle{Feasibility of using the privacy-preserving large
  language model Vicuna for labeling radiology reports}.
\newblock \bibinfo{journal}{\emph{Radiology}} \bibinfo{volume}{309},
  \bibinfo{number}{1} (\bibinfo{year}{2023}), \bibinfo{pages}{e231147}.
\newblock


\bibitem[Muller et~al\mbox{.}(2021)]%
        {Muller2021}
\bibfield{author}{\bibinfo{person}{Michael Muller},
  \bibinfo{person}{Christine~T. Wolf}, \bibinfo{person}{Josh Andres},
  \bibinfo{person}{Michael Desmond}, \bibinfo{person}{Narendra~Nath Joshi},
  \bibinfo{person}{Zahra Ashktorab}, \bibinfo{person}{Aabhas Sharma},
  \bibinfo{person}{Kristina Brimijoin}, \bibinfo{person}{Qian Pan},
  \bibinfo{person}{Evelyn Duesterwald}, {and} \bibinfo{person}{Casey Dugan}.}
  \bibinfo{year}{2021}\natexlab{}.
\newblock \showarticletitle{Designing Ground Truth and the Social Life of
  Labels}. In \bibinfo{booktitle}{\emph{CHI Conference on Human Factors in
  Computing Systems}} (New York, NY, USA). \bibinfo{publisher}{ACM},
  \bibinfo{pages}{1--16}.
\newblock
\showISBNx{9781450380966}
\urldef\tempurl%
\url{https://doi.org/10.1145/3411764.3445402}
\showDOI{\tempurl}


\bibitem[Naqvi et~al\mbox{.}(2023)]%
        {naqvi2023skin}
\bibfield{author}{\bibinfo{person}{Maryam Naqvi}, \bibinfo{person}{Syed~Qasim
  Gilani}, \bibinfo{person}{Tehreem Syed}, \bibinfo{person}{Oge Marques}, {and}
  \bibinfo{person}{Hee-Cheol Kim}.} \bibinfo{year}{2023}\natexlab{}.
\newblock \showarticletitle{Skin cancer detection using deep learning—a
  review}.
\newblock \bibinfo{journal}{\emph{Diagnostics}} \bibinfo{volume}{13},
  \bibinfo{number}{11} (\bibinfo{year}{2023}), \bibinfo{pages}{1911}.
\newblock


\bibitem[Niso et~al\mbox{.}(2022)]%
        {niso2022}
\bibfield{author}{\bibinfo{person}{Guiomar Niso}, \bibinfo{person}{Rotem
  Botvinik-Nezer}, \bibinfo{person}{Stefan Appelhoff},
  \bibinfo{person}{Alejandro {De La Vega}}, \bibinfo{person}{Oscar Esteban},
  \bibinfo{person}{Joset~A. Etzel}, \bibinfo{person}{Karolina Finc},
  \bibinfo{person}{Melanie Ganz}, \bibinfo{person}{Rémi Gau},
  \bibinfo{person}{Yaroslav~O. Halchenko}, \bibinfo{person}{Peer Herholz},
  \bibinfo{person}{Agah Karakuzu}, \bibinfo{person}{David~B. Keator},
  \bibinfo{person}{Christopher~J. Markiewicz}, \bibinfo{person}{Camille
  Maumet}, \bibinfo{person}{Cyril~R. Pernet}, \bibinfo{person}{Franco
  Pestilli}, \bibinfo{person}{Nazek Queder}, \bibinfo{person}{Tina Schmitt},
  \bibinfo{person}{Weronika Sójka}, \bibinfo{person}{Adina~S. Wagner},
  \bibinfo{person}{Kirstie~J. Whitaker}, {and} \bibinfo{person}{Jochem~W.
  Rieger}.} \bibinfo{year}{2022}\natexlab{}.
\newblock \showarticletitle{Open and reproducible neuroimaging: From study
  inception to publication}.
\newblock \bibinfo{journal}{\emph{NeuroImage}}  \bibinfo{volume}{263}
  (\bibinfo{year}{2022}), \bibinfo{pages}{119623}.
\newblock
\showISSN{1053-8119}
\urldef\tempurl%
\url{https://doi.org/10.1016/j.neuroimage.2022.119623}
\showDOI{\tempurl}


\bibitem[Oakden-Rayner(2020)]%
        {oakden2020exploring}
\bibfield{author}{\bibinfo{person}{Lauren Oakden-Rayner}.}
  \bibinfo{year}{2020}\natexlab{}.
\newblock \showarticletitle{Exploring Large-scale Public Medical Image
  Datasets}.
\newblock \bibinfo{journal}{\emph{Academic Radiology}} \bibinfo{volume}{27},
  \bibinfo{number}{1} (\bibinfo{year}{2020}), \bibinfo{pages}{106--112}.
\newblock


\bibitem[Oakden-Rayner(2025)]%
        {oakdenrayner2019xraysimpressions}
\bibfield{author}{\bibinfo{person}{Lauren Oakden-Rayner}.}
  \bibinfo{year}{2025}\natexlab{}.
\newblock \bibinfo{title}{Half a million x-rays! First impressions of the
  Stanford and MIT chest x-ray datasets}.
\newblock
  \bibinfo{howpublished}{https://laurenoakdenrayner.com/2019/02/25/half-a-million-x-rays-first-impressions-of-the-stanford-and-mit-chest-x-ray-datasets/}.
\newblock
\newblock
\shownote{Accessed: 2025-01-07}.


\bibitem[Oakden-Rayner et~al\mbox{.}(2020)]%
        {oakden2020hidden}
\bibfield{author}{\bibinfo{person}{Luke Oakden-Rayner}, \bibinfo{person}{Jared
  Dunnmon}, \bibinfo{person}{Gustavo Carneiro}, {and}
  \bibinfo{person}{Christopher R{\'e}}.} \bibinfo{year}{2020}\natexlab{}.
\newblock \showarticletitle{Hidden stratification causes clinically meaningful
  failures in machine learning for medical imaging}. In
  \bibinfo{booktitle}{\emph{Conference on Health, Inference, and Learning
  (CHIL)}}. \bibinfo{publisher}{ACM}, \bibinfo{pages}{151--159}.
\newblock


\bibitem[Oakden-Rayner et~al\mbox{.}(2022)]%
        {oakden2022validation}
\bibfield{author}{\bibinfo{person}{Lauren Oakden-Rayner},
  \bibinfo{person}{William Gale}, \bibinfo{person}{Thomas~A Bonham},
  \bibinfo{person}{Matthew~P Lungren}, \bibinfo{person}{Gustavo Carneiro},
  \bibinfo{person}{Andrew~P Bradley}, {and} \bibinfo{person}{Lyle~J Palmer}.}
  \bibinfo{year}{2022}\natexlab{}.
\newblock \showarticletitle{Validation and algorithmic audit of a deep learning
  system for the detection of proximal femoral fractures in patients in the
  emergency department: a diagnostic accuracy study}.
\newblock \bibinfo{journal}{\emph{The Lancet Digital Health}}
  \bibinfo{volume}{4}, \bibinfo{number}{5} (\bibinfo{year}{2022}),
  \bibinfo{pages}{e351--e358}.
\newblock


\bibitem[Oca{\~n}a-Tienda et~al\mbox{.}(2023)]%
        {ocana2023comprehensive}
\bibfield{author}{\bibinfo{person}{Beatriz Oca{\~n}a-Tienda},
  \bibinfo{person}{Juli{\'a}n P{\'e}rez-Beteta}, \bibinfo{person}{Jos{\'e}~D
  Villanueva-Garc{\'\i}a}, \bibinfo{person}{Jos{\'e}~A Romero-Rosales},
  \bibinfo{person}{David Molina-Garc{\'\i}a}, \bibinfo{person}{Yannick Suter},
  \bibinfo{person}{Beatriz Asenjo}, \bibinfo{person}{David Albillo},
  \bibinfo{person}{Ana Ortiz~de Mendivil}, \bibinfo{person}{Luis~A
  P{\'e}rez-Romasanta}, {et~al\mbox{.}}} \bibinfo{year}{2023}\natexlab{}.
\newblock \showarticletitle{A comprehensive dataset of annotated brain
  metastasis MR images with clinical and radiomic data}.
\newblock \bibinfo{journal}{\emph{Scientific Data}} \bibinfo{volume}{10},
  \bibinfo{number}{1} (\bibinfo{year}{2023}), \bibinfo{pages}{208}.
\newblock


\bibitem[Ogasawara(2009)]%
        {ogasawaraVariation2009}
\bibfield{author}{\bibinfo{person}{Keith~K. Ogasawara}.}
  \bibinfo{year}{2009}\natexlab{}.
\newblock \showarticletitle{Variation in Fetal Ultrasound Biometry Based on
  Differences in Fetal Ethnicity}.
\newblock \bibinfo{journal}{\emph{American Journal of Obstetrics and
  Gynecology}} \bibinfo{volume}{200}, \bibinfo{number}{6} (\bibinfo{date}{June}
  \bibinfo{year}{2009}), \bibinfo{pages}{676.e1--676.e4}.
\newblock
\showISSN{0002-9378}
\urldef\tempurl%
\url{https://doi.org/10.1016/j.ajog.2009.02.031}
\showDOI{\tempurl}


\bibitem[Oh et~al\mbox{.}(2021)]%
        {oh2021background}
\bibfield{author}{\bibinfo{person}{Youngmin Oh}, \bibinfo{person}{Beomjun Kim},
  {and} \bibinfo{person}{Bumsub Ham}.} \bibinfo{year}{2021}\natexlab{}.
\newblock \showarticletitle{Background-aware pooling and noise-aware loss for
  weakly-supervised semantic segmentation}. In
  \bibinfo{booktitle}{\emph{Computer Vision and Pattern Recognition (CVPR)}}.
  \bibinfo{pages}{6913--6922}.
\newblock


\bibitem[Olesen et~al\mbox{.}(2024)]%
        {olesen2024slicing}
\bibfield{author}{\bibinfo{person}{Vincent Olesen}, \bibinfo{person}{Nina
  Weng}, \bibinfo{person}{Aasa Feragen}, {and} \bibinfo{person}{Eike
  Petersen}.} \bibinfo{year}{2024}\natexlab{}.
\newblock \showarticletitle{Slicing Through Bias: Explaining Performance Gaps
  in Medical Image Analysis Using Slice Discovery Methods}. In
  \bibinfo{booktitle}{\emph{MICCAI Workshop on Fairness of AI in Medical
  Imaging (FAIMI)}}. Springer, \bibinfo{pages}{3--13}.
\newblock


\bibitem[Ong~Ly et~al\mbox{.}(2024)]%
        {ong2024shortcut}
\bibfield{author}{\bibinfo{person}{Cathy Ong~Ly}, \bibinfo{person}{Balagopal
  Unnikrishnan}, \bibinfo{person}{Tony Tadic}, \bibinfo{person}{Tirth Patel},
  \bibinfo{person}{Joe Duhamel}, \bibinfo{person}{Sonja Kandel},
  \bibinfo{person}{Yasbanoo Moayedi}, \bibinfo{person}{Michael Brudno},
  \bibinfo{person}{Andrew Hope}, \bibinfo{person}{Heather Ross},
  {et~al\mbox{.}}} \bibinfo{year}{2024}\natexlab{}.
\newblock \showarticletitle{Shortcut learning in medical AI hinders
  generalization: method for estimating AI model generalization without
  external data}.
\newblock \bibinfo{journal}{\emph{NPJ Digital Medicine}} \bibinfo{volume}{7},
  \bibinfo{number}{1} (\bibinfo{year}{2024}), \bibinfo{pages}{124}.
\newblock


\bibitem[{\O}rting et~al\mbox{.}(2020)]%
        {orting2020survey}
\bibfield{author}{\bibinfo{person}{Silas~Nyboe {\O}rting},
  \bibinfo{person}{Andrew Doyle}, \bibinfo{person}{Arno van Hilten},
  \bibinfo{person}{Matthias Hirth}, \bibinfo{person}{Oana Inel},
  \bibinfo{person}{Christopher~R Madan}, \bibinfo{person}{Panagiotis Mavridis},
  \bibinfo{person}{Helen Spiers}, {and} \bibinfo{person}{Veronika Cheplygina}.}
  \bibinfo{year}{2020}\natexlab{}.
\newblock \showarticletitle{A survey of crowdsourcing in medical image
  analysis}.
\newblock \bibinfo{journal}{\emph{Human Computation}}  \bibinfo{volume}{7}
  (\bibinfo{year}{2020}), \bibinfo{pages}{1--26}.
\newblock


\bibitem[Pacheco and Krohling(2021)]%
        {pacheco2021attention}
\bibfield{author}{\bibinfo{person}{Andre~GC Pacheco} {and}
  \bibinfo{person}{Renato~A Krohling}.} \bibinfo{year}{2021}\natexlab{}.
\newblock \showarticletitle{An attention-based mechanism to combine images and
  metadata in deep learning models applied to skin cancer classification}.
\newblock \bibinfo{journal}{\emph{IEEE Journal of Biomedical and Health
  Informatics}} \bibinfo{volume}{25}, \bibinfo{number}{9}
  (\bibinfo{year}{2021}), \bibinfo{pages}{3554--3563}.
\newblock


\bibitem[Pacheco et~al\mbox{.}(2020)]%
        {pacheco2020pad}
\bibfield{author}{\bibinfo{person}{Andre~GC Pacheco},
  \bibinfo{person}{Gustavo~R Lima}, \bibinfo{person}{Amanda~S Salomao},
  \bibinfo{person}{Breno Krohling}, \bibinfo{person}{Igor~P Biral},
  \bibinfo{person}{Gabriel~G de Angelo}, \bibinfo{person}{F{\'a}bio~CR
  Alves~Jr}, \bibinfo{person}{Jos{\'e}~GM Esgario}, \bibinfo{person}{Alana~C
  Simora}, \bibinfo{person}{Pedro~BC Castro}, {et~al\mbox{.}}}
  \bibinfo{year}{2020}\natexlab{}.
\newblock \showarticletitle{PAD-UFES-20: A skin lesion dataset composed of
  patient data and clinical images collected from smartphones}.
\newblock \bibinfo{journal}{\emph{Data in Brief}}  \bibinfo{volume}{32}
  (\bibinfo{year}{2020}), \bibinfo{pages}{106221}.
\newblock


\bibitem[Pan and Yang(2010)]%
        {pan2010survey}
\bibfield{author}{\bibinfo{person}{Sinno~Jialin Pan} {and}
  \bibinfo{person}{Qiang Yang}.} \bibinfo{year}{2010}\natexlab{}.
\newblock \showarticletitle{A survey on transfer learning}.
\newblock \bibinfo{journal}{\emph{IEEE Transactions on Knowledge and Data
  Engineering}} \bibinfo{volume}{22}, \bibinfo{number}{10}
  (\bibinfo{year}{2010}), \bibinfo{pages}{1345--1359}.
\newblock


\bibitem[Paul et~al\mbox{.}(2022)]%
        {paul2022demographic}
\bibfield{author}{\bibinfo{person}{H~Yi Paul}, \bibinfo{person}{Tae~Kyung Kim},
  \bibinfo{person}{Eliot Siegel}, {and} \bibinfo{person}{Noushin
  Yahyavi-Firouz-Abadi}.} \bibinfo{year}{2022}\natexlab{}.
\newblock \showarticletitle{Demographic reporting in publicly available chest
  radiograph data sets: opportunities for mitigating sex and racial disparities
  in deep learning models}.
\newblock \bibinfo{journal}{\emph{Journal of the American College of
  Radiology}} \bibinfo{volume}{19}, \bibinfo{number}{1} (\bibinfo{year}{2022}),
  \bibinfo{pages}{192--200}.
\newblock


\bibitem[Paullada et~al\mbox{.}(2021)]%
        {paullada2021data}
\bibfield{author}{\bibinfo{person}{Amandalynne Paullada},
  \bibinfo{person}{Inioluwa~Deborah Raji}, \bibinfo{person}{Emily~M Bender},
  \bibinfo{person}{Emily Denton}, {and} \bibinfo{person}{Alex Hanna}.}
  \bibinfo{year}{2021}\natexlab{}.
\newblock \showarticletitle{Data and its (dis) contents: A survey of dataset
  development and use in machine learning research}.
\newblock \bibinfo{journal}{\emph{Patterns}} \bibinfo{volume}{2},
  \bibinfo{number}{11} (\bibinfo{year}{2021}).
\newblock


\bibitem[Peng et~al\mbox{.}(2021)]%
        {peng2021mitigating}
\bibfield{author}{\bibinfo{person}{Kenneth Peng}, \bibinfo{person}{Arunesh
  Mathur}, {and} \bibinfo{person}{Arvind Narayanan}.}
  \bibinfo{year}{2021}\natexlab{}.
\newblock \showarticletitle{Mitigating dataset harms requires stewardship:
  Lessons from 1000 papers}. In \bibinfo{booktitle}{\emph{Neural Information
  Processing Systems (NeurIPS) Datasets and Benchmarks Track}},
  \bibfield{editor}{\bibinfo{person}{J.~Vanschoren} {and}
  \bibinfo{person}{S.~Yeung}} (Eds.), Vol.~\bibinfo{volume}{1}.
\newblock


\bibitem[Pernet et~al\mbox{.}(2023)]%
        {pernet2023long}
\bibfield{author}{\bibinfo{person}{Cyril Pernet}, \bibinfo{person}{Claus
  Svarer}, \bibinfo{person}{Ross Blair}, \bibinfo{person}{John~D Van~Horn},
  {and} \bibinfo{person}{Russell~A Poldrack}.} \bibinfo{year}{2023}\natexlab{}.
\newblock \showarticletitle{On the long-term archiving of research data}.
\newblock \bibinfo{journal}{\emph{Neuroinformatics}} \bibinfo{volume}{21},
  \bibinfo{number}{2} (\bibinfo{year}{2023}), \bibinfo{pages}{243--246}.
\newblock


\bibitem[Philipp et~al\mbox{.}(2024)]%
        {philipp2024annotation}
\bibfield{author}{\bibinfo{person}{Lena Philipp}, \bibinfo{person}{Maarten de
  Rooij}, \bibinfo{person}{John Hermans}, \bibinfo{person}{Matthieu Rutten},
  \bibinfo{person}{Horst~Karl Hahn}, \bibinfo{person}{Bram van Ginneken}, {and}
  \bibinfo{person}{Alessa Hering}.} \bibinfo{year}{2024}\natexlab{}.
\newblock \showarticletitle{Annotation-Efficient Strategy for Segmentation of
  3D Body Composition}. In \bibinfo{booktitle}{\emph{Medical Imaging with Deep
  Learning (MIDL)}}.
\newblock


\bibitem[Poldrack et~al\mbox{.}(2024)]%
        {poldrack2024past}
\bibfield{author}{\bibinfo{person}{Russell~A Poldrack},
  \bibinfo{person}{Christopher~J Markiewicz}, \bibinfo{person}{Stefan
  Appelhoff}, \bibinfo{person}{Yoni~K Ashar}, \bibinfo{person}{Tibor Auer},
  \bibinfo{person}{Sylvain Baillet}, \bibinfo{person}{Shashank Bansal},
  \bibinfo{person}{Leandro Beltrachini}, \bibinfo{person}{Christian~G Benar},
  \bibinfo{person}{Giacomo Bertazzoli}, {et~al\mbox{.}}}
  \bibinfo{year}{2024}\natexlab{}.
\newblock \showarticletitle{The past, present, and future of the brain imaging
  data structure (BIDS)}.
\newblock \bibinfo{journal}{\emph{Imaging Neuroscience}}  \bibinfo{volume}{2}
  (\bibinfo{year}{2024}), \bibinfo{pages}{1--19}.
\newblock


\bibitem[Priem et~al\mbox{.}(2022)]%
        {priem2022openalex}
\bibfield{author}{\bibinfo{person}{Jason Priem}, \bibinfo{person}{Heather
  Piwowar}, {and} \bibinfo{person}{Richard Orr}.}
  \bibinfo{year}{2022}\natexlab{}.
\newblock \bibinfo{title}{OpenAlex: A fully-open index of scholarly works,
  authors, venues, institutions, and concepts}.
\newblock
\newblock
\showeprint[arxiv]{2205.01833}~[cs.DL]


\bibitem[R{\"a}dsch et~al\mbox{.}(2025)]%
        {radsch2025quality}
\bibfield{author}{\bibinfo{person}{Tim R{\"a}dsch}, \bibinfo{person}{Annika
  Reinke}, \bibinfo{person}{Vivienn Weru}, \bibinfo{person}{Minu~D Tizabi},
  \bibinfo{person}{Nicholas Heller}, \bibinfo{person}{Fabian Isensee},
  \bibinfo{person}{Annette Kopp-Schneider}, {and} \bibinfo{person}{Lena
  Maier-Hein}.} \bibinfo{year}{2025}\natexlab{}.
\newblock \showarticletitle{Quality Assured: Rethinking Annotation Strategies
  in Imaging AI}. In \bibinfo{booktitle}{\emph{European Conference on Computer
  Vision}}. Springer, \bibinfo{pages}{52--69}.
\newblock


\bibitem[R{\"a}dsch et~al\mbox{.}(2023)]%
        {radsch2023labelling}
\bibfield{author}{\bibinfo{person}{Tim R{\"a}dsch}, \bibinfo{person}{Annika
  Reinke}, \bibinfo{person}{Vivienn Weru}, \bibinfo{person}{Minu~D Tizabi},
  \bibinfo{person}{Nicholas Schreck}, \bibinfo{person}{A~Emre Kavur},
  \bibinfo{person}{B{\"u}nyamin Pekdemir}, \bibinfo{person}{Tobias Ro{\ss}},
  \bibinfo{person}{Annette Kopp-Schneider}, {and} \bibinfo{person}{Lena
  Maier-Hein}.} \bibinfo{year}{2023}\natexlab{}.
\newblock \showarticletitle{Labelling instructions matter in biomedical image
  analysis}.
\newblock \bibinfo{journal}{\emph{Nature Machine Intelligence}}
  \bibinfo{volume}{5}, \bibinfo{number}{3} (\bibinfo{year}{2023}),
  \bibinfo{pages}{273--283}.
\newblock


\bibitem[Raji et~al\mbox{.}(2021)]%
        {raji2021ai}
\bibfield{author}{\bibinfo{person}{Inioluwa~Deborah Raji},
  \bibinfo{person}{Emily~M Bender}, \bibinfo{person}{Amandalynne Paullada},
  \bibinfo{person}{Emily Denton}, {and} \bibinfo{person}{Alex Hanna}.}
  \bibinfo{year}{2021}\natexlab{}.
\newblock \showarticletitle{{AI} and the everything in the whole wide world
  benchmark}.
\newblock \bibinfo{journal}{\emph{arXiv preprint arXiv:2111.15366}}
  (\bibinfo{year}{2021}).
\newblock


\bibitem[Rajpurkar(2017)]%
        {rajpurkar2017chexnet}
\bibfield{author}{\bibinfo{person}{P Rajpurkar}.}
  \bibinfo{year}{2017}\natexlab{}.
\newblock \showarticletitle{CheXNet: Radiologist-Level Pneumonia Detection on
  Chest X-Rays with Deep Learning}.
\newblock \bibinfo{journal}{\emph{arXiv preprint arXiv:1711.05225}}
  (\bibinfo{year}{2017}).
\newblock


\bibitem[Rajpurkar et~al\mbox{.}(2022)]%
        {rajpurkar2022ai}
\bibfield{author}{\bibinfo{person}{Pranav Rajpurkar}, \bibinfo{person}{Emma
  Chen}, \bibinfo{person}{Oishi Banerjee}, {and} \bibinfo{person}{Eric~J
  Topol}.} \bibinfo{year}{2022}\natexlab{}.
\newblock \showarticletitle{{AI} in health and medicine}.
\newblock \bibinfo{journal}{\emph{Nature Medicine}} \bibinfo{volume}{28},
  \bibinfo{number}{1} (\bibinfo{year}{2022}), \bibinfo{pages}{31--38}.
\newblock


\bibitem[Raumanns et~al\mbox{.}(2021)]%
        {raumanns2021enhance}
\bibfield{author}{\bibinfo{person}{Ralf Raumanns}, \bibinfo{person}{Gerard
  Schouten}, \bibinfo{person}{Max Joosten}, \bibinfo{person}{Josien~PW Pluim},
  \bibinfo{person}{Veronika Cheplygina}, {et~al\mbox{.}}}
  \bibinfo{year}{2021}\natexlab{}.
\newblock \showarticletitle{ENHANCE (ENriching Health data by ANnotations of
  Crowd and Experts): A case study for skin lesion classification}.
\newblock \bibinfo{journal}{\emph{Machine Learning for Biomedical Imaging}}
  \bibinfo{volume}{1}, \bibinfo{number}{December 2021 issue}
  (\bibinfo{year}{2021}), \bibinfo{pages}{1--26}.
\newblock


\bibitem[Reinke et~al\mbox{.}(2023)]%
        {reinke2023commonlimitationsimageprocessing}
\bibfield{author}{\bibinfo{person}{Annika Reinke}, \bibinfo{person}{Minu~D.
  Tizabi}, \bibinfo{person}{Carole~H. Sudre}, \bibinfo{person}{Matthias
  Eisenmann}, \bibinfo{person}{Tim Rädsch}, \bibinfo{person}{Michael
  Baumgartner}, \bibinfo{person}{Laura Acion}, \bibinfo{person}{Michela
  Antonelli}, \bibinfo{person}{Tal Arbel}, \bibinfo{person}{Spyridon Bakas},
  \bibinfo{person}{Peter Bankhead}, \bibinfo{person}{Arriel Benis},
  \bibinfo{person}{Matthew Blaschko}, \bibinfo{person}{Florian Buettner},
  \bibinfo{person}{M.~Jorge Cardoso}, \bibinfo{person}{Jianxu Chen},
  \bibinfo{person}{Veronika Cheplygina}, \bibinfo{person}{Evangelia
  Christodoulou}, \bibinfo{person}{Beth Cimini}, \bibinfo{person}{Gary~S.
  Collins}, \bibinfo{person}{Sandy Engelhardt}, \bibinfo{person}{Keyvan
  Farahani}, \bibinfo{person}{Luciana Ferrer}, \bibinfo{person}{Adrian
  Galdran}, \bibinfo{person}{Bram van Ginneken}, \bibinfo{person}{Ben Glocker},
  \bibinfo{person}{Patrick Godau}, \bibinfo{person}{Robert Haase},
  \bibinfo{person}{Fred Hamprecht}, \bibinfo{person}{Daniel~A. Hashimoto},
  \bibinfo{person}{Doreen Heckmann-Nötzel}, \bibinfo{person}{Peter Hirsch},
  \bibinfo{person}{Michael~M. Hoffman}, \bibinfo{person}{Merel Huisman},
  \bibinfo{person}{Fabian Isensee}, \bibinfo{person}{Pierre Jannin},
  \bibinfo{person}{Charles~E. Kahn}, \bibinfo{person}{Dagmar Kainmueller},
  \bibinfo{person}{Bernhard Kainz}, \bibinfo{person}{Alexandros Karargyris},
  \bibinfo{person}{Alan Karthikesalingam}, \bibinfo{person}{A.~Emre Kavur},
  \bibinfo{person}{Hannes Kenngott}, \bibinfo{person}{Jens Kleesiek},
  \bibinfo{person}{Andreas Kleppe}, \bibinfo{person}{Sven Kohler},
  \bibinfo{person}{Florian Kofler}, \bibinfo{person}{Annette Kopp-Schneider},
  \bibinfo{person}{Thijs Kooi}, \bibinfo{person}{Michal Kozubek},
  \bibinfo{person}{Anna Kreshuk}, \bibinfo{person}{Tahsin Kurc},
  \bibinfo{person}{Bennett~A. Landman}, \bibinfo{person}{Geert Litjens},
  \bibinfo{person}{Amin Madani}, \bibinfo{person}{Klaus Maier-Hein},
  \bibinfo{person}{Anne~L. Martel}, \bibinfo{person}{Peter Mattson},
  \bibinfo{person}{Erik Meijering}, \bibinfo{person}{Bjoern Menze},
  \bibinfo{person}{David Moher}, \bibinfo{person}{Karel G.~M. Moons},
  \bibinfo{person}{Henning Müller}, \bibinfo{person}{Brennan Nichyporuk},
  \bibinfo{person}{Felix Nickel}, \bibinfo{person}{M.~Alican Noyan},
  \bibinfo{person}{Jens Petersen}, \bibinfo{person}{Gorkem Polat},
  \bibinfo{person}{Susanne~M. Rafelski}, \bibinfo{person}{Nasir Rajpoot},
  \bibinfo{person}{Mauricio Reyes}, \bibinfo{person}{Nicola Rieke},
  \bibinfo{person}{Michael Riegler}, \bibinfo{person}{Hassan Rivaz},
  \bibinfo{person}{Julio Saez-Rodriguez}, \bibinfo{person}{Clara~I. Sánchez},
  \bibinfo{person}{Julien Schroeter}, \bibinfo{person}{Anindo Saha},
  \bibinfo{person}{M.~Alper Selver}, \bibinfo{person}{Lalith Sharan},
  \bibinfo{person}{Shravya Shetty}, \bibinfo{person}{Maarten van Smeden},
  \bibinfo{person}{Bram Stieltjes}, \bibinfo{person}{Ronald~M. Summers},
  \bibinfo{person}{Abdel~A. Taha}, \bibinfo{person}{Aleksei Tiulpin},
  \bibinfo{person}{Sotirios~A. Tsaftaris}, \bibinfo{person}{Ben~Van Calster},
  \bibinfo{person}{Gaël Varoquaux}, \bibinfo{person}{Manuel Wiesenfarth},
  \bibinfo{person}{Ziv~R. Yaniv}, \bibinfo{person}{Paul Jäger}, {and}
  \bibinfo{person}{Lena Maier-Hein}.} \bibinfo{year}{2023}\natexlab{}.
\newblock \bibinfo{title}{Common Limitations of Image Processing Metrics: A
  Picture Story}.
\newblock
\newblock
\urldef\tempurl%
\url{https://arxiv.org/abs/2104.05642}
\showURL{%
\tempurl}


\bibitem[Review(2025)]%
        {aclrollingreview}
\bibfield{author}{\bibinfo{person}{ACL~Rolling Review}.}
  \bibinfo{year}{2025}\natexlab{}.
\newblock \bibinfo{title}{Call for papers}.
\newblock \bibinfo{howpublished}{https://aclrollingreview.org/cfp}.
\newblock
\newblock
\shownote{Accessed: 2025-04-22}.


\bibitem[Ricci~Lara et~al\mbox{.}(2022)]%
        {ricci2022addressing}
\bibfield{author}{\bibinfo{person}{Mar{\'\i}a~Agustina Ricci~Lara},
  \bibinfo{person}{Rodrigo Echeveste}, {and} \bibinfo{person}{Enzo Ferrante}.}
  \bibinfo{year}{2022}\natexlab{}.
\newblock \showarticletitle{Addressing fairness in artificial intelligence for
  medical imaging}.
\newblock \bibinfo{journal}{\emph{Nature Communications}} \bibinfo{volume}{13},
  \bibinfo{number}{1} (\bibinfo{year}{2022}), \bibinfo{pages}{1--6}.
\newblock


\bibitem[Ricci~Lara et~al\mbox{.}(2023)]%
        {ricci2023dataset}
\bibfield{author}{\bibinfo{person}{Mar{\'\i}a~Agustina Ricci~Lara},
  \bibinfo{person}{Mar{\'\i}a~Victoria Rodr{\'\i}guez~Kowalczuk},
  \bibinfo{person}{Maite Lisa~Eliceche},
  \bibinfo{person}{Mar{\'\i}a~Guillermina Ferraresso},
  \bibinfo{person}{Daniel~Roberto Luna}, \bibinfo{person}{Sonia~Elizabeth
  Benitez}, {and} \bibinfo{person}{Luis~Daniel Mazzuoccolo}.}
  \bibinfo{year}{2023}\natexlab{}.
\newblock \showarticletitle{A dataset of skin lesion images collected in
  Argentina for the evaluation of AI tools in this population}.
\newblock \bibinfo{journal}{\emph{Scientific Data}} \bibinfo{volume}{10},
  \bibinfo{number}{1} (\bibinfo{year}{2023}), \bibinfo{pages}{712}.
\newblock


\bibitem[Rieke et~al\mbox{.}(2020)]%
        {rieke2020future}
\bibfield{author}{\bibinfo{person}{Nicola Rieke}, \bibinfo{person}{Jonny
  Hancox}, \bibinfo{person}{Wenqi Li}, \bibinfo{person}{Fausto Milletari},
  \bibinfo{person}{Holger~R Roth}, \bibinfo{person}{Shadi Albarqouni},
  \bibinfo{person}{Spyridon Bakas}, \bibinfo{person}{Mathieu~N Galtier},
  \bibinfo{person}{Bennett~A Landman}, \bibinfo{person}{Klaus Maier-Hein},
  {et~al\mbox{.}}} \bibinfo{year}{2020}\natexlab{}.
\newblock \showarticletitle{The future of digital health with federated
  learning}.
\newblock \bibinfo{journal}{\emph{NPJ digital medicine}} \bibinfo{volume}{3},
  \bibinfo{number}{1} (\bibinfo{year}{2020}), \bibinfo{pages}{1--7}.
\newblock


\bibitem[Rostamzadeh et~al\mbox{.}(2022)]%
        {rostamzadeh2022healthsheet}
\bibfield{author}{\bibinfo{person}{Negar Rostamzadeh}, \bibinfo{person}{Diana
  Mincu}, \bibinfo{person}{Subhrajit Roy}, \bibinfo{person}{Andrew Smart},
  \bibinfo{person}{Lauren Wilcox}, \bibinfo{person}{Mahima Pushkarna},
  \bibinfo{person}{Jessica Schrouff}, \bibinfo{person}{Razvan Amironesei},
  \bibinfo{person}{Nyalleng Moorosi}, {and} \bibinfo{person}{Katherine
  Heller}.} \bibinfo{year}{2022}\natexlab{}.
\newblock \showarticletitle{Healthsheet: development of a transparency artifact
  for health datasets}. In \bibinfo{booktitle}{\emph{Proceedings of the 2022
  ACM Conference on Fairness, Accountability, and Transparency}}.
  \bibinfo{pages}{1943--1961}.
\newblock


\bibitem[Routier et~al\mbox{.}(2021)]%
        {Routier2021}
\bibfield{author}{\bibinfo{person}{Alexandre Routier}, \bibinfo{person}{Ninon
  Burgos}, \bibinfo{person}{Mauricio Díaz}, \bibinfo{person}{Michael Bacci},
  \bibinfo{person}{Simona Bottani}, \bibinfo{person}{Omar El-Rifai},
  \bibinfo{person}{Sabrina Fontanella}, \bibinfo{person}{Pietro Gori},
  \bibinfo{person}{Jérémy Guillon}, \bibinfo{person}{Alexis Guyot},
  \bibinfo{person}{Ravi Hassanaly}, \bibinfo{person}{Thomas Jacquemont},
  \bibinfo{person}{Pascal Lu}, \bibinfo{person}{Arnaud Marcoux},
  \bibinfo{person}{Tristan Moreau}, \bibinfo{person}{Jorge Samper-González},
  \bibinfo{person}{Marc Teichmann}, \bibinfo{person}{Elina Thibeau-Sutre},
  \bibinfo{person}{Ghislain Vaillant}, \bibinfo{person}{Junhao Wen},
  \bibinfo{person}{Adam Wild}, \bibinfo{person}{Marie~Odile Habert},
  \bibinfo{person}{Stanley Durrleman}, {and} \bibinfo{person}{Olivier
  Colliot}.} \bibinfo{year}{2021}\natexlab{}.
\newblock \showarticletitle{Clinica: An Open-Source Software Platform for
  Reproducible Clinical Neuroscience Studies}.
\newblock \bibinfo{journal}{\emph{Frontiers in Neuroinformatics}}
  \bibinfo{volume}{15} (\bibinfo{date}{8} \bibinfo{year}{2021}).
\newblock
\showISSN{16625196}
\urldef\tempurl%
\url{https://doi.org/10.3389/fninf.2021.689675}
\showDOI{\tempurl}


\bibitem[Rowhani-Farid and Barnett(2016)]%
        {rowhani2016has}
\bibfield{author}{\bibinfo{person}{Anisa Rowhani-Farid} {and}
  \bibinfo{person}{Adrian~G Barnett}.} \bibinfo{year}{2016}\natexlab{}.
\newblock \showarticletitle{Has open data arrived at the British Medical
  Journal (BMJ)? An observational study}.
\newblock \bibinfo{journal}{\emph{BMJ Open}} \bibinfo{volume}{6},
  \bibinfo{number}{10} (\bibinfo{year}{2016}), \bibinfo{pages}{e011784}.
\newblock


\bibitem[Russakovsky et~al\mbox{.}(2015)]%
        {russakovsky2015imagenet}
\bibfield{author}{\bibinfo{person}{Olga Russakovsky}, \bibinfo{person}{Jia
  Deng}, \bibinfo{person}{Hao Su}, \bibinfo{person}{Jonathan Krause},
  \bibinfo{person}{Sanjeev Satheesh}, \bibinfo{person}{Sean Ma},
  \bibinfo{person}{Zhiheng Huang}, \bibinfo{person}{Andrej Karpathy},
  \bibinfo{person}{Aditya Khosla}, \bibinfo{person}{Michael Bernstein},
  {et~al\mbox{.}}} \bibinfo{year}{2015}\natexlab{}.
\newblock \showarticletitle{{ImageNet Large Scale Visual Recognition
  Challenge}}.
\newblock \bibinfo{journal}{\emph{{International Journal of Computer Vision}}}
  \bibinfo{volume}{115}, \bibinfo{number}{3} (\bibinfo{year}{2015}),
  \bibinfo{pages}{211--252}.
\newblock


\bibitem[Sambasivan et~al\mbox{.}(2021)]%
        {sambasivan2021everyone}
\bibfield{author}{\bibinfo{person}{Nithya Sambasivan}, \bibinfo{person}{Shivani
  Kapania}, \bibinfo{person}{Hannah Highfill}, \bibinfo{person}{Diana Akrong},
  \bibinfo{person}{Praveen Paritosh}, {and} \bibinfo{person}{Lora~M Aroyo}.}
  \bibinfo{year}{2021}\natexlab{}.
\newblock \showarticletitle{“{Everyone} wants to do the model work, not the
  data work”: Data Cascades in High-Stakes {AI}}. In
  \bibinfo{booktitle}{\emph{Proceedings of the 2021 CHI Conference on Human
  Factors in Computing Systems}}. \bibinfo{pages}{1--15}.
\newblock


\bibitem[Sappia(2024)]%
        {ACOUSLIC}
\bibfield{author}{\bibinfo{person}{Mar{\'i}a~Sof{\'i}a Sappia}.}
  \bibinfo{year}{2024}\natexlab{}.
\newblock \bibinfo{booktitle}{\emph{{{ACOUSLIC-AI}} : {{Abdominal Circumference
  Operator-agnostic UltraSound}} Measurement}}.
\newblock
\urldef\tempurl%
\url{https://doi.org/10.5281/zenodo.12697994}
\showDOI{\tempurl}


\bibitem[Savage and Vickers(2009)]%
        {savage2009empirical}
\bibfield{author}{\bibinfo{person}{Caroline~J Savage} {and}
  \bibinfo{person}{Andrew~J Vickers}.} \bibinfo{year}{2009}\natexlab{}.
\newblock \showarticletitle{Empirical study of data sharing by authors
  publishing in PLoS journals}.
\newblock \bibinfo{journal}{\emph{PLoS One}} \bibinfo{volume}{4},
  \bibinfo{number}{9} (\bibinfo{year}{2009}), \bibinfo{pages}{e7078}.
\newblock


\bibitem[Scarpace et~al\mbox{.}(2016)]%
        {scarpace2016cancer}
\bibfield{author}{\bibinfo{person}{L Scarpace}, \bibinfo{person}{T Mikkelsen},
  \bibinfo{person}{S Cha}, \bibinfo{person}{S Rao}, \bibinfo{person}{S
  Tekchandani}, \bibinfo{person}{D Gutman}, \bibinfo{person}{JH Saltz},
  \bibinfo{person}{BJ Erickson}, \bibinfo{person}{N Pedano},
  \bibinfo{person}{AE Flanders}, {et~al\mbox{.}}}
  \bibinfo{year}{2016}\natexlab{}.
\newblock \showarticletitle{The cancer genome atlas glioblastoma multiforme
  collection (TCGA-GBM)(Version 4)[data set]}.
\newblock \bibinfo{journal}{\emph{The Cancer Imaging Archive}}
  \bibinfo{volume}{10} (\bibinfo{year}{2016}), \bibinfo{pages}{K9}.
\newblock


\bibitem[Scheuerman et~al\mbox{.}(2021)]%
        {scheuerman2021datasets}
\bibfield{author}{\bibinfo{person}{Morgan~Klaus Scheuerman},
  \bibinfo{person}{Alex Hanna}, {and} \bibinfo{person}{Emily Denton}.}
  \bibinfo{year}{2021}\natexlab{}.
\newblock \showarticletitle{Do datasets have politics? Disciplinary values in
  computer vision dataset development}.
\newblock \bibinfo{journal}{\emph{Proceedings of the ACM on Human-Computer
  Interaction}} \bibinfo{volume}{5}, \bibinfo{number}{CSCW2}
  (\bibinfo{year}{2021}), \bibinfo{pages}{1--37}.
\newblock


\bibitem[Schmidt et~al\mbox{.}(2023)]%
        {schmidt2023automated}
\bibfield{author}{\bibinfo{person}{Lena Schmidt}, \bibinfo{person}{Saleh
  Mohamed}, \bibinfo{person}{Nick Meader}, \bibinfo{person}{Jaume Bacardit},
  {and} \bibinfo{person}{Dawn Craig}.} \bibinfo{year}{2023}\natexlab{}.
\newblock \showarticletitle{Automated data analysis of unstructured grey
  literature in health research: A mapping review}.
\newblock \bibinfo{journal}{\emph{Research Synthesis Methods}}
  (\bibinfo{year}{2023}).
\newblock


\bibitem[Schmidt et~al\mbox{.}(2021)]%
        {schmidt2021data}
\bibfield{author}{\bibinfo{person}{Lena Schmidt}, \bibinfo{person}{Ailbhe
  N~Finnerty Mutlu}, \bibinfo{person}{Rebecca Elmore},
  \bibinfo{person}{Babatunde~K Olorisade}, \bibinfo{person}{James Thomas},
  {and} \bibinfo{person}{Julian~PT Higgins}.} \bibinfo{year}{2021}\natexlab{}.
\newblock \showarticletitle{Data extraction methods for systematic review
  (semi) automation: Update of a living systematic review}.
\newblock \bibinfo{journal}{\emph{F1000Research}}  \bibinfo{volume}{10}
  (\bibinfo{year}{2021}).
\newblock


\bibitem[Sellergren et~al\mbox{.}(2023)]%
        {sellergrengeneralized}
\bibfield{author}{\bibinfo{person}{Andrew Sellergren}, \bibinfo{person}{Atilla
  Kiraly}, \bibinfo{person}{Tom Pollard}, \bibinfo{person}{Wei-Hung Weng},
  \bibinfo{person}{Yun Liu}, \bibinfo{person}{Akib Uddin}, {and}
  \bibinfo{person}{Christina Chen}.} \bibinfo{year}{2023}\natexlab{}.
\newblock \bibinfo{title}{Generalized Image Embeddings for the MIMIC Chest
  X-Ray dataset}.
\newblock
\newblock
\urldef\tempurl%
\url{https://doi.org/10.13026/PXC2-VX69}
\showDOI{\tempurl}


\bibitem[Seyyed-Kalantari et~al\mbox{.}(2020)]%
        {seyyed2020chexclusion}
\bibfield{author}{\bibinfo{person}{Laleh Seyyed-Kalantari},
  \bibinfo{person}{Guanxiong Liu}, \bibinfo{person}{Matthew McDermott},
  \bibinfo{person}{Irene~Y Chen}, {and} \bibinfo{person}{Marzyeh Ghassemi}.}
  \bibinfo{year}{2020}\natexlab{}.
\newblock \showarticletitle{CheXclusion: Fairness gaps in deep chest X-ray
  classifiers}. In \bibinfo{booktitle}{\emph{Pacific Symposium on
  Biocomputing}}. World Scientific, \bibinfo{pages}{232--243}.
\newblock


\bibitem[Seyyed-Kalantari et~al\mbox{.}(2021)]%
        {seyyed2021underdiagnosis}
\bibfield{author}{\bibinfo{person}{Laleh Seyyed-Kalantari},
  \bibinfo{person}{Haoran Zhang}, \bibinfo{person}{Matthew~BA McDermott},
  \bibinfo{person}{Irene~Y Chen}, {and} \bibinfo{person}{Marzyeh Ghassemi}.}
  \bibinfo{year}{2021}\natexlab{}.
\newblock \showarticletitle{Underdiagnosis bias of artificial intelligence
  algorithms applied to chest radiographs in under-served patient populations}.
\newblock \bibinfo{journal}{\emph{Nature Medicine}} \bibinfo{volume}{27},
  \bibinfo{number}{12} (\bibinfo{year}{2021}), \bibinfo{pages}{2176--2182}.
\newblock


\bibitem[Shamshad et~al\mbox{.}(2023)]%
        {shamshad2023transformers}
\bibfield{author}{\bibinfo{person}{Fahad Shamshad}, \bibinfo{person}{Salman
  Khan}, \bibinfo{person}{Syed~Waqas Zamir}, \bibinfo{person}{Muhammad~Haris
  Khan}, \bibinfo{person}{Munawar Hayat}, \bibinfo{person}{Fahad~Shahbaz Khan},
  {and} \bibinfo{person}{Huazhu Fu}.} \bibinfo{year}{2023}\natexlab{}.
\newblock \showarticletitle{Transformers in medical imaging: A survey}.
\newblock \bibinfo{journal}{\emph{Medical Image Analysis}}
  \bibinfo{volume}{88} (\bibinfo{year}{2023}), \bibinfo{pages}{102802}.
\newblock
\showISSN{1361-8415}
\urldef\tempurl%
\url{https://doi.org/10.1016/j.media.2023.102802}
\showDOI{\tempurl}


\bibitem[Shen et~al\mbox{.}(2024)]%
        {shen2024data}
\bibfield{author}{\bibinfo{person}{Judy~Hanwen Shen},
  \bibinfo{person}{Inioluwa~Deborah Raji}, {and} \bibinfo{person}{Irene~Y
  Chen}.} \bibinfo{year}{2024}\natexlab{}.
\newblock \showarticletitle{The data addition dilemma}.
\newblock \bibinfo{journal}{\emph{arXiv preprint arXiv:2408.04154}}
  (\bibinfo{year}{2024}).
\newblock


\bibitem[Sletner et~al\mbox{.}(2015)]%
        {sletnerEthnic2015}
\bibfield{author}{\bibinfo{person}{Line Sletner}, \bibinfo{person}{Svein
  Rasmussen}, \bibinfo{person}{Anne~Karen Jenum}, \bibinfo{person}{Britt
  Nakstad}, \bibinfo{person}{Odd Harald~Rognerud Jensen}, {and}
  \bibinfo{person}{Siri Vangen}.} \bibinfo{year}{2015}\natexlab{}.
\newblock \showarticletitle{Ethnic Differences in Fetal Size and Growth in a
  Multi-Ethnic Population}.
\newblock \bibinfo{journal}{\emph{Early Human Development}}
  \bibinfo{volume}{91}, \bibinfo{number}{9} (\bibinfo{date}{Sept.}
  \bibinfo{year}{2015}), \bibinfo{pages}{547--554}.
\newblock
\showISSN{0378-3782}
\urldef\tempurl%
\url{https://doi.org/10.1016/j.earlhumdev.2015.07.002}
\showDOI{\tempurl}


\bibitem[Sourget et~al\mbox{.}(2024)]%
        {sourget2024citation}
\bibfield{author}{\bibinfo{person}{Th{\'e}o Sourget}, \bibinfo{person}{Ahmet
  Akko{\c{c}}}, \bibinfo{person}{Stinna Winther},
  \bibinfo{person}{Christine~Lyngbye Galsgaard}, \bibinfo{person}{Amelia
  Jim{\'e}nez-S{\'a}nchez}, \bibinfo{person}{Dovile Juodelyte},
  \bibinfo{person}{Caroline Petitjean}, {and} \bibinfo{person}{Veronika
  Cheplygina}.} \bibinfo{year}{2024}\natexlab{}.
\newblock \showarticletitle{{[{C}itation needed]} Data usage and citation
  practices in medical imaging conferences}. In
  \bibinfo{booktitle}{\emph{Medical Imaging with Deep Learing (MIDL), in
  press}}.
\newblock


\bibitem[Sterling et~al\mbox{.}(2022)]%
        {Sterling2022}
\bibfield{author}{\bibinfo{person}{Elijah Sterling}, \bibinfo{person}{Hannah
  Pearl}, \bibinfo{person}{Zexuan Liu}, \bibinfo{person}{Jason~W. Allen}, {and}
  \bibinfo{person}{Candace~C. Fleischer}.} \bibinfo{year}{2022}\natexlab{}.
\newblock \bibinfo{title}{Demographic reporting across a decade of
  neuroimaging: a systematic review}.
\newblock , \bibinfo{numpages}{2785-2796}~pages.
\newblock
Issue 6.
\showISSN{19317565}
\urldef\tempurl%
\url{https://doi.org/10.1007/s11682-022-00724-8}
\showDOI{\tempurl}


\bibitem[Suter et~al\mbox{.}(2022)]%
        {Suter2022}
\bibfield{author}{\bibinfo{person}{Yannick Suter}, \bibinfo{person}{Urspeter
  Knecht}, \bibinfo{person}{Waldo Valenzuela}, \bibinfo{person}{Michelle
  Notter}, \bibinfo{person}{Ekkehard Hewer}, \bibinfo{person}{Philippe
  Schucht}, \bibinfo{person}{Roland Wiest}, {and} \bibinfo{person}{Mauricio
  Reyes}.} \bibinfo{year}{2022}\natexlab{}.
\newblock \showarticletitle{The LUMIERE dataset: Longitudinal Glioblastoma MRI
  with expert RANO evaluation}.
\newblock \bibinfo{journal}{\emph{Scientific Data}}  \bibinfo{volume}{9}
  (\bibinfo{date}{12} \bibinfo{year}{2022}).
\newblock
Issue 1.
\showISSN{20524463}
\urldef\tempurl%
\url{https://doi.org/10.1038/s41597-022-01881-7}
\showDOI{\tempurl}


\bibitem[Ten~Berg et~al\mbox{.}(2024)]%
        {ten2024chatgpt}
\bibfield{author}{\bibinfo{person}{Hidde Ten~Berg}, \bibinfo{person}{Bram van
  Bakel}, \bibinfo{person}{Lieke van~de Wouw}, \bibinfo{person}{Kim~E Jie},
  \bibinfo{person}{Anoeska Schipper}, \bibinfo{person}{Henry Jansen},
  \bibinfo{person}{Rory~D O’Connor}, \bibinfo{person}{Bram van Ginneken},
  {and} \bibinfo{person}{Steef Kurstjens}.} \bibinfo{year}{2024}\natexlab{}.
\newblock \showarticletitle{ChatGPT and generating a differential diagnosis
  early in an emergency department presentation}.
\newblock \bibinfo{journal}{\emph{Annals of Emergency Medicine}}
  \bibinfo{volume}{83}, \bibinfo{number}{1} (\bibinfo{year}{2024}),
  \bibinfo{pages}{83--86}.
\newblock


\bibitem[Tong et~al\mbox{.}(2023)]%
        {Tong2023}
\bibfield{author}{\bibinfo{person}{Wen-Juan Tong}, \bibinfo{person}{Shao-Hong
  Wu}, \bibinfo{person}{Mei-Qing Cheng}, \bibinfo{person}{Hui Huang},
  \bibinfo{person}{Jin-Yu Liang}, \bibinfo{person}{Chao-Qun Li},
  \bibinfo{person}{Huan-Ling Guo}, \bibinfo{person}{Dan-Ni He},
  \bibinfo{person}{Yi-Hao Liu}, \bibinfo{person}{Han Xiao},
  \bibinfo{person}{Hang-Tong Hu}, \bibinfo{person}{Si-Min Ruan},
  \bibinfo{person}{Ming-De Li}, \bibinfo{person}{Ming-De Lu}, {and}
  \bibinfo{person}{Wei Wang}.} \bibinfo{year}{2023}\natexlab{}.
\newblock \showarticletitle{Integration of Artificial Intelligence Decision
  Aids to Reduce Workload and Enhance Efficiency in Thyroid Nodule Management}.
\newblock \bibinfo{journal}{\emph{JAMA Network Open}} \bibinfo{volume}{6},
  \bibinfo{number}{5} (\bibinfo{year}{2023}), \bibinfo{pages}{e2313674}.
\newblock
\showISSN{2574-3805}
\urldef\tempurl%
\url{https://doi.org/10.1001/jamanetworkopen.2023.13674}
\showDOI{\tempurl}


\bibitem[Tsai et~al\mbox{.}(2015)]%
        {tsaiObesity2015}
\bibfield{author}{\bibinfo{person}{Pai-Jong~Stacy Tsai},
  \bibinfo{person}{Matthew Loichinger}, {and} \bibinfo{person}{Ivica Zalud}.}
  \bibinfo{year}{2015}\natexlab{}.
\newblock \showarticletitle{Obesity and the Challenges of Ultrasound Fetal
  Abnormality Diagnosis}.
\newblock \bibinfo{journal}{\emph{Best Practice \& Research Clinical Obstetrics
  \& Gynaecology}} \bibinfo{volume}{29}, \bibinfo{number}{3}
  (\bibinfo{date}{April} \bibinfo{year}{2015}), \bibinfo{pages}{320--327}.
\newblock
\showISSN{1521-6934}
\urldef\tempurl%
\url{https://doi.org/10.1016/j.bpobgyn.2014.08.011}
\showDOI{\tempurl}


\bibitem[Tschandl et~al\mbox{.}(2018)]%
        {tschandl2018ham10000}
\bibfield{author}{\bibinfo{person}{Philipp Tschandl}, \bibinfo{person}{Cliff
  Rosendahl}, {and} \bibinfo{person}{Harald Kittler}.}
  \bibinfo{year}{2018}\natexlab{}.
\newblock \showarticletitle{The HAM10000 dataset, a large collection of
  multi-source dermatoscopic images of common pigmented skin lesions}.
\newblock \bibinfo{journal}{\emph{Scientific Data}} \bibinfo{volume}{5},
  \bibinfo{number}{1} (\bibinfo{year}{2018}), \bibinfo{pages}{1--9}.
\newblock


\bibitem[UNSCEAR et~al\mbox{.}(2000)]%
        {unscear2000sources}
\bibfield{author}{\bibinfo{person}{UN UNSCEAR} {et~al\mbox{.}}}
  \bibinfo{year}{2000}\natexlab{}.
\newblock \showarticletitle{Sources and effects of ionizing radiation}.
\newblock \bibinfo{journal}{\emph{United Nations Scientific Committee on the
  Effects of Atomic Radiation}} (\bibinfo{year}{2000}).
\newblock


\bibitem[van Assen et~al\mbox{.}(2024)]%
        {vanAssen2024implications}
\bibfield{author}{\bibinfo{person}{Marly van Assen}, \bibinfo{person}{Ashley
  Beecy}, \bibinfo{person}{Gabrielle Gershon}, \bibinfo{person}{Janice
  Newsome}, \bibinfo{person}{Hari Trivedi}, {and} \bibinfo{person}{Judy
  Gichoya}.} \bibinfo{year}{2024}\natexlab{}.
\newblock \showarticletitle{Implications of bias in artificial intelligence:
  considerations for cardiovascular imaging}.
\newblock \bibinfo{journal}{\emph{Current Atherosclerosis Reports}}
  \bibinfo{volume}{26}, \bibinfo{number}{4} (\bibinfo{year}{2024}),
  \bibinfo{pages}{91--102}.
\newblock


\bibitem[{van den Heuvel} et~al\mbox{.}(2018)]%
        {HC18}
\bibfield{author}{\bibinfo{person}{Thomas L.~A. {van den Heuvel}},
  \bibinfo{person}{Dagmar {de Bruijn}}, \bibinfo{person}{Chris~L. {de Korte}},
  {and} \bibinfo{person}{Bram {van Ginneken}}.}
  \bibinfo{year}{2018}\natexlab{}.
\newblock \bibinfo{booktitle}{\emph{Automated Measurement of Fetal Head
  Circumference Using {{2D}} Ultrasound Images}}.
\newblock
\urldef\tempurl%
\url{https://doi.org/10.5281/zenodo.1327317}
\showDOI{\tempurl}


\bibitem[Van~Ginneken et~al\mbox{.}(2001)]%
        {van2001computer}
\bibfield{author}{\bibinfo{person}{Bram Van~Ginneken},
  \bibinfo{person}{BM~Ter~Haar Romeny}, {and} \bibinfo{person}{Max~A
  Viergever}.} \bibinfo{year}{2001}\natexlab{}.
\newblock \showarticletitle{Computer-aided diagnosis in chest radiography: a
  survey}.
\newblock \bibinfo{journal}{\emph{IEEE Transactions on Medical Imaging}}
  \bibinfo{volume}{20}, \bibinfo{number}{12} (\bibinfo{year}{2001}),
  \bibinfo{pages}{1228--1241}.
\newblock


\bibitem[van Royen et~al\mbox{.}(2022)]%
        {van2022developing}
\bibfield{author}{\bibinfo{person}{Florien~S van Royen},
  \bibinfo{person}{Karel~GM Moons}, \bibinfo{person}{Geert-Jan Geersing}, {and}
  \bibinfo{person}{Maarten van Smeden}.} \bibinfo{year}{2022}\natexlab{}.
\newblock \showarticletitle{Developing, validating, updating and judging the
  impact of prognostic models for respiratory diseases}.
\newblock \bibinfo{journal}{\emph{European Respiratory Journal}}
  \bibinfo{volume}{60}, \bibinfo{number}{3} (\bibinfo{year}{2022}).
\newblock


\bibitem[Vapnik(1999)]%
        {vapnik1999overview}
\bibfield{author}{\bibinfo{person}{V.N. Vapnik}.}
  \bibinfo{year}{1999}\natexlab{}.
\newblock \showarticletitle{An overview of statistical learning theory}.
\newblock \bibinfo{journal}{\emph{IEEE Transactions on Neural Networks}}
  \bibinfo{volume}{10}, \bibinfo{number}{5} (\bibinfo{year}{1999}),
  \bibinfo{pages}{988--999}.
\newblock
\urldef\tempurl%
\url{https://doi.org/10.1109/72.788640}
\showDOI{\tempurl}


\bibitem[von Euler-Chelpin et~al\mbox{.}(2019)]%
        {voneuler-chelpin2019sensitivity}
\bibfield{author}{\bibinfo{person}{My von Euler-Chelpin},
  \bibinfo{person}{Martin Lillholm}, \bibinfo{person}{Ilse Vejborg},
  \bibinfo{person}{Mads Nielsen}, {and} \bibinfo{person}{Elsebeth Lynge}.}
  \bibinfo{year}{2019}\natexlab{}.
\newblock \showarticletitle{Sensitivity of screening mammography by density and
  texture: a cohort study from a population-based screening program in
  {Denmark}}.
\newblock \bibinfo{journal}{\emph{Breast Cancer Research}}
  \bibinfo{volume}{21}, \bibinfo{number}{1} (\bibinfo{date}{Oct.}
  \bibinfo{year}{2019}), \bibinfo{pages}{111}.
\newblock
\showISSN{1465-542X}
\urldef\tempurl%
\url{https://doi.org/10.1186/s13058-019-1203-3}
\showDOI{\tempurl}


\bibitem[Wang et~al\mbox{.}(2017)]%
        {wang2017chestx}
\bibfield{author}{\bibinfo{person}{Xiaosong Wang}, \bibinfo{person}{Yifan
  Peng}, \bibinfo{person}{Le Lu}, \bibinfo{person}{Zhiyong Lu},
  \bibinfo{person}{Mohammadhadi Bagheri}, {and} \bibinfo{person}{Ronald~M
  Summers}.} \bibinfo{year}{2017}\natexlab{}.
\newblock \showarticletitle{Chestx-ray8: Hospital-scale chest x-ray database
  and benchmarks on weakly-supervised classification and localization of common
  thorax diseases}. In \bibinfo{booktitle}{\emph{Computer Vision and Pattern
  Recognition (CVPR)}}. \bibinfo{pages}{2097--2106}.
\newblock


\bibitem[Wen et~al\mbox{.}(2022)]%
        {wen2022characteristics}
\bibfield{author}{\bibinfo{person}{David Wen}, \bibinfo{person}{Saad~M Khan},
  \bibinfo{person}{Antonio~Ji Xu}, \bibinfo{person}{Hussein Ibrahim},
  \bibinfo{person}{Luke Smith}, \bibinfo{person}{Jose Caballero},
  \bibinfo{person}{Luis Zepeda}, \bibinfo{person}{Carlos de Blas~Perez},
  \bibinfo{person}{Alastair~K Denniston}, \bibinfo{person}{Xiaoxuan Liu},
  {et~al\mbox{.}}} \bibinfo{year}{2022}\natexlab{}.
\newblock \showarticletitle{Characteristics of publicly available skin cancer
  image datasets: a systematic review}.
\newblock \bibinfo{journal}{\emph{The Lancet Digital Health}}
  \bibinfo{volume}{4}, \bibinfo{number}{1} (\bibinfo{year}{2022}),
  \bibinfo{pages}{e64--e74}.
\newblock


\bibitem[Wen et~al\mbox{.}(2024)]%
        {wen2024data}
\bibfield{author}{\bibinfo{person}{David Wen}, \bibinfo{person}{Andrew Soltan},
  \bibinfo{person}{Emanuele Trucco}, {and} \bibinfo{person}{Rubeta~N Matin}.}
  \bibinfo{year}{2024}\natexlab{}.
\newblock \showarticletitle{From data to diagnosis: skin cancer image datasets
  for artificial intelligence}.
\newblock \bibinfo{journal}{\emph{Clinical and Experimental Dermatology}}
  (\bibinfo{year}{2024}), \bibinfo{pages}{llae112}.
\newblock


\bibitem[Weng et~al\mbox{.}(2023)]%
        {weng2023sex}
\bibfield{author}{\bibinfo{person}{Nina Weng}, \bibinfo{person}{Siavash
  Bigdeli}, \bibinfo{person}{Eike Petersen}, {and} \bibinfo{person}{Aasa
  Feragen}.} \bibinfo{year}{2023}\natexlab{}.
\newblock \showarticletitle{Are Sex-Based Physiological Differences the Cause
  of Gender Bias for Chest X-Ray Diagnosis?}. In
  \bibinfo{booktitle}{\emph{MICCAI Workshop on Clinical Image-Based
  Procedures}}. Springer, \bibinfo{pages}{142--152}.
\newblock


\bibitem[(WHO)(2025)]%
        {who2024maternalmortality}
\bibfield{author}{\bibinfo{person}{World Health~Organization (WHO)}.}
  \bibinfo{year}{2025}\natexlab{}.
\newblock \bibinfo{title}{"Fact sheets: maternal mortality"}.
\newblock
  \bibinfo{howpublished}{https://www.who.int/news-room/fact-sheets/detail/maternal-mortality}.
\newblock
\newblock
\shownote{Accessed: 2024-08-20}.


\bibitem[Wilkinson et~al\mbox{.}(2016)]%
        {wilkinson2016fair}
\bibfield{author}{\bibinfo{person}{Mark~D Wilkinson}, \bibinfo{person}{Michel
  Dumontier}, \bibinfo{person}{IJsbrand~Jan Aalbersberg},
  \bibinfo{person}{Gabrielle Appleton}, \bibinfo{person}{Myles Axton},
  \bibinfo{person}{Arie Baak}, \bibinfo{person}{Niklas Blomberg},
  \bibinfo{person}{Jan-Willem Boiten}, \bibinfo{person}{Luiz~Bonino da
  Silva~Santos}, \bibinfo{person}{Philip~E Bourne}, {et~al\mbox{.}}}
  \bibinfo{year}{2016}\natexlab{}.
\newblock \showarticletitle{The {FAIR Guiding Principles} for scientific data
  management and stewardship}.
\newblock \bibinfo{journal}{\emph{Scientific data}} \bibinfo{volume}{3},
  \bibinfo{number}{1} (\bibinfo{year}{2016}), \bibinfo{pages}{1--9}.
\newblock


\bibitem[Winkler et~al\mbox{.}(2019)]%
        {winkler2019association}
\bibfield{author}{\bibinfo{person}{Julia~K Winkler}, \bibinfo{person}{Christine
  Fink}, \bibinfo{person}{Ferdinand Toberer}, \bibinfo{person}{Alexander Enk},
  \bibinfo{person}{Teresa Deinlein}, \bibinfo{person}{Rainer
  Hofmann-Wellenhof}, \bibinfo{person}{Luc Thomas}, \bibinfo{person}{Aimilios
  Lallas}, \bibinfo{person}{Andreas Blum}, \bibinfo{person}{Wilhelm Stolz},
  {et~al\mbox{.}}} \bibinfo{year}{2019}\natexlab{}.
\newblock \showarticletitle{Association between surgical skin markings in
  dermoscopic images and diagnostic performance of a deep learning
  convolutional neural network for melanoma recognition}.
\newblock \bibinfo{journal}{\emph{JAMA Dermatology}} \bibinfo{volume}{155},
  \bibinfo{number}{10} (\bibinfo{year}{2019}), \bibinfo{pages}{1135--1141}.
\newblock


\bibitem[Wynants et~al\mbox{.}(2020)]%
        {wynants2020prediction}
\bibfield{author}{\bibinfo{person}{Laure Wynants}, \bibinfo{person}{Ben
  Van~Calster}, \bibinfo{person}{Gary~S Collins}, \bibinfo{person}{Richard~D
  Riley}, \bibinfo{person}{Georg Heinze}, \bibinfo{person}{Ewoud Schuit},
  \bibinfo{person}{Elena Albu}, \bibinfo{person}{Banafsheh Arshi},
  \bibinfo{person}{Vanesa Bellou}, \bibinfo{person}{Marc~MJ Bonten},
  {et~al\mbox{.}}} \bibinfo{year}{2020}\natexlab{}.
\newblock \showarticletitle{Prediction models for diagnosis and prognosis of
  covid-19: systematic review and critical appraisal}.
\newblock \bibinfo{journal}{\emph{bmj}}  \bibinfo{volume}{369}
  (\bibinfo{year}{2020}).
\newblock


\bibitem[Yan et~al\mbox{.}(2020)]%
        {Yan2020}
\bibfield{author}{\bibinfo{person}{Wenjun Yan}, \bibinfo{person}{Lu Huang},
  \bibinfo{person}{Liming Xia}, \bibinfo{person}{Shengjia Gu},
  \bibinfo{person}{Fuhua Yan}, \bibinfo{person}{Yuanyuan Wang}, {and}
  \bibinfo{person}{Qian Tao}.} \bibinfo{year}{2020}\natexlab{}.
\newblock \showarticletitle{MRI manufacturer shift and adaptation: Increasing
  the generalizability of deep learning segmentation for MR images acquired
  with different scanners}.
\newblock \bibinfo{journal}{\emph{Radiology: Artificial Intelligence}}
  \bibinfo{volume}{2} (\bibinfo{year}{2020}), \bibinfo{pages}{1--10}.
\newblock
Issue 4.
\showISSN{26386100}
\urldef\tempurl%
\url{https://doi.org/10.1148/ryai.2020190195}
\showDOI{\tempurl}


\bibitem[Yang et~al\mbox{.}(2024)]%
        {yang2024navigating}
\bibfield{author}{\bibinfo{person}{Xinyu Yang}, \bibinfo{person}{Weixin Liang},
  {and} \bibinfo{person}{James Zou}.} \bibinfo{year}{2024}\natexlab{}.
\newblock \showarticletitle{{Navigating Dataset Documentations in AI: A
  Large-Scale Analysis of Dataset Cards on HuggingFace}}. In
  \bibinfo{booktitle}{\emph{International Conference on Learning
  Representations (ICLR)}}.
\newblock


\bibitem[Yi et~al\mbox{.}(2019)]%
        {yi2019generative}
\bibfield{author}{\bibinfo{person}{Xin Yi}, \bibinfo{person}{Ekta Walia}, {and}
  \bibinfo{person}{Paul Babyn}.} \bibinfo{year}{2019}\natexlab{}.
\newblock \showarticletitle{Generative adversarial network in medical imaging:
  A review}.
\newblock \bibinfo{journal}{\emph{Medical Image Analysis}}
  \bibinfo{volume}{58} (\bibinfo{year}{2019}), \bibinfo{pages}{101552}.
\newblock


\bibitem[Yoon et~al\mbox{.}(2023)]%
        {Yoon2023}
\bibfield{author}{\bibinfo{person}{Sung~Hyun Yoon}, \bibinfo{person}{Sunyoung
  Park}, \bibinfo{person}{Sowon Jang}, \bibinfo{person}{Junghoon Kim},
  \bibinfo{person}{Kyung~Won Lee}, \bibinfo{person}{Woojoo Lee},
  \bibinfo{person}{Seungjae Lee}, \bibinfo{person}{Gabin Yun}, {and}
  \bibinfo{person}{Kyung~Hee Lee}.} \bibinfo{year}{2023}\natexlab{}.
\newblock \showarticletitle{Use of artificial intelligence in triaging of chest
  radiographs to reduce radiologists’ workload}.
\newblock \bibinfo{journal}{\emph{European Radiology}} \bibinfo{volume}{34},
  \bibinfo{number}{2} (\bibinfo{year}{2023}), \bibinfo{pages}{1094–1103}.
\newblock
\showISSN{1432-1084}
\urldef\tempurl%
\url{https://doi.org/10.1007/s00330-023-10124-1}
\showDOI{\tempurl}


\bibitem[Yu et~al\mbox{.}(2023)]%
        {yu2023dataset}
\bibfield{author}{\bibinfo{person}{Ruonan Yu}, \bibinfo{person}{Songhua Liu},
  {and} \bibinfo{person}{Xinchao Wang}.} \bibinfo{year}{2023}\natexlab{}.
\newblock \showarticletitle{Dataset distillation: A comprehensive review}.
\newblock \bibinfo{journal}{\emph{IEEE Transactions on Pattern Analysis and
  Machine Intelligence}} (\bibinfo{year}{2023}).
\newblock


\bibitem[Zaj{\k{a}}c et~al\mbox{.}(2023)]%
        {zajkac2023ground}
\bibfield{author}{\bibinfo{person}{Hubert~Dariusz Zaj{\k{a}}c},
  \bibinfo{person}{Natalia~Rozalia Avlona}, \bibinfo{person}{Finn Kensing},
  \bibinfo{person}{Tariq~Osman Andersen}, {and} \bibinfo{person}{Irina
  Shklovski}.} \bibinfo{year}{2023}\natexlab{}.
\newblock \showarticletitle{{Ground Truth Or Dare: Factors Affecting The
  Creation Of Medical Datasets For Training {AI}}}. In
  \bibinfo{booktitle}{\emph{Conference on AI, Ethics, and Society (AIES)}}.
  \bibinfo{pages}{351--362}.
\newblock


\bibitem[Zech et~al\mbox{.}(2018)]%
        {zech2018variable}
\bibfield{author}{\bibinfo{person}{John~R Zech}, \bibinfo{person}{Marcus~A
  Badgeley}, \bibinfo{person}{Manway Liu}, \bibinfo{person}{Anthony~B Costa},
  \bibinfo{person}{Joseph~J Titano}, {and} \bibinfo{person}{Eric~Karl
  Oermann}.} \bibinfo{year}{2018}\natexlab{}.
\newblock \showarticletitle{Variable generalization performance of a deep
  learning model to detect pneumonia in chest radiographs: a cross-sectional
  study}.
\newblock \bibinfo{journal}{\emph{PLoS Medicine}} \bibinfo{volume}{15},
  \bibinfo{number}{11} (\bibinfo{year}{2018}), \bibinfo{pages}{e1002683}.
\newblock


\bibitem[Zhang et~al\mbox{.}(2022)]%
        {zhang2022improving}
\bibfield{author}{\bibinfo{person}{Haoran Zhang}, \bibinfo{person}{Natalie
  Dullerud}, \bibinfo{person}{Karsten Roth}, \bibinfo{person}{Lauren
  Oakden-Rayner}, \bibinfo{person}{Stephen Pfohl}, {and}
  \bibinfo{person}{Marzyeh Ghassemi}.} \bibinfo{year}{2022}\natexlab{}.
\newblock \showarticletitle{Improving the fairness of chest X-ray classifiers}.
  In \bibinfo{booktitle}{\emph{Conference on Health, Inference, and Learning
  (CHIL)}}. PMLR, \bibinfo{pages}{204--233}.
\newblock


\end{thebibliography}

\clearpage
\appendix
\setcounter{figure}{0}
\setcounter{table}{0}

\renewcommand\thefigure{\thesection\arabic{figure}}
\renewcommand\thetable{\thesection\arabic{table}}


\section{Shortcuts}
\begin{figure}[h!]
    \centering
    \includegraphics[width=1.0\columnwidth]{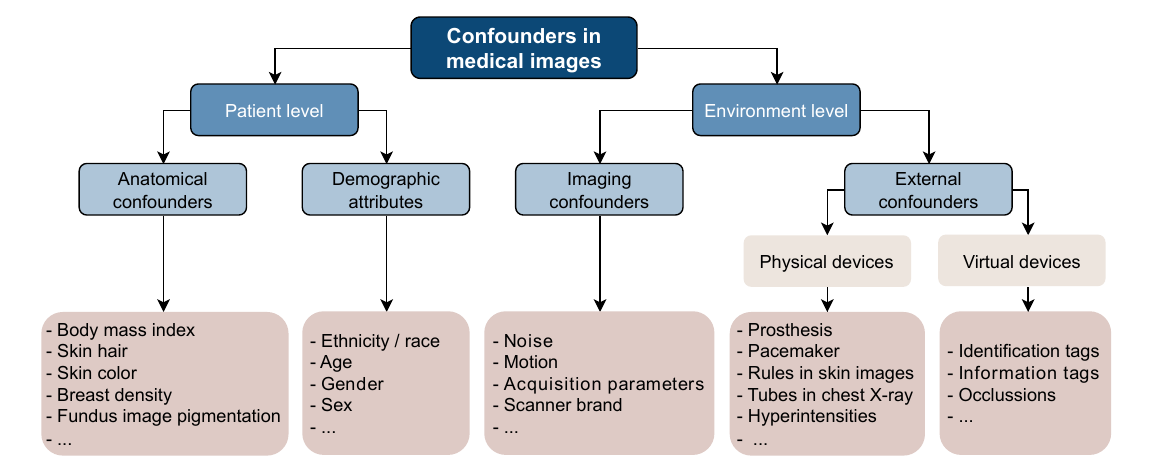}
    \caption{MICCAT: Medical Imaging Contextualized Confounder Taxonomy. Reproduced from \cite{juodelyte2024source} with permission of the authors.}
    \label{fig:miccat}
    \Description{MICCAT: Medical Imaging Contextualized Confounder Taxonomy. The patient level can be further classified into anatomical confounders and demographic attributes. The environment level can be further classified into imaging confounders and external confounders (physical or virtual devices).}
\end{figure}

Machine learning models are prone to rely on spurious correlations to make predictions, which are usually easier to detect than the genuine disease patterns, as explained in Section~\ref{sec:shortcuts}. Such shortcuts can be well-localized objects in the image -- e.g., mechanical ventilation tubes or pacemakers, which are often present in patients with certain diseases-- or more global ones, like specific noise patterns or intensity distributions associated to a certain acquisition setting or device brand. Demographic attributes like sex/gender, age or ethnicity can also lead to shortcut learning, disproportionately impacting historically underserved subgroups \cite{banerjee2023shortcuts}, especially when datasets are highly imbalanced. To categorize the different types of shortcuts, we adopt the Medical Imaging Contextualized Confounder Taxonomy (MICCAT) \cite{juodelyte2024source}, see Fig.~\ref{fig:miccat}, which we believe can be useful to understand how spurious correlations arise and how they can be mitigated. MICCAT extends beyond traditional demographic attributes, such as sex/gender, age, and ethnicity, to include a broader set of confounders that are domain- and context-specific. These confounders encompass patient-level and environment-level factors.

Patient-level confounders include both demographic attributes and anatomical confounders. Demographic attributes, such as gender \cite{larrazabal2020gender,abbasi2020risk}, age \cite{abbasi2020risk}, and ethnicity \cite{gichoya2022ai}, represent standard factors typically considered in bias analysis. Anatomical confounders, on the other hand, refer to physical or disease-related characteristics specific to organs or conditions. Examples include body mass index, tissue density, breast density, and bone density, as well as combinations of these factors. Such anatomical variations may define subgroups where models underperform, and their identification often requires analysis beyond standard demographic characteristics \cite{voneuler-chelpin2019sensitivity}.

Environment-level confounders include both external and imaging confounders. External confounders arise from physical or virtual elements within the image, such as chest drains \cite{oakden2020hidden,jimenez2023detecting}, pen marks near skin lesions \cite{winkler2019association}, patient positioning \cite{degrave2021ai}, or text and measurement calipers \cite{lin2024shortcut}. These elements typically produce localized artifacts visible to the human eye. In contrast, imaging confounders result from variations in the imaging process, including differences in equipment brands, scanner types, acquisition parameters, noise, or motion artifacts. Such confounders generally create global artifacts that affect the entire image and may not be perceptible to the human eye. Systematic differences, such as variations in exposure settings for chest X-rays \cite{lang2024using} or distinctive characteristics of imaging devices used across different medical centers \cite{compton2023more}, can unintentionally introduce shortcuts for machine learning models. Rather than learning clinically relevant features, models may instead rely on these acquisition-specific factors, which are associated with disease labels but do not represent true underlying medical conditions \cite{ong2024shortcut}.

\section{Clinical case studies} \label{sec:casestudies}

We present four case studies, illustrated in Fig.~\ref{fig:casestudies}, with key points highlighted in Table~\ref{tab:casestudies}.

\ourpar{Case study 1: chest X-rays.}
Chest X-rays are the most commonly performed radiologic examinations worldwide~\cite{unscear2000sources}, requiring significant expertise for accurate and meaningful interpretation~\cite{van2001computer}. The advent of artificial intelligence is a game changer for automatic chest X-ray diagnosis, marked by the release of the large-scale NIH-CXR14 dataset~\cite{wang2017chestx}. This milestone spurred a wave of studies, including claims of machine learning models achieving radiologist-level performance, exemplified by CheXNet's pneumonia detection model~\cite{rajpurkar2017chexnet}. However, these claims that have been criticized for relying on shortcuts, such as the use of chest drains for pneumothorax classification \cite{oakden2020hidden, jimenez2023detecting}, and for demonstrating low inter-rater agreement for pathologies like pneumonia \cite{damgaard2023augmenting}. The latter may be attributed to the fact that pneumonia is a differential diagnosis requiring clinical information -- laboratory tests, physical examinations, symptoms and signs, etc. -- that extends beyond chest X-ray images \cite{bustos2020padchest}. Subsequently, additional datasets such as CheXpert~\cite{irvin2019chexpert}, MIMIC-CXR~\cite{johnson2019mimic}, and PadChest~\cite{bustos2020padchest} have become widely used in the research community. 

With the primary purpose of diagnosing and/or monitoring treatment of various pathologies, those chest X-ray dataset often include associated demographic information such as age, gender/sex, and race/ethnicity, which enables further analysis of potential disparities and biases in automatic chest X-ray diagnosis. Among these key demographic attributes, age and sex/gender are commonly included in most datasets, whereas race/ethnicity is available in only a few, such as MIMIC-CXR and CheXpert~\cite{paul2022demographic}.

Chest X-ray datasets are generally skewed toward older populations, with PadChest's median age at 62, and MIMIC-CXR's largest group aged 60–80~\cite{zhang2022improving}. Studies have shown that age can be predicted from chest X-rays~\cite{ieki2022deep}, raising concerns about unintended information leakage favoring well-represented age groups. Widely used chest X-ray datasets show no significant gender distribution imbalance (e.g., NIH-CXR14: \{M:56.5\%, F:43.5\%\}, CheXpert: \{M:58.7\%, F:41.3\%\}). However, performance disparities between male and female groups in disease classifiers persist~\cite{larrazabal2020gender,seyyed2020chexclusion,seyyed2021underdiagnosis}, with unclear causes. Balancing the data has proven ineffective. Studies have ruled out hypothesis like under-representation~\cite{larrazabal2020gender}, physiological differences (e.g., the presence of breasts)~\cite{weng2023sex} and shortcut learning (e.g., support devices like chest drains~\cite{olesen2024slicing, oakden2020hidden, jimenez2023detecting}).

Race and ethnicity are important demographic attributes, but are rarely included in chest X-ray datasets. Among 23 reviewed datasets \cite{paul2022demographic}, only MIMIC-CXR and CheXpert report self-reported race, with predominantly white populations accounting for 60.7\% and 56.4\% of samples, respectively. Performance disparities between racial groups favoring white individuals have been noted \cite{seyyed2020chexclusion}. Studies show deep learning can infer protected attributes like race from chest X-rays, despite this being a challenging task for human experts~\cite{gichoya2022ai}. This raises critical concerns about the potential use of protected features in the decision-making process, which could lead to unfair biases. Recent research~\cite{glocker2023algorithmic} suggests that transfer learning alone cannot confirm the influence of protected attributes; instead combining methods like test-set resampling, multitask learning, and model inspection provides offer better insights within neural network feature representations.

\ourpar{Case study 2: Skin lesions}
Over the past few years, the use of computer vision algorithms in dermatology has experienced significant growth~\cite{naqvi2023skin}. This surge may be largely explained by the availability of publicly accessible skin lesion datasets -- such as Fitzpatrick17k~\cite{groh2021evaluating}, PAD-UFES-20~\cite{pacheco2020pad}, and HIBA~\cite{ricci2023dataset} -- as well as the International Skin Imaging Collaboration (ISIC) --, which aggregates different datasets such as HAM10000~\cite{tschandl2018ham10000} and BCN 20000 \cite{combalia2019bcn20000}, in an archive -- the ISIC Archive\footnote{\url{https://www.isic-archive.com/}}) that contains more than 490K skin lesion images. Dermoscopy and clinical photography are common imaging modalities in the development of ML models that address tasks such as classification~\cite{pacheco2021attention}, segmentation~\cite{mirikharaji2023survey}, and lesion localization \cite{maqsood2023multiclass}.

Despite the progress made in the field, the under-representation of demographic groups in skin lesion datasets limits the generalizability of ML models~\cite{daneshjou2021lack}. Skin lesions may manifest differently across populations, influenced by factors like skin tone, genetic background, age and UV light exposure \cite{wen2024data}. Nonetheless, most publicly available datasets are limited in terms of geographic and skin tone diversity, predominantly featuring of patients with Fitzpatrick skin types I to III~\cite{wen2022characteristics, groh2022towards}. This lack of diversity may lead to biased models that underperform for under-represented groups. For example, melanoma -- the deadliest skin cancer -- is much more prevalent in fair-skinned individuals but can also occur in those with darker skin tones, which may be misdiagnosed due to the limited representation in training data \cite{groh2024deep}.

Recent efforts to diversify skin lesion datasets, such as the inclusion of PAD-UFES-20~\cite{pacheco2020pad} and HIBA~\cite{ricci2023dataset} datasets in the ISIC archive, have improved the representation of Latin American individuals. A region that was previously under-represented in the ISIC archive. However, there is still a strong lack of representation in terms of diversity of skin tone. In this context, addressing these disparities is a global challenge that requires a collaborative effort from the research community, drawing on diverse perspectives and contributions from different backgrounds and regions worldwide. Such initiatives are important to enhance fairness and promote equitable AI solutions for skin lesion analysis, working toward technologies that better serve patients across diverse demographic groups.

\ourpar{Case study 3: fetal ultrasound}

Ultrasound is the fundamental imaging modality for antenatal care. The acquisition process consists of a physical examination with an ultrasound probe of the pregnant woman's womb looking for specific two-dimensional slices, called standard planes. After the location of one of these planes, the clinical expert often performs a series of annotations on the image to extract measurements of several structures of interest like the fetal head, femur, heart or abdomen, the placenta or the maternal cervix, among others, that are essential for the assessment of the pregnancy prognosis and the fetal well-being and growth. Several sources of variation might affect the image quality, such as the maternal body mass index (BMI) or the experience of the sonographer acquiring the scan. Maternal obesity, in particular, results in a higher difficulty to complete a full survey and a decreased visualization of fetal anatomies, being the face and the heart the most difficult anatomies to observe~\citep{dasheObesity2009,tsaiObesity2015}. One would expect detection accuracy disparities across varying BMI values, which could result in unfair predictions if not accounted for.

Another important factor is the ethnicity, which is associated with variations in the normal fetal growth according to several multi-ethnic studies~\citep{droogerEthnic2005,ogasawaraVariation2009,sletnerEthnic2015}. Thus, growth charts based on image biomarkers, such as the head circumference or the femur length, need to be adapted to the local population.

With all this evidence, ML models trained with one cohort will inevitable exhibit biases toward a specific population subgroup, emphasizing the importance of multi-cohort and multi-variate analysis. Unfortunately, few or no evidence is reported on the effect of these biases for ML models. One of the reasons might be the scarcity of existing open fetal ultrasound datasets, with none, to date, providing demographic information. The few available datasets are the Fetal Planes DB~\citep{FetalPlanesDB} and data from the HC18~\citep{HC18}, FH-PS-AOP~\citep{AOP}, and ACOUSLIC-AI~\citep{ACOUSLIC} challenges. Overall, the analysis of fetal ultrasound imaging is either conducted by teams with access to their own local cohort or constrained to the few existing open datasets and tasks, making the multi-cohort analysis very difficult. Hence, there is a strong need for more diverse datasets with detailed demographic information, including BMI and ethnicity, to develop fair and unbiased models. Among others applications, this could allow for the implementation of robust ML-based tools in low-resource settings through portable devices to facilitate prenatal screening and try to reduce the pressing challenge of fetal mortality in low- and middle-income countries.

\ourpar{Case study 4: brain MRI}
MRI is the third most commonly performed imaging modality after CT and X-rays. Its superior soft tissue contrast enables detailed visualization of brain anatomy, making it significantly more sensitive and specific for detecting abnormalities within the brain. This capability explains why most public MRI datasets available for AI research focus on the brain \cite{Dishner2024}. Key application areas for brain MRI and AI, along with some of the most notable datasets, include neurodegenerative diseases, represented by the OASIS Brains project datasets for Alzheimer’s disease \cite{marcus2007oasis, marcus2010open, lamontagne2019oasis, koenig2020oasis}; brain cancer, with datasets like the BraTS Challenge \cite{Baid2021}, LUMIERE \cite{Suter2022}, Ocana \cite{ocana2023comprehensive}, RHUH-GBM \cite{cepeda2023rio}, and TCGA-GBM \cite{scarpace2016cancer}; and stroke, with ISLES 2022 \cite{hernandez2022isles}, ATLAS v2.0 \cite{Liew2022}, among others. The neuroimaging community has made significant progress in advancing data-sharing practices by adopting standards like BIDS \cite{Gorgolewski2016} and employing tools like DataLad \cite{Halchenko2021} for versioning and tracking. Representative platforms for hosting neuroimaging datasets include NITRC.org \cite{kennedy2016nitrc} and OpenNeuro \cite{markiewicz2021openneuro}, with the latter adhering to the BIDS standard and hosting over 700 brain MRI datasets, as identified through our web scraping analysis. Additionally, tools for converting legacy datasets to these formats are actively being developed \cite{Routier2021}. 

The conversion from the standard clinical imaging format, DICOM, to NIfTI or other formats is complex and error-prone \cite{Li2016} and depends on how different formats implement DICOM standards. Several practical considerations drive the conversion. From a data science perspective, the DICOM standard's slice-wise storage of imaging volumes is less efficient for processing compared to volumetric formats like NIfTI. Moreover, most data science tools and libraries lack robust support for direct manipulation of DICOM files, favoring instead the use of formats designed for streamlined computational workflows. While these conversions simplify data processing pipelines, they often lead to an under-appreciation of the extensive clinical metadata embedded in DICOM files. This metadata is crucial because it captures details such as field strength, acquisition parameters, and vendor-specific variations—factors known to significantly influence the generalizability of deep learning models. Studies have demonstrated that neglecting these elements can reduce model robustness across different imaging settings \cite{Yan2020, Keenan2022, Kushol2023}, which is also reflected in the high interest in the field of domain adaptation for medical imaging applications \cite{guan2021domain, fang2024source}.

Many publicly available neuroimaging datasets undergo extensive preprocessing prior to release. Often they are standardized for open challenges, ensuring model evaluations focus on algorithm performance rather than preprocessing effects. While this approach has facilitated the comparison of methods within open challenges, it has also contributed to a disconnect between these standardized datasets and the complexities of real-world clinical data. This disconnect may inadvertently hinder technological advancements aimed at improving preprocessing techniques themselves. Skull stripping, also referred to as brain extraction, is a preprocessing step specific to neuroimaging that involves removing the skull and extracranial structures. It is particularly useful in tasks where these structures might lead to false-positive detections or over-segmentation. Additionally, skull stripping is often employed as a privacy mechanism, especially in structural MRI, to eliminate facial features. However, if performed improperly, it can result in the unintended removal of brain tissue, posing significant challenges in neuroimaging tasks where lesions are located near the brain’s periphery, such as meningiomas. Intensity normalization, while commonly used to address the non-quantitative nature of MRI, is among the most destructive preprocessing steps, as the original intensity values cannot be recovered post-normalization. This irreversible transformation can limit the utility of datasets for certain applications. Examples of datasets shared with minimal preprocessing are the ISLES 2022 dataset \cite{hernandez2022isles}, which was provided in nearly raw form, with skull stripping performed solely for anonymization purposes and data shared in NIfTI format, and the ISLES 2024 dataset, which was shared in NIfTY format as well as anonymized DICOMS. Such datasets offer valuable opportunities to study and address the challenges associated with preprocessing in a more authentic representation of real-world data.

Finally, reporting of demographics in brain MRI is generally very poor. For example, \cite{Sterling2022} analyzed demographic reporting in MR neuroimaging studies in the US over the past decade and found that biological sex was reported in 77\% of studies, while race and ethnicity were reported in only 10\% and 4\%, respectively. Recent efforts have assessed the impact of demographic factors on model performance. In \cite{dibaji2023studying}, disparities were found in brain age prediction models across sex subgroups and datasets. Similarly, \cite{ioannou2022study} found race bias was more pronounced than sex bias on CNN-based MR segmentation, with Black females being the most affected subgroup.

In conclusion, MRI remains a cornerstone imaging modality for neuroimaging research, driven by its unparalleled ability to visualize brain anatomy with high sensitivity and specificity. The wealth of publicly available brain MRI datasets has enabled significant advances in AI applications, but these datasets are often limited by extensive preprocessing and a lack of inclusion of key demographic information. Preprocessing steps, while facilitating standardized comparisons in open challenges, can obscure the challenges inherent in real-world data and hinder progress in improving preprocessing pipelines themselves. Additionally, the underutilization of rich metadata in DICOM files and the complexities of format conversions further highlight the need for tools and practices that bridge the gap between clinical imaging and AI research. Efforts to address demographic disparities in neuroimaging datasets are beginning to emerge, as seen in recent efforts like the addition of children (BraTS-PEDS~\cite{kazerooni2024bratspeds}) and Sub-Saharan African populations (BraTS-Africa~\cite{adewole2023brain}) to BraTS. Moving forward, the neuroimaging community must prioritize practices that maintain the authenticity of raw imaging data, incorporate diverse populations, and leverage metadata more effectively to maximize the potential of AI in medical imaging.

\setcounter{figure}{0}
\section{Living review} \label{sec:applivingreview} 

In Section~\ref{sec:livingreview}, we propose a proof of concept for the faliving review to keep track of medical imaging datasets and their related research artifacts. The resources that we consider as research artifacts are broadly classified into two categories, see Table~\ref{tab:resources}: 
\begin{itemize}
   \item resource roles: describe the purpose the artifact serves in the research process, generally categorized into material or method, and
    \item resource types: describe the format or content of the artifact, regardless or its role. 
\end{itemize}
We map the relation between the datasets and the research artifacts with the citation function (use, produce, extend, introduce, other), see Table \ref{tab:citationfunctions}. We show research artifacts and their citation function for CheXpert dataset in Table~\ref{tab:relations}.

Our living review consists of three parts, as illustrated in Fig.~\ref{fig:livingreviewzenododb}: 
\begin{enumerate} 
\item an overarching living review publication,
\item documentation of research artifacts via dataset-specific publications on Zenodo, which the overarching paper links to,
\item a SQL database for exploring the links between the datasets and the research artifacts. 
\end{enumerate}

Our demo is available at~\url{http://130.226.140.142}, and a screenshot is shown in Fig.~\ref{fig:demo}.

\ourpar{Technical details}
We collect datasets, papers, and their relations into a SQL database. We visualize the database with DBeaver as shown in Fig.~\ref{fig:database}. The database comprise of three tables: one for datasets, another for papers, and a third one for datasets usages. The datasets table includes details such as the dataset ID, name and modality of the dataset. The papers table contains information on the ID, BibTeX key name, DOI, and arXiv link. The datasets usages table stores the ID, paper\_ID, dataset\_ID, annotations, shortcuts, and the taxonomy to which the confounder belongs. Example of annotations include chest drains, segmentation masks, and bounding boxes. Example of shortcuts include dark corners and demographic attributes. Our demo is built with PostgreSQL, Python, Streamlit and the \texttt{st\_link\_analysis} package. 
\setcounter{figure}{0}
\setcounter{figure}{0}\begin{figure}[]
    \centering
    \includegraphics[width=0.95\columnwidth]{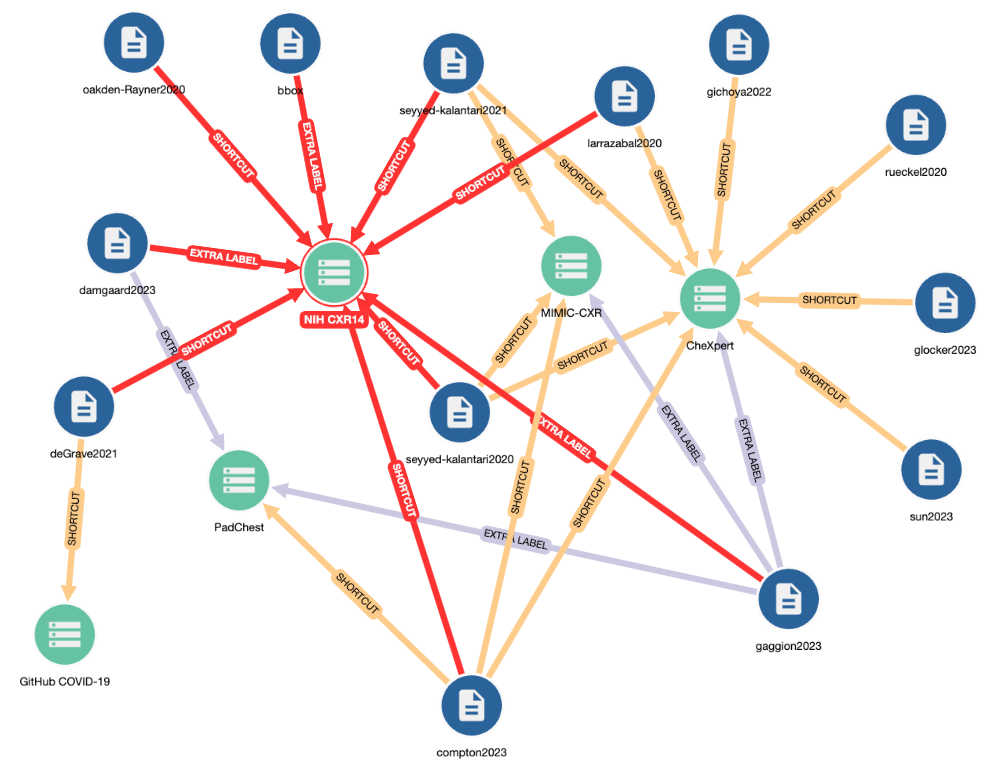}
    \caption{A screenshot of our demo living review. We highlight the relations for NIH-CXR14 dataset.}
    \Description{Visualization of the living review as a graph node, highlighting the relationships for the NIH-CXR14 dataset.}
    \label{fig:demo}
\end{figure}

\begin{figure*}[b]
    \centering
    \includegraphics[width=0.8\linewidth]{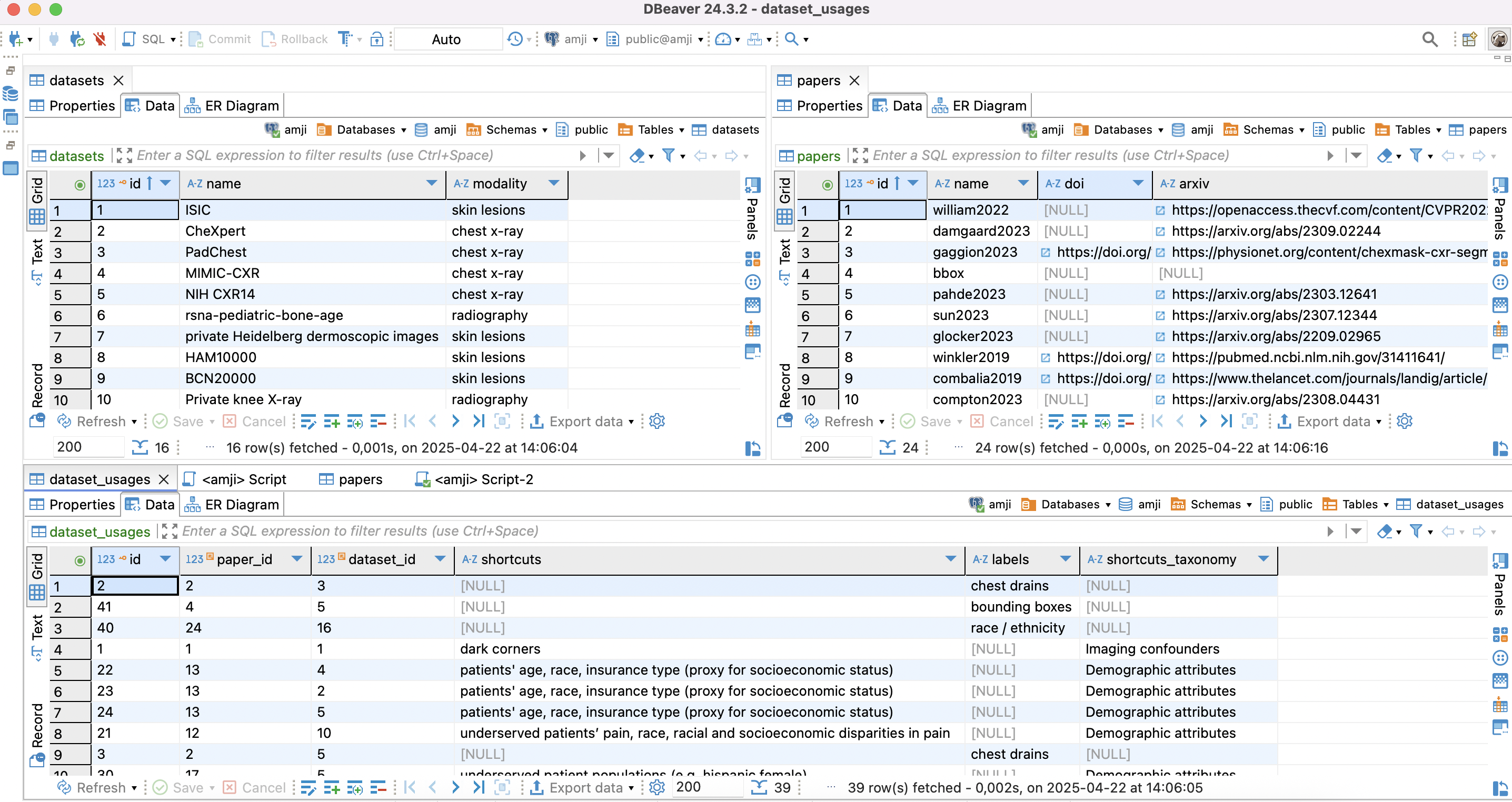}
    \caption{A screenshot of our SQL database in DBeaver.}
    \label{fig:database}
    \Description{A screenshot of our SQL database in DBeaver displaying the three tables: datasets, papers and dataset usages.}
\end{figure*}
\begin{table}[b]
    \centering
    \begin{tabular}{p{0.2\columnwidth}p{0.3\columnwidth}p{0.4\columnwidth}}
    \textbf{Resource role} & \textbf{Resource type} & \textbf{Description} \\ [0.6ex] 
    \hline
    \multirow{2}{*}{}Material & Dataset & corpus, image sets, etc. \\
                               & Additional labels & segmentation masks, demographic information, report of errors, etc. \\ 
                               & Shortcuts & resource finds evidence of errors in the annotations or shortcuts in the dataset. \\ \hline
    \multirow{2}{*}{}Method & Tool & toolkit, software, system, etc. \\ 
    & Code &  codebase, library, API, etc. \\ \hline
    Mixed & Mixed & citations referring to multiple resources \\
    \bottomrule
    \end{tabular}
    \caption{List of resource roles and resource types.} 
    \label{tab:resources}
\end{table}
\begin{table}[b]
\centering
    \begin{tabular}{p{0.2\columnwidth}p{0.8\columnwidth}}
    \textbf{Citation function} & \textbf{Description} \\ [0.6ex] 
    \hline
    Use & Used in the citing paper’s research \\
    \hline
    Produce & First produced or released by the citing paper’s research \\ 
    \hline
    Extend & Used in the citing paper’s research but are improved, upgraded, or changed to work for other problems in the course of the research \\
    \hline
    Introduce & The resources or the related information (\textit{e.g.}, background, applications) are introduced \\
    \hline
    Other & The citation does not belong to the above categories \\
    \bottomrule
    \end{tabular}
\caption{List of citation functions.} 
\label{tab:citationfunctions}
\end{table}
\begin{table}[b]
    \centering
    \begin{tabular}{p{0.32\columnwidth}p{0.15\columnwidth}p{0.24\columnwidth}p{0.2\columnwidth}}
    \textbf{Research artifact} & \textbf{Citation function} & \textbf{Resource type} & \textbf{Example} \\ [0.6ex] 
    \hline
     \citet{irvin2019chexpert} & Produce & Dataset & \chexpert{} dataset \\ \hline
     \citet{rajpurkar2022ai} & Use & - & - \\ \hline
     \citet{garbin2021structured} & Introduce & Documentation & Datasheet~\cite{gebru2021datasheets} for \chexpert \\ \hline
     \citet{larrazabal2020gender} & Extend & Shortcut & Gender bias \\ \hline
     \citet{gaggion2023chexmask} & Extend & Additional annotation & Segmentation masks \\
    \bottomrule
    \end{tabular}
    \caption{Example of research artifacts related to \chexpert{} dataset.} 
    \label{tab:relations}
\end{table}
\begin{table*}[]
\def\arraystretch{1.1}
\centering
    \begin{tabular}{p{0.1\columnwidth}p{0.34\columnwidth}p{0.34\columnwidth}p{0.34\columnwidth}p{0.34\columnwidth}p{0.34\columnwidth}} 
    \textbf{Case study} & \textbf{Tasks and Datasets} & \textbf{Annotation practices} & \textbf{Demographics} & \textbf{Shortcuts} & \textbf{Data lifecycle}  \\ [0.6ex] 
    \hline
    \multirow{2}{*}{\rotatebox[origin=c]{90}{1: Chest X-rays\ } } & T: Classification of chest X-ray diseases & Experience radiologists \textit{vs.} automatic NLP extraction & Self-reported race/ethinicity & Chest drains, hospital scanner, radiographic markers \\ 
    & D: NIH-CXR14, CheXpert, PadChest, MIMIC-CXR & & Skewed towards elder patients \\
    \hline
    \multirow{2}{*}{\rotatebox[origin=c]{90}{2: Skin lesions\  }} & T: Classification, segmentation and lesion localization. & & Skin type, genetic background & Dark corners, hair, patches, rulers, ink marking/staining \\ 
    & D: ISIC, HIBA, PAD-UFES-20, HAM10000, Fitzpatrick17k & & Recent additions of demographics with HIBA and PAD-UFES-2 \\
    \hline
    \multirow{2}{*}{\rotatebox[origin=c]{90}{3: Fetal ultrasound \  }} & T: 2D plane localization, fetal biometrics estimation & Experience of the sonographer might affect the image quality & None to date \\[16pt]
    & D: Fetal Planes DB, HC18, ACOUSLIC-AI, FH-FS-AOP \vspace*{10pt} \\[6pt] \hline
    \multirow{2}{*}{\rotatebox[origin=c]{90}{4:Neuroimaging\  }} & T: Neurogenerative diseases and stroke detection. & & Poorly reported: in US studies (2010–2020), 77\% report sex, but only 10\% report race and 4\% ethnicity.
    & & Standards (e.g. BIDS, preprocessing, conversion from DICOM to NIfTI format) \\
    & D: OASIS, LUMIERE, BraTS, Ocana, TCGA-GBM, ATLAS v2.0, ISLES 2022 & & Recent inclusion of children and Sub-Saharan African populations in BraTS & & Data sharing platforms like Datalad, NITRC.org, OpenNeuro \\
    \bottomrule
    \end{tabular}
\caption{Summary highlighting various aspects of the four case studies. T: tasks; D: datasets. Note: this table is not exhaustive but summarizes key points discussed in the manuscript.} 
\label{tab:casestudies}
\end{table*}

\end{document}